\documentclass{article}

\usepackage{arxiv}
\usepackage[utf8]{inputenc} % allow utf-8 input
\usepackage[T1]{fontenc}    % use 8-bit T1 fonts
\usepackage{hyperref}       % hyperlinks
\usepackage{url}            % simple URL typesetting
\usepackage{booktabs}       % professional-quality tables
\usepackage{array}
\usepackage{amsfonts}       % blackboard math symbols
\usepackage{nicefrac}       % compact symbols for 1/2, etc.
\usepackage{multirow} 
\usepackage{graphicx}
\usepackage{subcaption}% for subfigs
\usepackage{adjustbox}
\usepackage[numbers]{natbib}
\usepackage{doi}
\usepackage{amsmath}
\usepackage{amssymb}
\usepackage{algorithm}
\usepackage{algpseudocode}
\usepackage{xcolor}
\usepackage{fancyhdr}
\usepackage{amsthm}
\usepackage{cleveref}
\usepackage{txfonts}
\usepackage{booktabs}
\usepackage{siunitx} 
\usepackage{appendix}
\newtheorem{remark}{Remark}
\usepackage{tablefootnote}

\newtheoremstyle{remarkstyle}  % Name
  {5pt}                        % Space above
  {5pt}                        % Space below
  {}                           % Body font (upright)
  {}                           % Indent amount
  {\bfseries}                  % Theorem head font (bold)
  {.}                          % Punctuation after theorem head
  { }                          % Space after theorem head
  {}                           % Theorem head spec (empty = `normal`)

\theoremstyle{remarkstyle}

\crefname{figure}{Fig.}{Figs.}
\crefname{algorithm}{Algorithm}{Algorithms}

\title{A Physics-informed Multi-resolution Neural Operator}
\author{
        Sumanta Roy\thanks{These authors have contributed equally.}\\
        Dept. of Civil and Systems Engineering\\
	Johns Hopkins University\\
	Baltimore, MD, USA\\
	\texttt{sroy41@jhu.edu} \\
        \And
        Bahador Bahmani\textsuperscript{*}\thanks{Corresponding authors.}\\
	Depart. of Mechanical Engineering\\
	Northwestern University\\
	Evanston, Illinois\\
	\texttt{bahador.bahmani@northwestern.edu} \\
        \And
        Ioannis G. Kevrekidis\\
        Dept. of Chemical and Biomolecular Engineering\\
        Dept. of Applied Mathematics and Statistics\\
	Johns Hopkins University\\
	Baltimore, MD, USA\\
	\texttt{yannisk@jhu.edu} \\
	\And
        Michael D. Shields\textsuperscript{\textdagger}\\
        Dept. of Civil and Systems Engineering\\
	Johns Hopkins University\\
	Baltimore, MD, USA\\
	\texttt{michael.shields@jhu.edu} \\
}
\begin{document}

\maketitle

\begin{abstract}
    The predictive accuracy of operator learning frameworks depends on the quality and quantity of available training data (input-output function pairs), often requiring substantial amounts of high-fidelity data, which can be challenging to obtain in some real-world engineering applications. These datasets may be unevenly discretized from one realization to another, with the grid resolution varying across samples.
    In this study, we introduce a physics-informed operator learning approach by extending the Resolution Independent Neural Operator (RINO) framework~\cite{bahmani2025resolution} to a fully data-free setup, addressing both challenges simultaneously. Here, the arbitrarily (but sufficiently finely) discretized input functions are projected onto a \textit{latent} embedding space (i.e., a vector space of finite dimensions), using pre-trained basis functions. The operator associated with the underlying partial differential equations (PDEs) is then approximated by a simple multi-layer perceptron (MLP), which takes as input a latent code along with spatiotemporal coordinates to produce the solution in the \textit{physical} space. The PDEs are enforced via a finite difference solver in the physical space.
    The validation and performance of the proposed method are benchmarked on several numerical examples with multi-resolution data, where input functions are sampled at varying resolutions, including both coarse and fine discretizations.
\end{abstract}

\section{Introduction}\label{sec:introduction}

There has been significant research on leveraging neural network architectures as global parameterizations of solution fields for physical processes and partial differential equations (PDEs). Among these developments, the Physics-Informed Neural Network (PINN) framework~\cite{raissi2019physics} has emerged as a useful new paradigm, where deep networks are trained not only to fit observed data but also to satisfy the governing PDEs by minimizing its residuals during training~\cite{raissi2019physics,karniadakis2021physics,lagaris1998artificial}. While PINNs have shown great promise across diverse applications in science and engineering, for example in solid mechanics~\cite{roy2023deep,haghighat2021physics}, fluid mechanics~\cite{cai2021physics,wessels2020neural}, heat transfer~\cite{cai2021physicsheat,laubscher2021simulation}, subsurface mechanics~\cite{liu2024novel,roy2025adaptive}, and interface problems~\cite{sarma2024interface,roy2024adaptive} to name a few, they face a fundamental limitation: retraining is required whenever parameters such as boundary conditions, initial conditions, or material properties change. Addressing this challenge necessitates a shift from solving individual problems to learning operators that map input parameters to solutions of these parametric PDEs.

Learning such a solution operator involves learning the latent mapping between infinite dimensional spaces. Modern operator learning frameworks use deep neural networks (DNNs), and these type of operators are called neural operators~\cite{lu2021learning,azizzadenesheli2024neural,kovachki2023neural}. They have emerged as a powerful method for constructing general emulators for PDEs that describe physical systems. 
%Different types of deep neural operators have been developed over the years, all based on the universal approximation theorem originally proposed by Chen \& Chen~\cite{chen1995universal}.
The Deep Operator Network (DeepONet)~\cite{lu2021learning} is one of the first deep neural architectures introduced for operator learning, building on the seminal work of Chen and Chen~\cite{chen1995universal}.
Other popular architectures include the 
Fourier neural operator (FNO)~\cite{li2020fourier,li2023fourier},
wavelet neural operator (WNO)~\cite{tripura2022wavelet,tripura2023wavelet}, 
%reduced order DeepONets~\cite{lu2022comprehensive,meng2024general}, 
proper orthogonal decomposition DeepONets (POD-DeepONets)~\cite{lu2022comprehensive,meng2024general}, latent-DeepONets~\cite{kontolati2024learning,karumuri2025physics},
multi-scale operators~\cite{liu2021multiscale,lin2021seamless}, non-local kernel networks~\cite{you2022nonlocal}, and \textit{local} Neural Operators~\cite{fabiani2025enabling}. 
%
%These data-driven methods, while powerful, may require substantial training data to achieve high predictive accuracy and do not inherently respect the governing physical laws. 
These data‑driven methods, while powerful, require prior access to data generated from the solution operator—particularly when its governing equations are known \textit{a priori}—making them dependent on another solver, most often a classical one such as the finite element method (FEM). Generating the necessary training data typically entails extensive simulations with potentially costly solvers to ensure reliability at inference.
%To address these limitations, Wang et al.~\cite{wang2021learning} proposed physics-informed DeepONets, which integrate automatic differentiation to enforce governing PDEs as soft constraints in the loss functional, in a manner similar to standard PINN. This reduces reliance on large datasets by requiring only input functions and boundary/initial conditions. Physics-informed DeepONets have been successfully applied in various domains~\cite{wang2023long,li2023phase,ahmed2024physicsin,zhong2024physicsinformed}, but they still require large amounts of high-quality (high-fidelity) input data to train the high-accuracy and reliable deep learning surrogate model.
To remedy these challenges, Wang et al.~\cite{wang2021learning} proposed to augment DeepONet training with the governing equations of the PDEs, incorporating them as an additional regularization term in the loss function. This approach, which uses automatic differentiation to enforce the PDEs as soft constraints—similar to standard PINNs—effectively bypasses or reduces the reliance on another (classical) solver. Another class of data-free neural operators, based on an encoder–decoder U-Net architecture \cite{ronneberger2015u}, was introduced by Zhu et al.~\cite{zhu2019physics}, where the discrete form of the governing equations is enforced through finite differences.

However, a fundamental limitation of modern operator learning frameworks is that, although they are discretization- and resolution-independent in the output function space, they are not in the input function space. Typically, input data are supplied to the neural operator through measurements of the input function at fixed sensor locations, which remain consistent across all realizations. It is important to note that even in physics-informed variants, where paired input–output data are not required, the input functions must still be sampled consistently at sufficiently fine and identical locations across all samples. This requirement poses a significant bottleneck, as acquiring such data can be time-consuming or prohibitively expensive. In practice, data are often generated from multi-resolution modeling setups, where simulations are performed on grids of varying, possibly even adaptive discretizations. Experimental data acquisition presents similar challenges, as it is rarely feasible to obtain measurements from thousands of samples with perfectly aligned sensor locations~\cite{wang2024optimal,absi2019sensor}. For instance, experimental measurements may come from non-uniformly distributed sensors, and simulations may employ adaptive remeshing techniques. Several recent efforts have attempted to address these challenges. Wei et al.~\cite{li2024multi} proposed a multi-resolution FNO to process input functions at different resolutions, but their method was limited to data-driven training with aligned datasets, where input and output functions were sampled at the same locations. Similarly, You et al.~\cite{you2022nonlocal} developed a nonlocal kernel network (NKN), another data-driven framework that achieved resolution-independence in input functions, but again under the restriction that input data were aligned with output data. More recently, Abueidda et al.~\cite{abueidda2025time} introduced NCDE-DeepONet, a continuous-time operator framework that achieves resolution-independence in an unaligned manner, but only in the temporal domain, not in space. In this context, we believe that the Resolution-Independent Neural Operator (RINO) proposed by Bahmani et al.~\cite{bahmani2025resolution} provides a promising advance, accommodating input functions sampled at irregular spatial locations. Their approach employs a dictionary learning algorithm that projects each arbitrarily discretized input realization onto a fixed set of continuous and differentiable basis functions shared across all realizations. The resulting projection coefficients serve as embedding coordinates in a latent space of fixed dimensionality, enabling direct compatibility with operator learning architectures such as DeepONet, without requiring modifications to the original architecture.

Although the RINO framework demonstrated superior performance on input functions arbitrarily simulated on a mix of coarse and refined grids, the framework is completely data-driven, hence there is no guarantee that the learned solution operator follows the underlying PDE. In addition to that, RINO uses a linear decoder that approximates the set of target functions in a finite dimensional linear subspace. 
It has been shown that when target functional data lies on a low-dimensional nonlinear submanifold, nonlinear decoders are better suited for capturing the underlying manifold structure~\cite{gupta2022non,seidman2022nomad,chen2024nonlinear,kovachki2024operatorlearning}. In this work we aim to address the aforementioned challenges by extending RINO to a completely \textit{physics-informed, multi-resolution, non-linear decoder based} operator learning setup. Firstly, the multi-resolution data (in the form of arbitrarily discretized input functions) are projected onto a set of pre-trained basis functions shared across all realizations, where the projection coefficients serve as the embedding coordinates. We then concatenate these embeddings with the collocation points (locations where we want to evaluate the solution), and use a simple multi-layer perceptron (MLP) to approximate the solution at those collocation points. We then incorporate a finite difference solver within the optimization loop to enforce the governing PDEs in the physical space as additional loss terms. This framework is termed a \textbf{P}hysics-\textbf{I}nformed \textbf{R}esolution \textbf{I}ndependent \textbf{N}eural \textbf{O}perator (\textbf{PI-RINO}).

The rest of the paper is organized as follows. In Section~\ref{sec:operator_learning_between_function_spaces}, we present a brief overview of the concept of operator learning between function spaces. In Section~\ref{sec:pi_rino}, the proposed PI-RINO framework is introduced. The efficacy and accuracy of the proposed framework are demonstrated via solving four benchmark problems in Section~\ref{sec:num_examples}. We also compare the finite difference method used in this study with the traditional automatic differentiation method that conventional PINN based methods use in Section~\ref{sec:autodiff_compare}. Finally, we summarize our observations and provide concluding remarks in Section~\ref{sec:conclusion}.

\section{Operator learning between function spaces}\label{sec:operator_learning_between_function_spaces}

In this section, we present a brief overview of the operator learning framework for partial differential equations (PDEs). The input parameters (input functions) of the PDE include (but are not limited to) initial and boundary conditions, coefficients of the PDE, source/sink terms, and/or the shape of the physical domain, etc. Let us consider the PDE (and boundary condition) of the form
\begin{equation}
    \mathcal{N}(u,s;\Lambda)=0, \quad \mathcal{B}(u,s)=0, \quad \mathcal{I}(u,s)=0,
    \label{eq:PDE}
\end{equation}
where $\mathcal{N}$, $\mathcal{B}$, and $\mathcal{I}$ are the PDE, the boundary condition, and initial condition operators, respectively; $u \in \mathcal{U}$ and $s \in \mathcal{S}$ denote the vector-valued input and output functions, respectively; $\Lambda$ denotes the parameter vector defining the PDE coefficients; and $\mathcal{U}$ and $\mathcal{S}$ define Banach spaces as
\begin{align}
    \mathcal{U} = \{u: \mathcal{X} \rightarrow \mathcal{R}^{d_u} \}, \quad \mathcal{X} \subseteq \mathcal{R}^{d_x},\\
    \mathcal{S} = \{s: \mathcal{Y} \rightarrow \mathcal{R}^{d_s} \}, \quad \mathcal{Y} \subseteq \mathcal{R}^{d_y}.
\end{align}\label{eq:banach_spaces}

We assume that, for any $u \in \mathcal{U}$, there exist an unique solution $s \in \mathcal{S}$ to the PDE given by Eq.~\eqref{eq:PDE}, such that there is a ground-truth operator $\mathcal{G}$, that maps the input space $\mathcal{U}$ to the output space $\mathcal{S}$, such that $\mathcal{G}:\mathcal{U} \rightarrow \mathcal{S}$. The aim of the operator learning task is to approximate the operator using a parameterized surrogate model $\mathcal{G}^\text{approx}:\mathcal{U} \times \Theta \mapsto \mathcal{S}$, where $\Theta \in \mathcal{R}^{|\Theta|}$ denotes the parameter space of the surrogate model. %One common way to solve this problem is to approximate the operator using three maps: an encoder, approximate operator, and a decoder, as depicted in Figure~\ref{fig:schematic_encoder_approx_decoder}.  
One of the first operator learning frameworks was presented by Chen and Chen~\cite{chen1995universal} based on the universal approximation theorem of operators, which was later adapted to deep neural networks with the development of the DeepONet by Lu et al.~\cite{lu2021learning}. The original DeepONet
% approximates the operator $\mathcal{G}^\text{approx}$, with $\Theta$ being the parameters of the DeepONet. It 
consists of two DNNs, a branch net \textit{br} and a trunk net \textit{tr}. The trunk net learns a set of basis functions over the solution domain $\mathcal{S}$, and the branch net maps the input function to the coefficients for those bases. The inputs to the trunk net are the continuous coordinates $\boldsymbol{y} \in \mathcal{Y}$ in the output space, while the input to the branch net is a finite-dimensional vector representation $\boldsymbol{u}$ of the input function $u \in \mathcal{U}$. The final solution operator is approximated by the dot product of the branch and trunk networks as 
\begin{equation}
    s(\boldsymbol{y}) = \mathcal{G}(u)(\boldsymbol{y}) \approx \mathcal{G}^\text{DON}(\boldsymbol{u})(\boldsymbol{y}) = \sum^{P}_{k=1}br_k(\boldsymbol{u},\boldsymbol{\theta})~tr_k(\boldsymbol{y},\boldsymbol{\theta}),
\end{equation}\label{eq:deeponet_approximation}
where $\boldsymbol{\theta}$ represents the parameters of the DNNs.

\subsection{Multi-resolution dataset}\label{sec:operator_learning_dataset}

The efficacy and accuracy of the operator learning frameworks discussed in the previous section depend on the quality and the resolution of the available training data. The dataset for training neural operators generally consist of $N$ sets of paired discretized input-output functions $\{\boldsymbol{u}^{(i)},\boldsymbol{s}^{(i)} \}_{i= 1}^N$. These pairs are typically assumed to be independent and identically distributed (i.i.d.) samples from a probability measure, such that for each $u^{(i)}(\boldsymbol{x}) \in \mathcal{U}$, there exists a solution $s^{(i)}(\boldsymbol{y})\in\mathcal{S}$ to the PDE given by Eq.~\eqref{eq:PDE}, such that $s^{(i)}(\boldsymbol{y})=\mathcal{G}(u^{(i)}(\boldsymbol{x}))$. 
% The input function $u^{(i)}(\boldsymbol{x})$ is converted to a finite-dimensional vector valued function $\boldsymbol{\overline{u}}^{(i)}$, to make it compatible to act as an input to the deep neural networks. 
For all the aforementioned frameworks, the vector $\boldsymbol{u}^{(i)}$ is constructed by discretizing $u^{(i)}(\boldsymbol{x})$ at $M_\text{in}$ input sensor locations, and is stored as the point cloud $\mathcal{D}_\text{in}^{(i)}=\{ \boldsymbol{x}^{(i,j)},\boldsymbol{u}^{(i,j)}\}_{j=1}^{M_\text{in}}$, where $\boldsymbol{x}^{(i,j)}\in \mathcal{X}$. Similarly, the output function $s^{(i)}(\boldsymbol{y})$ are discretized at $M_\text{out}$ sensor locations, and is stored as a point cloud $\mathcal{D}_\text{out}^{(i)}=\{ \boldsymbol{y}^{(i,j)},\boldsymbol{s}^{(i,j)}\}_{j=1}^{M_\text{output}}$, where $\boldsymbol{y}^{(i,j)}\in \mathcal{Y}$. 

\begin{figure}[!hbt]
\begin{centering}
\includegraphics[width=0.8\textwidth]{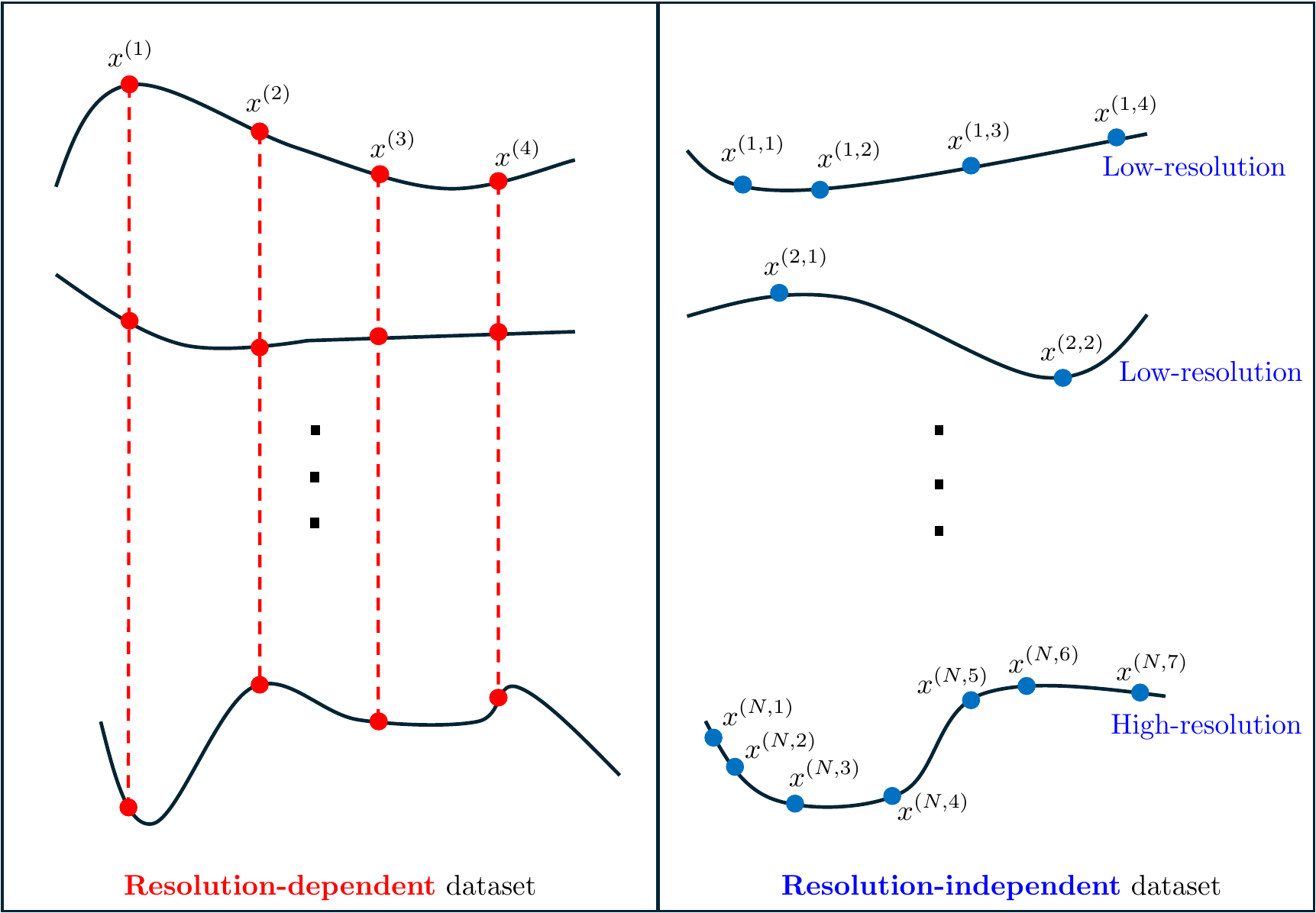}
\par\end{centering}
\caption{Illustration of a single-resolution dataset that typical neural operators use (left), compared to the multi-resolution dataset used in this study (right). The single-resolution dataset requires the input functions to be sampled at consistent location across all the realizations, while in the multi-resolution dataset does not have that requirement, as the input functions can be arbitrarily sampled across all the realizations.}\label{fig:multi_fidelity_dataset}
\end{figure}

The DeepONet framework and its variants require the input function $u(x)$ to be sufficiently richly discretized across $M_\text{in}$ locations which, critically, \textit{must remain consistent} across all $N$ realizations. This makes the framework resolution-dependent. However, in many practical scenarios, the data are not consistently discretized across all realizations.  That is, the data may have a \textit{multi-resolution} character in the sense that, for each realization $i$, the value of $u(\boldsymbol{x})$ is available at an arbitrary number of locations $M^{(i)}_\text{in}$ that is specific to that realization and such that $M^{(i)}_\text{in}\ne M^{(j)}_\text{in}$ for $i\ne j$  \cite{bahmani2025resolution}. We could have some realizations from low-resolution sources with smaller value of $M^{(i)}_\text{in}$ and some from high-resolution sources with larger values of $M^{(i)}_\text{in}$. This is illustrated in Figure~\ref{fig:multi_fidelity_dataset}. 

Note that this multi-resolution framework requires that all data be generated from the same set of PDEs with sufficient resolution to accurately characterize the input and solution. Moreover, we distinguish multi-resolution from multi-fidelity in the sense that multi-fidelity typically implies a loss of information from high-fidelity to low-fidelity.
While the proposed multi-resolution framework may be potentially fruitfully applicable to be applied in multi-fidelity frameworks in the future, we do not do so here. The next section introduces a neural-operator that handles this type of multi-resolution dataset. 

\section{Physics-informed resolution independent neural operator (PI-RINO)}\label{sec:pi_rino}

To overcome the limitations of many conventional operator learning methods in handling multi-resolution data, one may employ a function encoder that embeds each arbitrarily sampled realization into a finite-dimensional vector $\boldsymbol{\alpha} \in \mathbb{R}^Q$. This embedded vector is then used as input to the branch network of DeepONet without requiring any additional architectural changes. This approach is referred to as the Resolution Independent DeepONet (RI-DeepONet)~\cite{bahmani2025resolution} and can be written as follows:
%To handle multi-fidelity data, where the input functions $u(x)$ are available at arbitrary locations across realizations, we build a resolution-independent encoder that finds a vector representation of the multi-resolution data. This basically means finding a consistent-dimensional embedding $\boldsymbol{\alpha}\in \mathcal{R}^Q$ for each input function $u(x)$. We use the idea proposed by Bahmani et al.~\cite{bahmani2025resolution}, such that the modified neural operator (for example, the DeepONet) is written as 
\begin{equation}
    s(\boldsymbol{y}) = \mathcal{G}(u)(\boldsymbol{y}) \approx \mathcal{G}^\text{RI-DON}(\boldsymbol{u})(\boldsymbol{y})=\sum^{P}_{k=1}br_k(\boldsymbol{\alpha}(\boldsymbol{u}),\boldsymbol{\theta})~tr_k(\boldsymbol{y},\boldsymbol{\theta}).
\end{equation}\label{eq:rino_deeponet_approximation}
%
%The above method is termed as resolution-independent DeepONet (RI-DeepONet) where the output functions are approximated on a linear subspace. 
%Moreover, although the universal approximation theorem of operators guarantee that there exists a linear subspace of a large enough dimension which approximates the target functions to any prescribed accuracy~\cite{kovachki2021universal,lanthaler2022error}, more often than not there are scenarios where the target functional data concentrates on sub-manifold~\cite{cohen2015approximation}. 
%
%Seidman et al.~\cite{seidman2022nomad} refers to this  phenomenon as the \textit{Operator Learning Manifold Hypothesis}.
The linear assumption of DeepOnet architectures (with respect to the solution space basis functions $tr_k$) can potentially be inefficient for problems where the solution lies on a lower-dimensional manifold rather than a linear subspace. While it is possible to represent such data using linear theory, doing so may require an unnecessarily large number of basis functions \citep{bahmani2025resolution,seidman2022nomad,cohen2015approximation}.
%
%Thus for certain classes of parametric PDEs, the solutions lie in low dimensional nonlinear manifolds, and thus linear representations can be very inefficient in representing these spaces as the dimensions can become  very large and thus inefficient in capturing the true low dimensional structure of the data. 
%Hence, they suggest using a non-linear decoder, termed \textit{NOMAD}, to approximate the solutions in the solution space, as given by:
Hence, in this work, to maintain generality, we approximate the operator using a single nonlinear function $s_{\boldsymbol{\theta}}$, parameterized by a deep neural network with parameters $\boldsymbol{\theta}$, as follows:
\begin{equation}
    s(\boldsymbol{y}) = \mathcal{G}(u)(\boldsymbol{y}) \approx s_{\boldsymbol{\theta}}(\boldsymbol{\alpha}(\boldsymbol{u}),\boldsymbol{y}).
    \label{eq:nomad_decoder}
\end{equation}
Such operator approximations have been used elsewhere \cite{seidman2022nomad,karumuri2020simulator}. We then aim to ensure that the approximate solution $s_{\boldsymbol{\theta}}(\boldsymbol{\alpha}(\boldsymbol{u}),\boldsymbol{y})$ satisfies the PDE and its boundary conditions for all $\boldsymbol{y}$, which leads to the proposed \textbf{P}hysics-\textbf{I}nformed \textbf{R}esolution \textbf{I}ndependent \textbf{N}eural \textbf{O}perator (\textbf{PI-RINO}). 

\begin{figure}[!ht]
\begin{centering}
\includegraphics[width=1.0\textwidth]{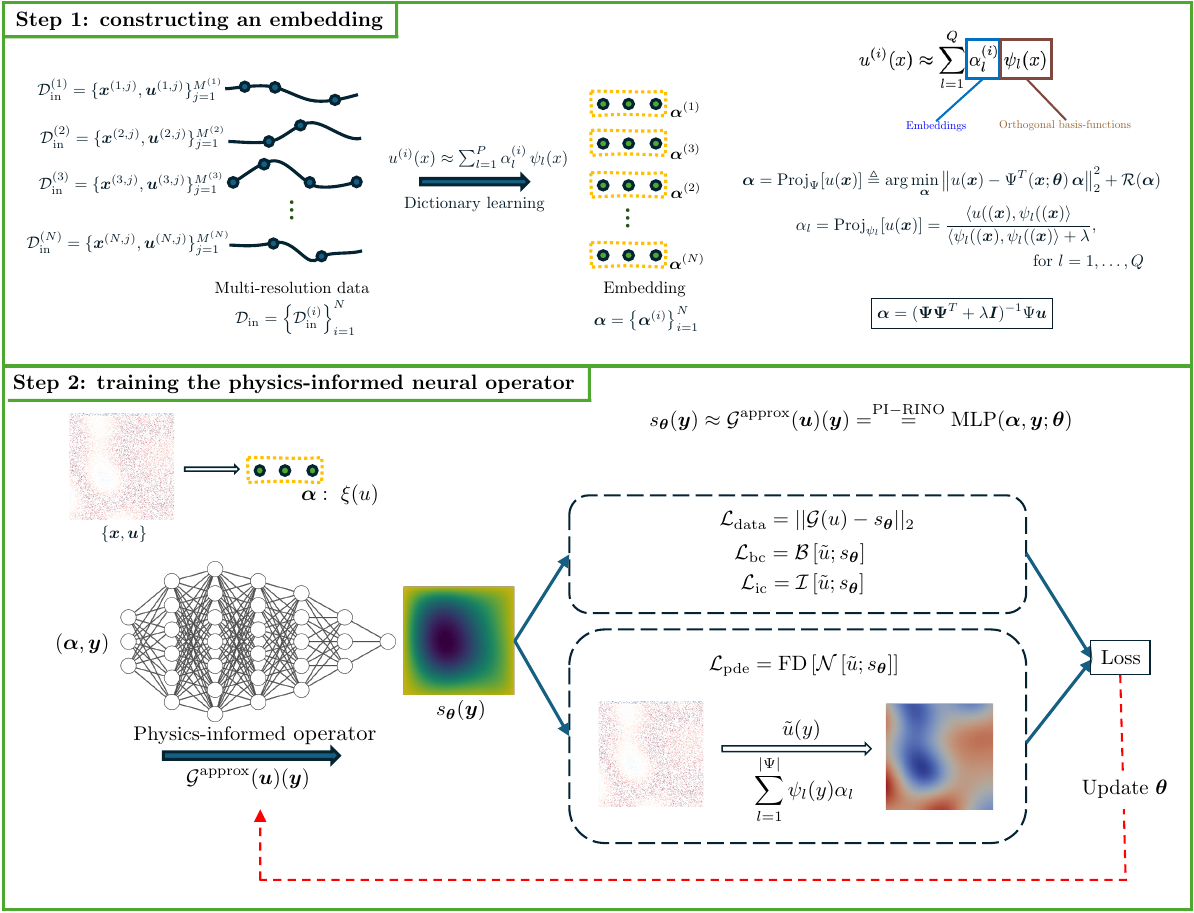}
\par\end{centering}
\caption{Schematic of the architecture of Physics-informed Resolution Independent Neural Operator (PI-RINO) for approximating the operator $\mathcal{G}:u \mapsto s$, following the PDE: $\mathcal{N}(u,s)=0$.}\label{fig:pi_rino_schematic}
\end{figure}

The PI-RINO framework consists of two general steps:
\begin{enumerate}
 \item \textbf{Learning a consistent embedding basis:} In this step, the arbitrarily discretized input function data are projected onto a set of continuous and orthogonal basis functions (e.g., orthogonal polynomials), with their projection coefficients serving as the embedding coordinates. Although many types of embeddings have been proposed in the literature (e.g., \cite{adcock2019frames,klus2015numerical,ingebrand2025basis}), in this work we select a dictionary learning algorithm due to its accuracy and implementational flexibility. Specifically, the algorithm identifies a set of basis functions parameterized by implicit neural representations (INRs), following the methods developed in~\cite{bahmani2025resolution}, which are used to approximate arbitrary input functions defined on point cloud data. The resulting embedding coordinates, $\boldsymbol{\alpha}(\boldsymbol{u})$, serve as the latent-space input to the neural operator in Eq.~\ref{eq:nomad_decoder}.

    \item \textbf{Training the physics-informed neural operator:} 
    % The embedding represents the input function in a latent-space. 
    The embedding coordinates, $\boldsymbol{\alpha}(\boldsymbol{u})$, are concatenated with the location coordinates $\boldsymbol{y}$ at which the output function $s(\boldsymbol{y})$ is to be evaluated. Together, these serve as the inputs to a deep neural network (multi-layer perception, MLP, in this case), which approximates the action of the PDE operator and produces the output in the physical domain, in accordance with Eq.~\eqref{eq:nomad_decoder}. The input functions are \textit{reconstructed in the physical-space} using the pre-trained basis-functions and coefficients and the governing PDE is enforced by constructing a physics-based loss functional, which acts as a guiding constraint in the training process. A finite-difference solver is used in the physical space to compute the necessary gradient operations of the PDE at specified collocation points.
\end{enumerate}
Figure~\ref{fig:pi_rino_schematic} illustrates the overall PI-RINO framework used to approximate the solution $s_{\boldsymbol{\theta}}(\boldsymbol{y})$ and the following subsections describe its components. The approach is detailed in Algorithm~\ref{alg:algorithm}.

\begin{algorithm}[!ht]
\caption{Physics-informed Resolution Independent Neural Operator (PI-RINO)}\label{alg:pi_rino}
\begin{algorithmic}[1]
\State \textbf{Data:} $\boldsymbol{\mathcal{D}}_\text{in}=\left \{ \boldsymbol{\mathcal{D}}^{(i)}_\text{in} \right \}_{i=1}^{N}$, ~$\boldsymbol{\mathcal{D}}^{(i)}_\text{in}= \left\{ \boldsymbol{x}^{(i,j)},\boldsymbol{u}^{(i,j)} \right\}_{j=1}^{M^{(i)}}$ \Comment{$N$ samples of input data, each sampled at $M^{(i)}$ \hfill locations}
\vspace{1em}

\State \textbf{PDE:} $\mathcal{N}(u,s)=0$ \Comment{PDE operator}
\vspace{1em}

\State \textbf{Step 1: Constructing an embedding:}
\State \(\boldsymbol{\Psi} = \{1\}\), \(\eta \in \mathbb{R}^{+}\), \(0 < \text{Tol} \ll 1\) \Comment{Initializing the dictionary, learning rate, and reconstruction error tolerance}
\While{\text{Recon. Err.} \(\geq\) \text{Tol}}
    \State Add a randomly initialized neural basis \(\psi^{\text{new}}(\boldsymbol{x}; \boldsymbol{\theta}^{\text{new}})\) to \(\boldsymbol{\Psi}(\boldsymbol{x}; \boldsymbol{\theta}, \boldsymbol{\theta}^{\text{new}})\)
    \For{\(e \gets 1\) to \(N_{\text{epoch}}\)} \Comment{Loop over epochs}
        \For{\(i \gets 1\) to \(N\)} \Comment{Loop over samples}
            \State \(\displaystyle
            \boldsymbol{\alpha}^{(i)} = \text{Proj}_{\Psi}[u^{(i)}(\boldsymbol{x})] \approx \arg\min_{\boldsymbol{\alpha}} 
            \mathbb{E}_{(\boldsymbol{x}, \boldsymbol{u}) \sim \mathcal{D}_{u}^{(i)}} \left[
            \left\| \boldsymbol{u} - \boldsymbol{\Psi}^T(\boldsymbol{x}; \boldsymbol{\theta}, \boldsymbol{\theta}^{\text{new}}) \boldsymbol{\alpha} \right\|_2^2
            \right] + \mathcal{R}(\boldsymbol{\alpha})\)
        \EndFor
        \State \(\displaystyle
        \boldsymbol{\theta}^{\text{new}} \gets \boldsymbol{\theta}^{\text{new}} - \eta \nabla_{\boldsymbol{\theta}^{\text{new}}} 
        \mathbb{E}_{i} \left[
        \mathbb{E}_{(\boldsymbol{x}, {u}) \sim \mathcal{D}_{u}^{(i)}} \left[
        \left\| \boldsymbol{u}(\boldsymbol{x}) - \boldsymbol{\Psi}^T(\boldsymbol{x}; \boldsymbol{\theta}, \boldsymbol{\theta}^{\text{new}}) \boldsymbol{\alpha}^{(i)} \right\|_2^2
        \right]
        \right]\) \Comment{Update the basis function parameters}
    \EndFor
    \State Calculate reconstruction error.
\EndWhile
\vspace{1em}

\State \textbf{Step 2: Training the physics-informed neural operator:}
\State \(\eta \in \mathbb{R}^{+}\) \Comment{Initializing the optimizer learning rate}
\For{\(e \gets 1\) to \(N_{\text{epoch}}\)} \Comment{Loop over epochs}
        \For{\(i \gets 1\) to \(N\)} \Comment{Loop over realizations (can be batched)}
            \State Concatenate $\boldsymbol{y}$ and $\boldsymbol{\alpha}^{(i)}$
            \State $s^{\text{pred},(i)}_{\boldsymbol{\theta}}(\boldsymbol{y}) = \phi(\boldsymbol{y}) ~ \text{MLP}(\boldsymbol{\alpha}^{(i)},\boldsymbol{y};\boldsymbol{\theta}) + P(\boldsymbol{y})$\Comment{Output prediction (hard-constrained)}
            \State $\tilde{u}^{(i)}(\boldsymbol{y}) = \sum_{l=1}^{|\Psi|}\psi_l(\boldsymbol{y}) \alpha_l^{(i)}$\Comment{Full-field input reconstruction}
            \State $\mathcal{L}_{\text{data}}^{(i)}=\left\|\mathcal{G}(u^{(i)})(\boldsymbol{y})-s^{\text{pred},(i)}_{\boldsymbol{\theta}}(\boldsymbol{y})\right\|^2_{2}$ \Comment{Data-loss (if available)}
            \For{$(p,q) \sim \text{domain}$} 
                \State $\mathcal{L}_{\text{pde}: ~\text{[p,q]}}^{(i)}=\text{FD}\left[ \mathcal{N}\left( \tilde{u}^{(i)}_{[p,q]}, s^{(i)}_{\boldsymbol{\theta}:~[p,q]} \right)  \right]$\Comment{FD loss on $[p,q]$-th collocation point}
            \EndFor
            \State $\mathcal{L}_{\text{pde}}^{(i)}=\sum_{p,q}\mathcal{L}_{\text{pde}: ~\text{[p,q]}}^{(i)}$\Comment{Physics-loss (sum of FD losses)}
            \State \(\displaystyle
        \boldsymbol{\theta} \gets \boldsymbol{\theta} - \eta \nabla_{\boldsymbol{\theta}} 
        \mathbb{E}_{i}
         \left[ \mathcal{L}_{\text{pde}}^{(i)} + \mathcal{L}_{\text{data}}^{(i)}
        \right]\) \Comment{Update neural operator parameters}
        \EndFor
    \EndFor
\end{algorithmic}\label{alg:algorithm}
\end{algorithm}

%Note: We hypothesize that this framework approximates the solution with a fully nonlinear decoder (MLP) to represent target functions in low dimensional solution manifolds. Thus we use just one deep neural network for the process, instead of three as used by Seidman et al.~\cite{seidman2022nomad}.

\subsection{Step 1: Constructing an embedding (\textit{function encoder})}\label{sec:pi_rino_dictionary}
The first step to handle the multi-resolution data is to identify a set of basis functions, which are used to approximate arbitrary input functions defined as point cloud data. Numerous options are available such as Fourier basis or polynomials basis functions. We use the dictionary learning algorithm introduced in \cite{bahmani2025resolution} because it is data-driven and offers great flexibility. The $i$-th input function realization is represented by a linear combination of basis functions, as follows:
\begin{equation}
    u^{(i)}(\boldsymbol{x}) \approx \tilde{u}^{(i)}(\boldsymbol{x}) = \sum_{l=1}^P \alpha_l^{(i)} \, \psi_l(\boldsymbol{x}) = \boldsymbol{\Psi}^T(\boldsymbol{x})\, \boldsymbol{\alpha}^{(i)},
\end{equation}
where $\alpha_l^{(i)}$ is the coefficient corresponding to the $l$-th basis function $\psi_l(\boldsymbol{x})$. The coefficients $\boldsymbol{\alpha}^{(i)}$ serve as the embeddings that capture the \textit{latent} information of the input function $u^{(i)}(\boldsymbol{x})$, which will later serve as inputs to the neural operator.
The data-driven basis functions, $\psi_l(\boldsymbol{x})$ are determined adaptively, each time based on the reconstruction residual obtained from the projection onto the subspace spanned by the dictionary's current state up to that iteration. In this way, each newly discovered basis function is designed to be weakly orthogonal to the existing dictionary $\{\psi_l\}_{l}$. These basis functions are parameterized by neural networks with sinusoidal activation functions following SIREN \cite{sitzmann2020implicit}. An effective dictionary learning algorithm has been introduced in \cite{bahmani2025resolution} to learn these basis functions and coefficients simultaneously, and further details can be found in Appendix~\ref{app:dictionary_learning}. Once the basis functions are learned, any sampled input function $\{\boldsymbol{x}^{(i)}, \boldsymbol{u}^{(i)}\}$ can be encoded via these basis functions by:
\begin{equation}
\boldsymbol{\alpha}^{(i)} =
\text{Proj}_{\boldsymbol{\Psi}}[u^{(i)}(\boldsymbol{x})] = (\boldsymbol{\Psi}(\boldsymbol{x}) \boldsymbol{\Psi}(\boldsymbol{x})^T + \lambda \boldsymbol{I})^{-1} \boldsymbol{\Psi}(\boldsymbol{x}) \boldsymbol{u}^{(i)},
\end{equation}
where $\lambda$ is an algorithmic parameter used to promote sparsity and help avoid ill-conditioned optimization during training.

\subsection{Step 2: Training the physics-informed neural operator (\textit{approximator})}\label{sec:pi_rino_no_training}

After obtaining the embedding $\boldsymbol{\alpha}^{(i)} \in \mathbb{R}^Q$ for each input function $u^{(i)}(\boldsymbol{x})$, we use it as input to the neural operator in Eq.~\ref{eq:nomad_decoder}.
%Therefore, following \cite{lanthaler2022error,seidman2022nomad} (see Eq.~\eqref{eq:nomad_decoder}), we aim to learn a nonlinear mapping that approximates the output function $s(x)$ by taking the embedding $\boldsymbol{\alpha}$ and the output coordinates $x$ as input. The authors in \cite{seidman2022nomad} retain the branch and trunk networks from the traditional DeepONet framework but replace the dot product with an additional MLP, allowing the output to lie in a nonlinear submanifold. Inspired by their approach, we adopt a simpler method: we use a single MLP (parameterized by $\boldsymbol{\theta}$) that takes the embedding $\boldsymbol{\alpha}$ and the output coordinates $x$ as input, and predicts the solution $s(x)$ in physical space:
%\begin{equation}
   % s_{\boldsymbol{\theta}}(x) \approx \mathcal{G}^\text{approx}(u)(x) \overset{\mathrm{PI\text{-}RINO}}{=} \text{MLP}(\boldsymbol{\alpha},x;\boldsymbol{\theta}),
%    \label{eq:pi-rino_equation}
%\end{equation}
%
%
Although the neural network $s_{\boldsymbol{\theta}}$ can be any appropriately chosen architecture, in this work we opt for a simple MLP to take advantage of its fully continuous nature, and thus its resolution independence, in the output space.
%Note that this parametrization is inherently resolution-independent (i.e., continuous) in the output space.
%where $\boldsymbol{\theta}$ denotes the weights and biases of the neural network. 
% This MLP combines the roles of both the approximator and the nonlinear decoder as shown in Fig.~\ref{fig:schematic_encoder_approx_decoder}. 
The parameters $\boldsymbol{\theta}$, which include all the weights and biases of the MLP, are optimized using the physics-informed loss function:
\begin{equation}
    \mathcal{L}(\boldsymbol{\theta}) = \mathcal{L}_{\text{pde}} + \mathcal{L}_{\text{bc}} + \mathcal{L}_{\text{ic}} = \frac{1}{Nm} \sum_{i=1}^{N} \sum_{j=1}^{m} \left |\mathcal{N}\left( \tilde{u}^{(i)}(\boldsymbol{y}_j), s_{\boldsymbol{\theta}}(\boldsymbol{y}_j) \right) \right |^2 
    +
    \frac{1}{Nb} \sum_{i=1}^{N} \sum_{j=1}^{b} \left |\mathcal{B}\left( \tilde{u}^{(i)}(\boldsymbol{y}_j), s_{\boldsymbol{\theta}}(\boldsymbol{y}_j) \right) \right |^2
    +
    \frac{1}{Nt} \sum_{i=1}^{N} \sum_{j=1}^{t} \left |\mathcal{I}\left( \tilde{u}^{(i)}(\boldsymbol{y}_j), s_{\boldsymbol{\theta}}(\boldsymbol{y}_j) \right) \right |^2,
    \label{eq:pi-rino_loss_function}
\end{equation}
where $\mathcal{N}$, $\mathcal{B}$, and $\mathcal{I}$ denote the PDE, boundary, and initial condition operators, respectively. Here, $N$ is the number of input function realizations, $m$ is the number of collocation points used to evaluate the physics loss (first term),  $b$ and $t$ is the number of points used for the boundary and initial loss (second and third term).

% To overcome the potential challenge of satisfying the governing PDE in the latent space, we advocate enforcing such constraints directly in the \textit{physical space}. 
Because these loss terms are often not well-defined in the latent space, we must enforce the associated constraints in the physical space.
Meanwhile, the input functions are encoded into a consistent latent space to ensure resolution independence (from the input function space perspective) and potentially reduce the dimensionality of the input function space. It is important to note that when computing the PDE residuals, we may need to evaluate the input functions at arbitrary spatiotemporal points (in the output function space). This is straightforward in the proposed approach, as the input function can be reconstructed at any point $\boldsymbol{y}$ given its embedding coordinates as:
%It is important to note that the output $s(y)$ is predicted in the \textit{physical space}, whereas the input functions $u(x)$ are only available at scattered, potentially arbitrary locations (the whole point of multi-fidelity learning). 
%
%Thus, to enforce the physics loss in physical space, we reconstruct the full-field version $\tilde{u}^{(i)}(x)$ of each input function $u^{(i)}(x)$ using the learned basis functions and the corresponding embeddings:
% Hence, when needed, one can reconstruct the full-field version $ u^{\text{reconstruct},(i)}(\boldsymbol{y})$ of each input function $u^{(i)}(\boldsymbol{y})$ using the learned basis functions and the corresponding embeddings as:
\begin{equation}
    \tilde{u}^{(i)}(\boldsymbol{y}) = \sum_{l=1}^{|\Psi|}\psi_l(\boldsymbol{y}) \alpha_l^{(i)},
    \label{eq:full-field-reconstruciton}
\end{equation}
which ensures a continuous representation of the input function that allows $u(\boldsymbol{y})$ to be evaluated at any ouput fuction point in the loss function, Eq.~\eqref{eq:pi-rino_loss_function}. 

\subsubsection{Finite difference method for physics loss}\label{sec:pi_rino_no_training_fd_loss}

Most traditional physics-informed machine learning methods use automatic differentiation~\cite{rall1981automatic} to compute derivatives in the output space. However, this can be computationally expensive, especially when higher-order derivatives are required~\cite{baydin2018automatic}, as it involves backpropagation through the entire network. To improve computational efficiency, we instead employ a finite difference (FD) convolutional stencil in physical space to approximate the necessary gradients \citep{gonzalez1998identification,raissi2018hidden,zhu2019physics}.

\begin{figure}[!ht]
\begin{centering}
\includegraphics[width=1.0\textwidth]{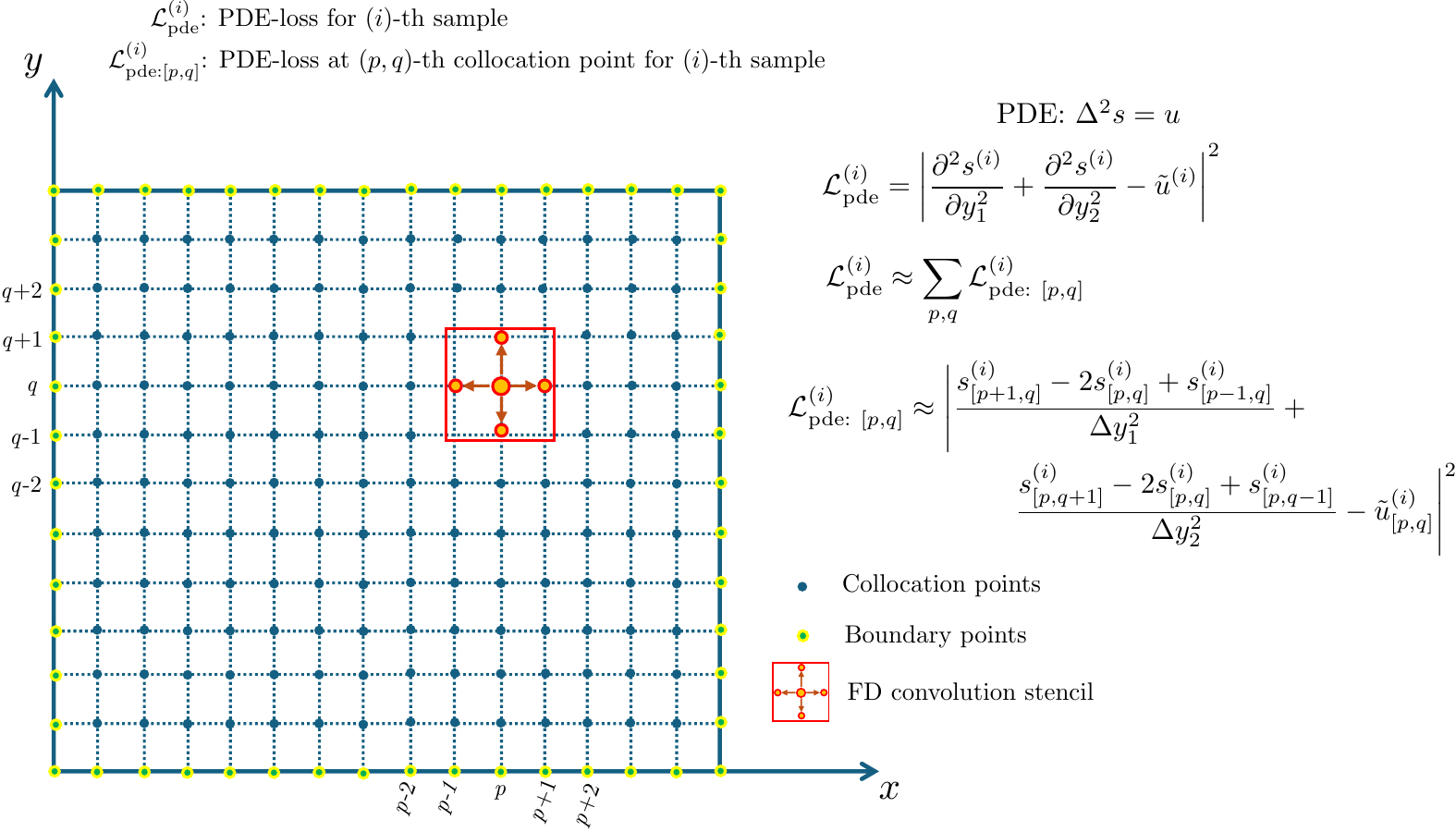}
\par\end{centering}
\caption{The physics loss enforced by convoluting a finite difference stencil across the entire grid of structured collocation points in the domain. This figure demonstrates the FD loss term computed for the $(i)$-th sample at the $(p,q)$-th collocation point for the representative PDE: $\Delta^2 s=u$. The final physics loss for the $(i)$-th sample is computed by taking the sum of all the residuals/losses across all the collocation points.}\label{fig:fd_convolution_schematic}
\end{figure}

This approach requires sampling collocation points on a structured grid, as illustrated in Figure~\ref{fig:fd_convolution_schematic}, in accordance with the requirements of the finite difference method. This structured sampling is feasible because both the neural operator output (Eq.~\eqref{eq:nomad_decoder}) and the reconstructed full-field input functions (Eq.~\eqref{eq:full-field-reconstruciton}) are continuous in the physical space $\boldsymbol{y}$, allowing us to evaluate them at any desired collocation point. For each sample realization $(i)$, we approximate 2-dimensional spatial derivatives using Taylor series expansions as follows:
\begin{equation}
\begin{split}
\left[\frac{\partial s^{(i)}}{\partial y_1} \right]_{p,q} &= \frac{s^{(i)}_{[p+1,q]} - s^{(i)}_{[p-1,q]}}{2\Delta y_1}, 
\quad 
\left[\frac{\partial s^{(i)}}{\partial y_2} \right]_{p,q} = \frac{s^{(i)}_{[p,q+1]} - s^{(i)}_{[p,q-1]}}{2\Delta y_2}, \\
\left[\frac{\partial^2 s^{(i)}}{\partial y_1^2} \right]_{p,q} &= \frac{s^{(i)}_{[p+1,q]} - 2s^{(i)}_{[p,q]} + s^{(i)}_{[p-1,q]}}{\Delta y_1^2}, 
\quad 
\left[\frac{\partial^2 s^{(i)}}{\partial y_2^2} \right]_{p,q} = \frac{s^{(i)}_{[p,q+1]} - 2s^{(i)}_{[p,q]} + s^{(i)}_{[p,q-1]}}{\Delta y_2^2},
\end{split}
\label{eq:taylor_approximations}
\end{equation}
where \( [p, q] \) refer to the indices of points in the domain (see Fig.~\ref{fig:fd_convolution_schematic} for reference), and $\Delta y_1$ and $\Delta y_2$ are grid sizes in the horizontal and vertical dimensions, respectively.
%While the expressions above apply to the reconstructed input function $u^{(i)}(x)$, analogous approximations are used for the output $s^{(i)}(x)$ as well.

These derivative approximations are used in the physics-informed loss computation. 
The FD stencil is applied (convolved) across the entire structured grid, and the local 
physics loss at each point $\boldsymbol{y}_{[p,q]}$ is given by:
\[
\mathcal{L}^{(i)}_{\text{pde}:[p,q]} 
= \text{FD} \Big[ \mathcal{N}\!\left( \tilde{u}^{(i)}_{[p,q]},\; 
s^{(i)}_{\boldsymbol{\theta}:[p,q]} \right) \Big].
\]
The total physics loss is then computed by summing these local losses across the domain. 
Figure~\ref{fig:fd_convolution_schematic} illustrates this stencil-based computation for 
a representative PDE: $\Delta^2 s = u$. 
% The FD loss is computed for the $[p,q]$-th 
% collocation point of the $i$-th input sample as
% \begin{equation}
%     \mathcal{L}_{\text{physics}:[p,q]}^{(i)} \approx 
%     \Bigg(
%     \frac{s^{(i)}_{[p+1,q]} - 2s^{(i)}_{[p,q]} + s^{(i)}_{[p-1,q]}}{\Delta y_1^2} 
%     + \frac{s^{(i)}_{[p,q+1]} - 2s^{(i)}_{[p,q]} + s^{(i)}_{[p,q-1]}}{\Delta y_2^2} 
%     - u^{\text{reconstruct}, (i)}_{[p,q]}
%     \Bigg)^{2}.
%     \label{eq:fd_local_loss}
% \end{equation}
% The FD loss for the $i$-th sample is obtained by convoluting this across all collocation points:
% \begin{equation}
%     \mathcal{L}_{\text{physics}}^{(i)} \approx 
%     \sum_{p,q} \mathcal{L}_{\text{physics}:[p,q]}^{(i)}.
%     \label{eq:fd_total_loss}
% \end{equation}
\paragraph{Remark} Although the FD discretizations in
% Eq.~\eqref{eq:fd_local_loss}--\eqref{eq:fd_total_loss} 
Figure~\ref{fig:fd_convolution_schematic} resemble standard textbook formulas, their role in physics-informed loss computation introduces a key distinction compared to autodiff-based methods. In conventional PINNs, collocation points are unstructured and independent, which allows batching to be performed over the joint pool of $(\text{\#samples} \times \text{\#collocation points})$. In contrast, the FD-based residual at each point $[p,q]$ depends on its neighbors through the stencil, thereby coupling all collocation points of a given sample. Consequently, the PDE loss for the $i$-th sample must be computed over the \emph{entire structured grid} associated with that sample, and batching is restricted to occur \emph{only across samples}, not across individual collocation points. This distinction, while subtle, has important implications for training efficiency and memory usage in FD-based physics-informed learning.

As we will demonstrate later in Section~\ref{sec:autodiff_compare}, PI-RINO using FD significantly outperforms its Autodiff-based counterpart in terms of convergence speed, while achieving comparable levels of approximation accuracy in both the predicted outputs and the physics residuals. Note, however, that other factors influencing the computational gain include the network size (which has the greatest impact on autodiff) as well as the stencil size, and one can imagine configurations in which this gain could be diminished. For demonstration purposes, we apply the basic FD scheme presented here. Additional details on the specific FD stencils used in the numerical experiments of this study are provided in Appendix~\ref{app:fd_loss}.

%As we will demonstrate later in Section~\ref{sec:autodiff_compare}, PI-RINO using the FD significantly outperforms its Autodiff-based counterpart in terms of convergence speed, while achieving comparable levels of approximation accuracy in both the predicted outputs and the physics residuals. Note that for advanced problems, more sophisticated FD schemes using higher-order stencils and/or unequal mesh spacing can be employed within the proposed method. However, higher-order numerical schemes come at greater cost and their advantages over automatic differentiation may diminish or vanish entirely. We apply the basic FD scheme presented here for demonstration. Additional details on the specific FD stencils used in the numerical experiments of this study are provided in Appendix~\ref{app:fd_loss}.

\subsubsection{Hard-constraining boundary and/or initial conditions}\label{sec:pi_rino_no_training_hard_constraint}

The loss function presented in Eq.~\eqref{eq:pi-rino_loss_function} adopts a soft-constraint formulation, where boundary conditions (and initial conditions, if applicable) are enforced through the loss function. This means that the optimizer attempts to minimize the deviation from the true boundary or initial conditions, but there is no strict guarantee that the solution will exactly satisfy them. While this approach is widely used in the PINNs community, it may result in suboptimal approximations, as noted in~\cite{zubov2021neuralpde,alkhadhr2023wave}. Typically, the more objectives included in the loss function, the more difficult the training becomes, as some objectives may conflict with one another \cite{bahmani2021training}. Therefore, designing the architecture to satisfy as many constraints as possible can facilitate training and potentially improve generalizability.

%To address this issue, we introduce a modification that enforces the boundary or initial conditions in a \textit{hard} manner, ensuring that the solution exactly satisfies these constraints. 
To address this issue, one can reparametrize the neural operator $s_{\boldsymbol{\theta}}$ to enforce the boundary or initial conditions in a \textit{hard} manner, ensuring that the solution exactly satisfies these constraints, as follows
%This is achieved by reformulating the solution approximation as:
\begin{equation}
s_{\boldsymbol{\theta}}(\boldsymbol{y}) = \phi(\boldsymbol{y}) ~ \text{MLP}(\boldsymbol{\alpha},\boldsymbol{y};\boldsymbol{\theta}) + P(\boldsymbol{y}), \quad \boldsymbol{y}\in \Omega,
\label{eq:hard_constraint}
\end{equation}
where $P(\boldsymbol{y})$ is designed to satisfy the prescribed boundary (or initial) values on the domain boundary $\partial \Omega$, and $\phi$ is a distance function that vanishes on the boundary (or at the initial time) ~\cite{sukumar2022exact}. 

In all experiments conducted in this study, we enforce Dirichlet boundary conditions (or initial conditions) in a hard manner using this approach. The function $\phi$ depends on the spatial coordinate $\boldsymbol{y}$ (or on time $t$ for enforcing initial conditions). In cases where both spatial and temporal constraints are enforced simultaneously, $\phi$ becomes a function of both $\boldsymbol{y}$ and $t$. For readers interested in hard-enforcing both essential and mixed boundary conditions, we refer to the work by Sukumar et al.~\cite{sukumar2022exact}, which presents a methodology for constructing distance functions to a domain’s boundary using R-functions and the theory of mean value potential fields.

\section{Numerical Examples}
\label{sec:num_examples}

In this section, we demonstrate the effectiveness of the proposed framework on various benchmark examples in 1D and 2D. In all of the examples (unless otherwise stated), the input functions $u(\boldsymbol{x})$ are initially generated using zero-mean Gaussian random fields (GRFs) as
$$
u \sim \mathcal{GP}(0,k_l(x_1,x_2)),
$$
with an exponential quadratic kernel $k_l(x_1,x_2)=\exp\left(-||x_1-x_2||^2/2l^2\right)$, with a length scale parameter $l>0$. 
%The parameter $l$ will be used to control the complexity and spread of the input functions $\mathbf{u}$. 
In order to mimic a multi-resolution setup, we only use sparse representations of the input functions $u$, by assuming that the data is only available to us using sensors placed sparsely (and randomly) across the realizations. The locations and numbers of these sensors vary from sample to sample. If the original input function is discretized at $M$ points for all realizations, we randomly subsample by selecting $M_\text{rand}$ points without replacement, where $M_\text{rand}$ is independently drawn from the uniform distribution, $M_\text{rand} \sim \text{Uniform}[M_\text{min},M_\text{max}]$, where  $0< M_\text{min}<M_\text{max}\leq M$. Except for the last example, all cases use the same input function data as used in \cite{bahmani2025resolution}.

%For all the examples, we have employed various activation functions, like the hyperbolic tangent (tanh), Rectified Linear Unit (ReLU),  Mish, and have compared the performance of them (in terms of accuracy on unseen test data and computational cost/training-times). 
The parameters (weights and biases) of the neural operator are initialized using the Xavier initialization scheme~\cite{glorot2010understanding}. The networks are trained using the mini-batch gradient descent using the Adam optimizer~\cite{kingma2014adam} with default settings. The hyperparameters used for training the operators are detailed in Appendix~\ref{app:hyperparameters}. 

\subsection{Antiderivative}\label{sec:num_examples_anti_derrivatve}
%To demonstrate the capability of the PI-RINO framework, let us take the classical example of the anti-derivative operator:
In the first numerical example, the antiderivative problem is chosen as a verification case, which has been extensively studied in the literature \citep{lu2021learning, wang2021learning}:
\begin{equation}
    \frac{ds}{dy}=u(y); \quad y \in [0,1],
\end{equation}\label{eq:antiderrivative_a}
with the initial condition $s(0)=0$, such that our goal is to learn the anti-derivative operator
\begin{equation}
G: u(x) \rightarrow s(y) = s(0) + \int_{0}^{y}u(t)dt, \quad y\in[0,1],
\end{equation}\label{eq:antiderrivative_b} where $u(x)$ is modeled using a GRF with length scale $l = 0.2$. In this study, we use the same dataset as the open-source data provided in \cite{lu2021learning}.
The original dataset ($\boldsymbol{u}(\boldsymbol{x}),\boldsymbol{s}(\boldsymbol{y})$ pairs) contains 150  samples (realizations) for the training set and 1000 samples (realizations) for the test set. The functions are discretized at 101 equally spaced sensor points. As described above, we generate the multi-resolution setup by sub-sampling from the set such that each training data function has a random discretization of between 
% 10 and 60 points. This is done by generating point-clouds by setting 
$M_\text{min}=10$ and  $M_\text{max}=60$. 
%We first learn the dictionary from the randomly discretized input functions. We then use the embeddings to train the physics-informed neural operator. 
All the experiments outlined in this section are trained using a consistent batch size of 64 and 25000 iterations of Adam. We study the performance of three activation functions: \textit{ReLU, Tanh} and \textit{Mish}~\cite{misra2019mish}. %Moreover, we re-frame the loss functional by hard-constraining the initial condition as outlined in Section~\ref{sec:pi_rino_no_training_hard_constraint}. To enforce physics, the finite difference solver is employed in physical space by reconstructing the input functions $u(x)$ using the pre-trained basis functions. 

\begin{figure}[!ht]
\begin{centering}
\includegraphics[width=0.98\textwidth]{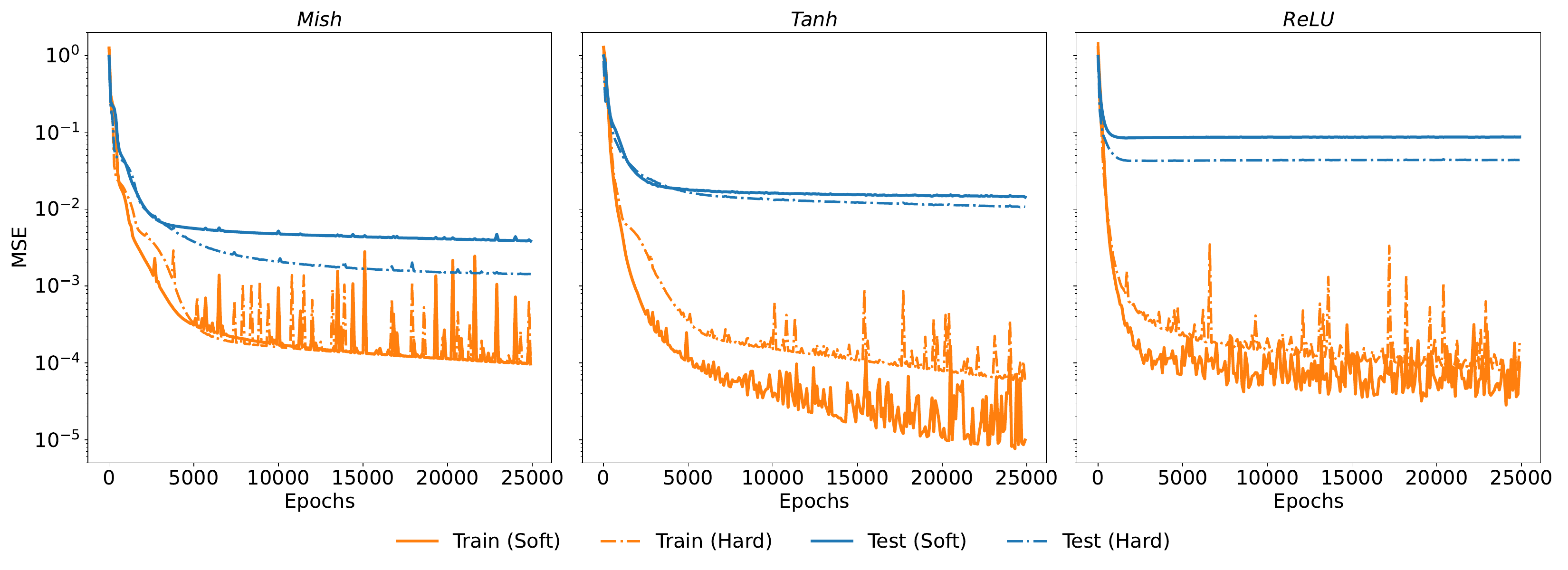}
\par\end{centering}
\caption{Antiderivative Example: Convergence profiles (MSE vs. Epochs) for the training and test sets for three different activation functions when the initial conditions are hard/soft constrained.}\label{fig:1d_antiderrivative_convergence}
\end{figure}

\begin{figure}[!ht]
    \centering
    \begin{subfigure}[b]{0.3\textwidth}
        \centering
        \includegraphics[width=\textwidth]{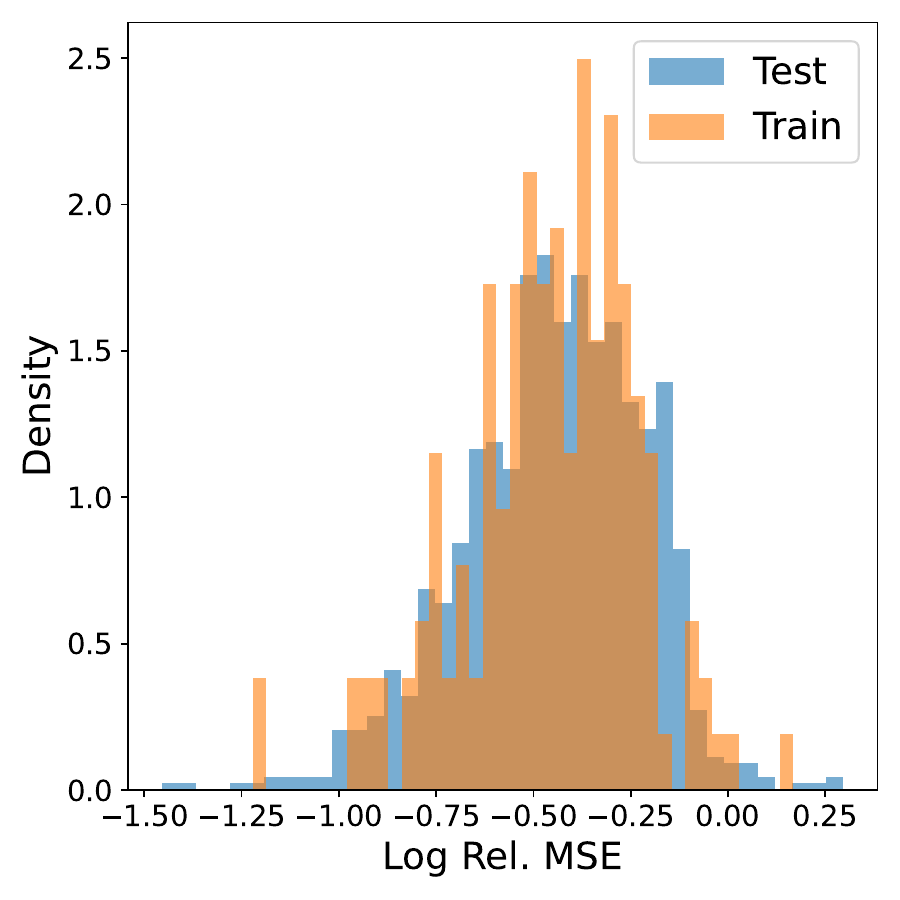}
        \caption{}
        \label{fig:1d_density_hard_mish}
    \end{subfigure}
    \hspace{2.0pt}
    \begin{subfigure}[b]{0.3\textwidth}
        \centering
        \includegraphics[width=\textwidth]{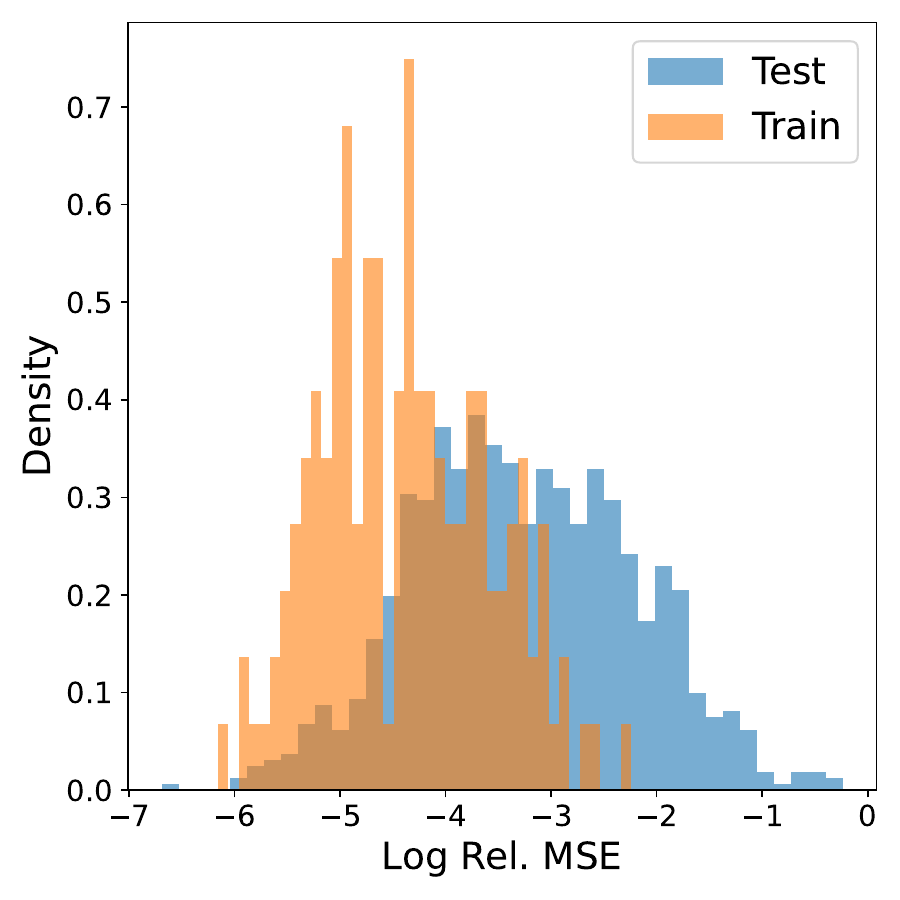}
        \caption{}
        \label{fig:1d_density_hard_tanh}
    \end{subfigure}    
    \hspace{2.0pt}
    \begin{subfigure}[b]{0.3\textwidth}
        \centering
        \includegraphics[width=\textwidth]{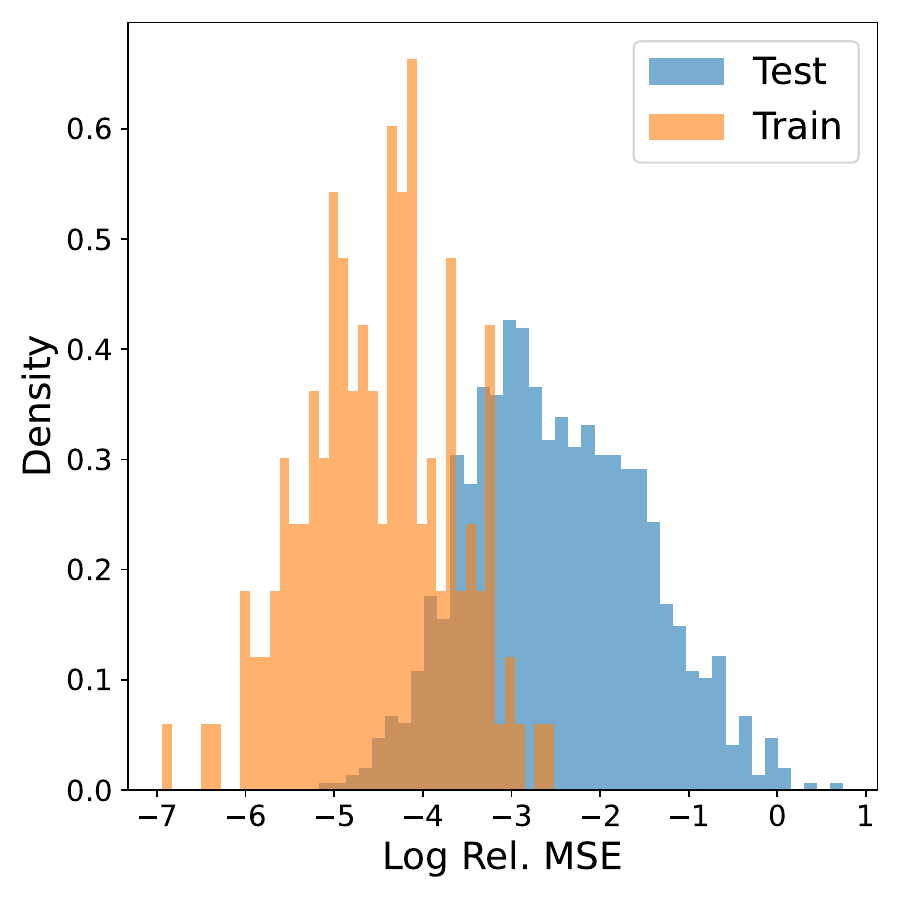}
        \caption{}
        \label{fig:1d_density_hard_relu}
    \end{subfigure} \\
    \caption{Antiderivative Example: 
 of output function prediction errors (in log rel. MSE) for the training and testing datasets after training for (a) Mish, (b) Tanh, and (c) ReLU activation functions when the initial condition was hard constrained during training.}\label{fig:1d_antiderrivative_histogram}
\end{figure}

Figure~\ref{fig:1d_antiderrivative_convergence} shows the convergence profiles using the three activation functions. The solid lines show the covergence profile when the initial condition is soft-constrained, while the dash-dot lines when it is hard-constrained. As it is observed, the generalization capability of \textit{Tanh} and \textit{ReLU} is weaker than \textit{Mish} for both approaches. When compared between the hard and soft constraints, the hard constraints show better convergence for both train and test sets, which is expected. Figure~\ref{fig:1d_antiderrivative_histogram} shows the distribution of output function prediction errors after training for both the training and testing sets when the initial condition is hard constrained. Motivated by these results, from this point forward all loss functions are hard-constrained wherever Dirichlet boundary conditions are imposed.

\begin{figure}[!ht]
    \centering
    \begin{subfigure}[b]{0.32\textwidth}
        \centering
        \includegraphics[width=\textwidth]{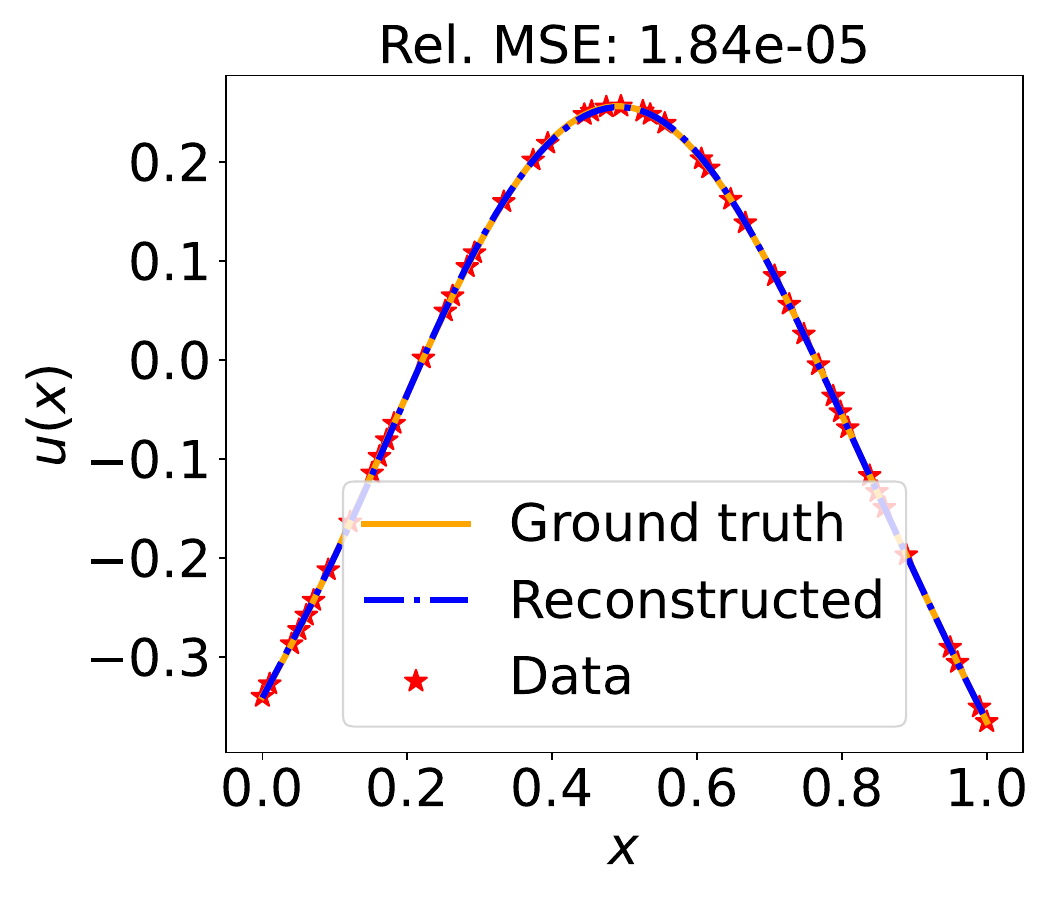}
        \caption{}
        \label{fig:1d_max_train_u}
    \end{subfigure}
    \hspace{2.0pt}
    \begin{subfigure}[b]{0.32\textwidth} 
        \centering
        \includegraphics[width=\textwidth]{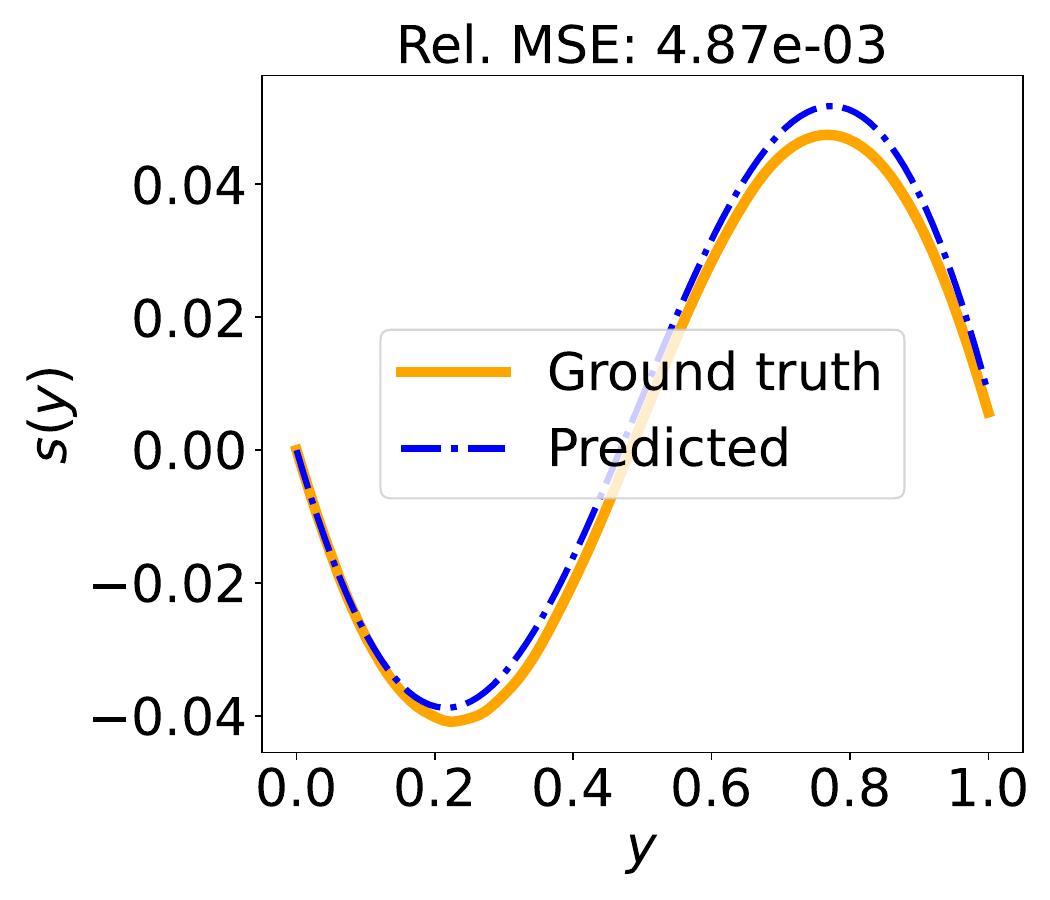}
        \caption{}
        \label{fig:1d_max_train_s} 
    \end{subfigure}
    \hspace{2.0pt}
    \begin{subfigure}[b]{0.32\textwidth} 
        \centering
        \includegraphics[width=\textwidth]{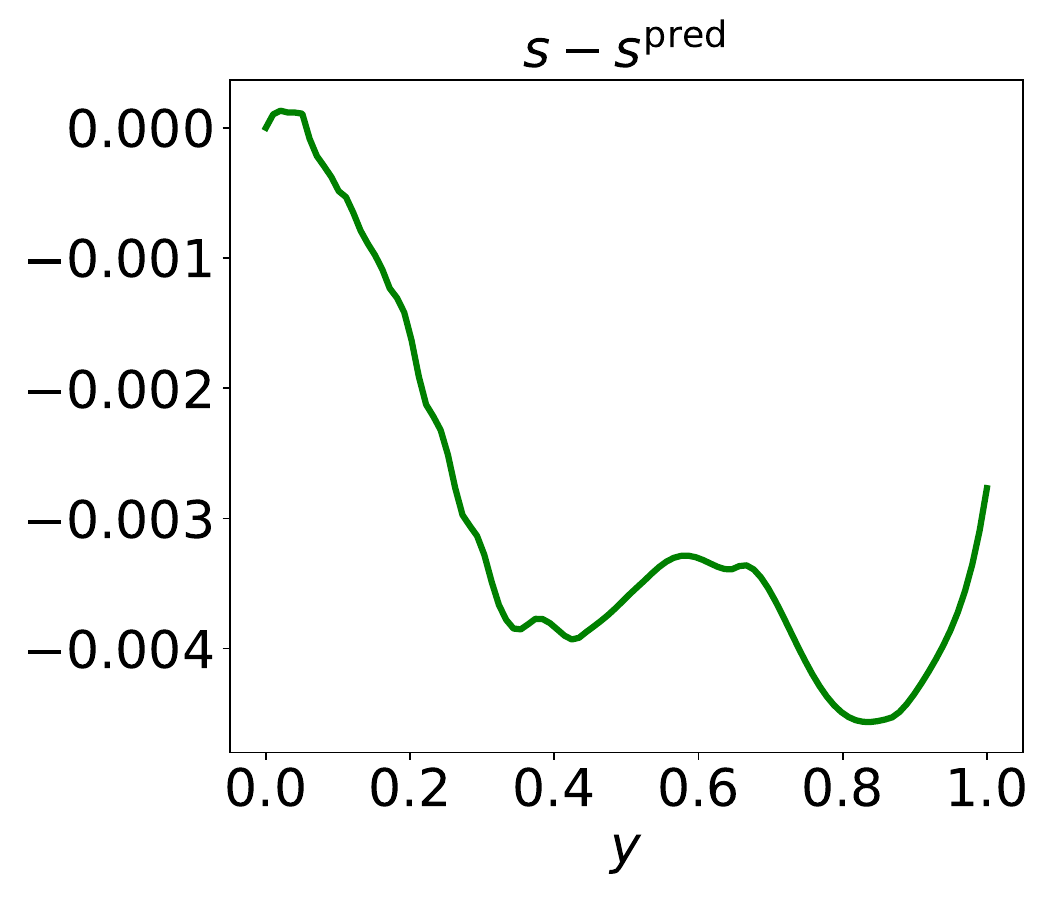}
        \caption{}
        \label{fig:1d_max_train_s_error} 
    \end{subfigure}\\
    \centering
     \begin{subfigure}[b]{0.32\textwidth}
        \centering
        \includegraphics[width=\textwidth]{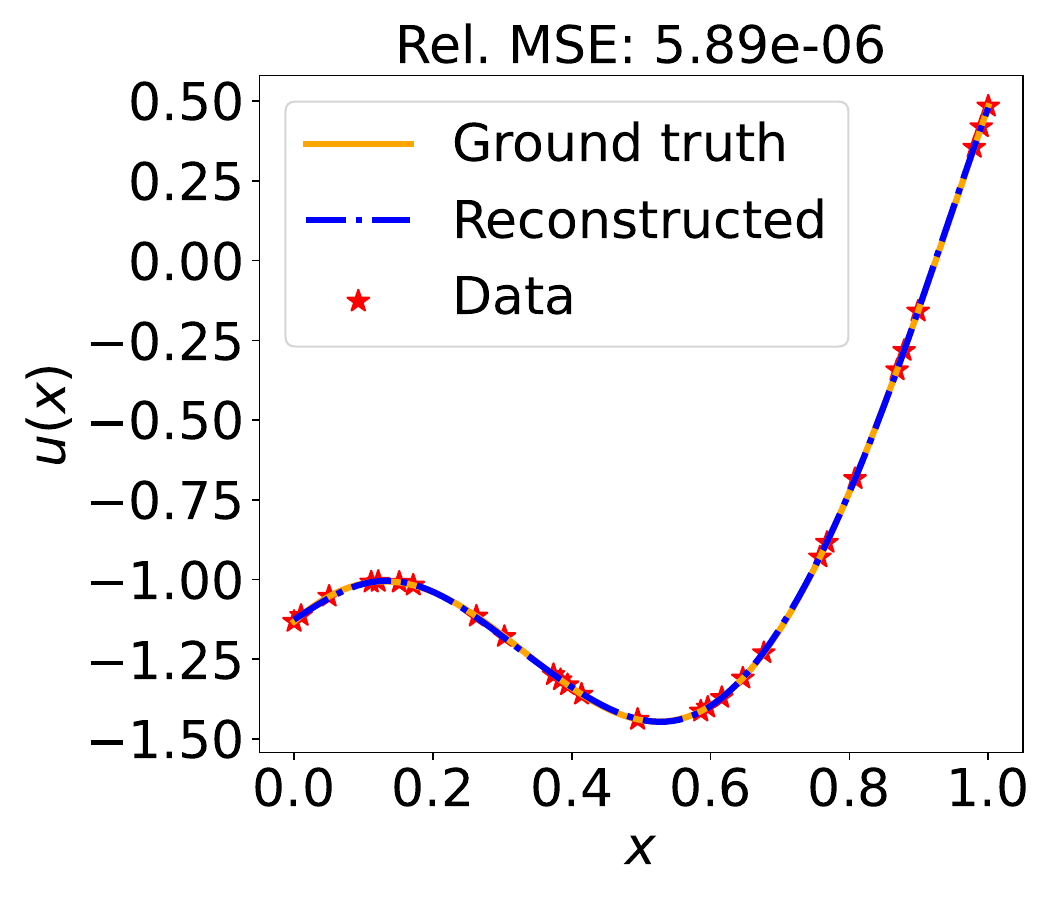}
        \caption{}
        \label{fig:1d_max_test_u}
    \end{subfigure}
    \hspace{2.0pt}
    \begin{subfigure}[b]{0.32\textwidth}
        \centering
        \includegraphics[width=\textwidth]{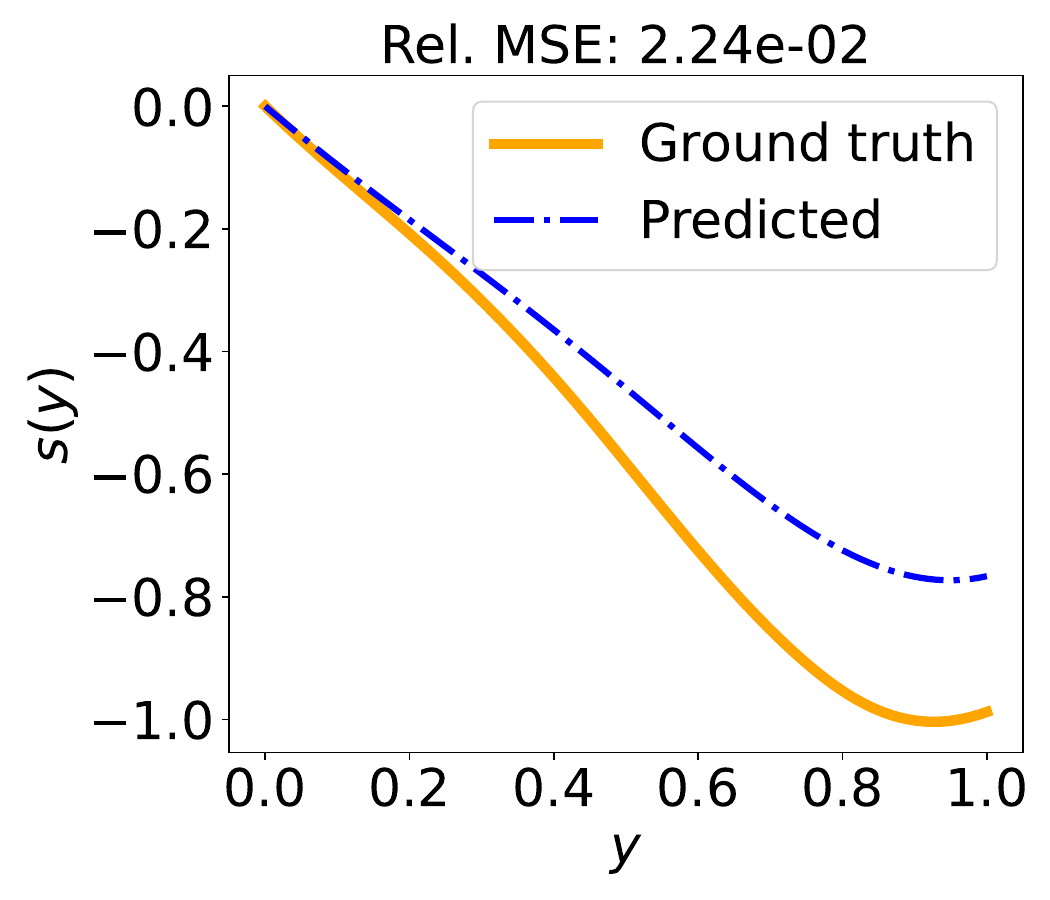}
        \caption{}
        \label{fig:1d_max_test_s}
    \end{subfigure}    
    \hspace{2.0pt}
    \begin{subfigure}[b]{0.33\textwidth}
        \centering
        \includegraphics[width=\textwidth]{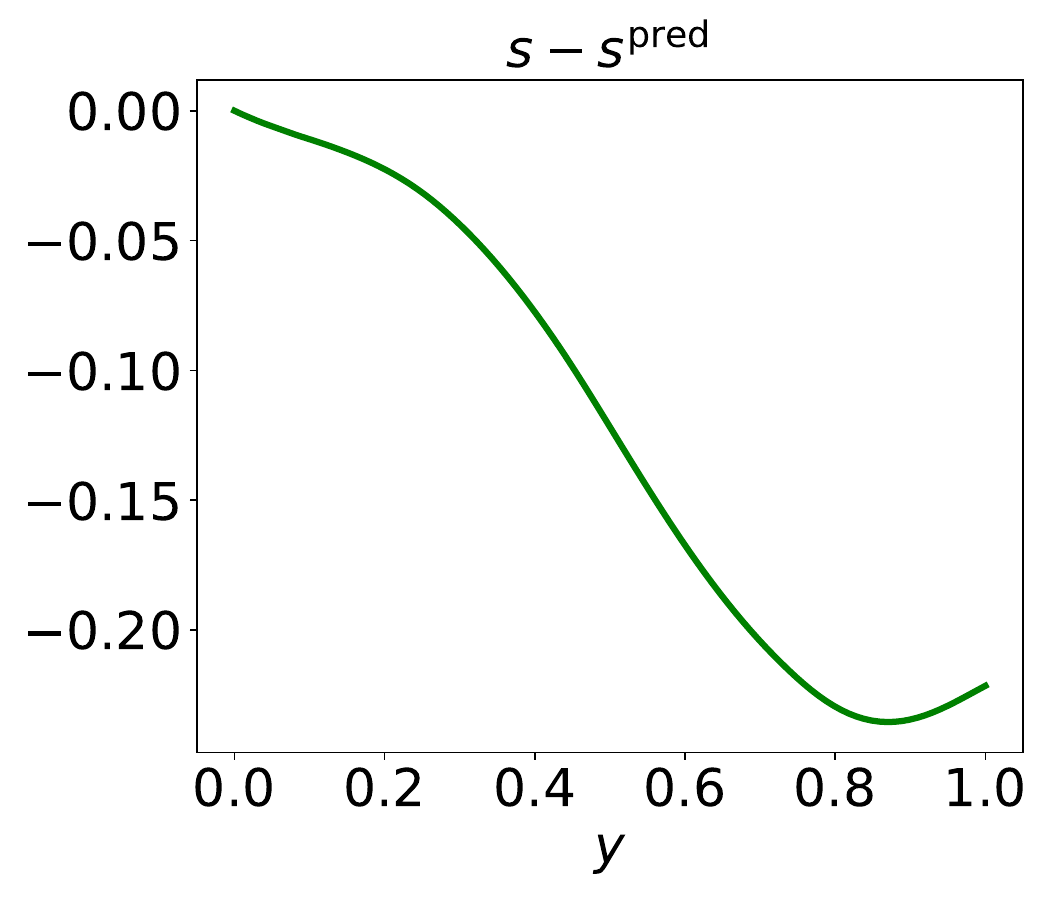}
        \caption{}
        \label{fig:1d_max_test_s_error}
    \end{subfigure}
    \caption{Antiderivative Example: Worst-case PI-RINO prediction. The sample with the highest error in the training set is shown in (a)--(c) and sample with the highest error in the test set is shown in (d)--(f).  The first column (a,d) shows the queried input function data as ‘$\star$’ along with the reconstruction from dictionary learning. The second column (b,e) shows the PI-RINO predictions along with the true solutions, and the third column (c,f) shows the absolute error of output prediction distribution across the domain. All models use the \textit{Mish} activation function.}
    \label{fig:1d_antiderrivative_results-worst}
\end{figure}
\begin{figure}
    \centering
    \begin{subfigure}[b]{0.32\textwidth}
        \centering
        \includegraphics[width=\textwidth]{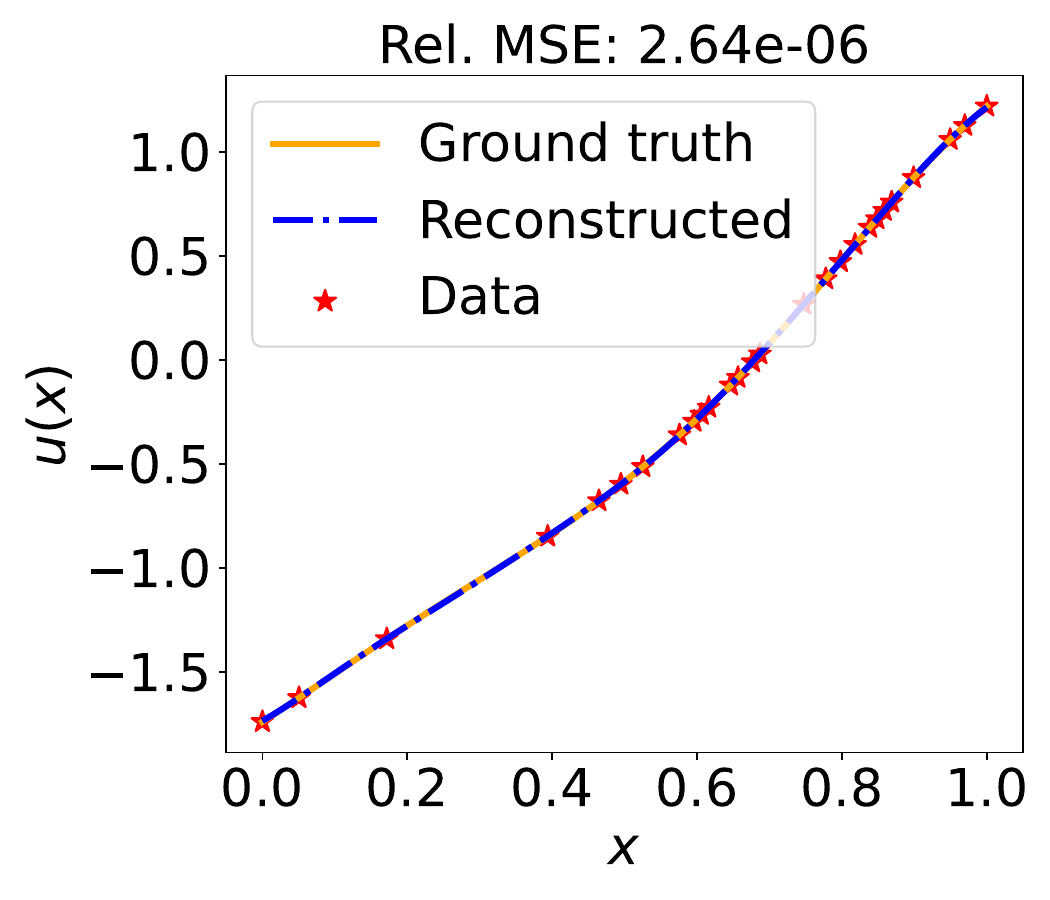}
        \caption{}
        \label{fig:1d_25per_train_u}
    \end{subfigure}
    \hspace{2.0pt}
    \begin{subfigure}[b]{0.32\textwidth}
        \centering
        \includegraphics[width=\textwidth]{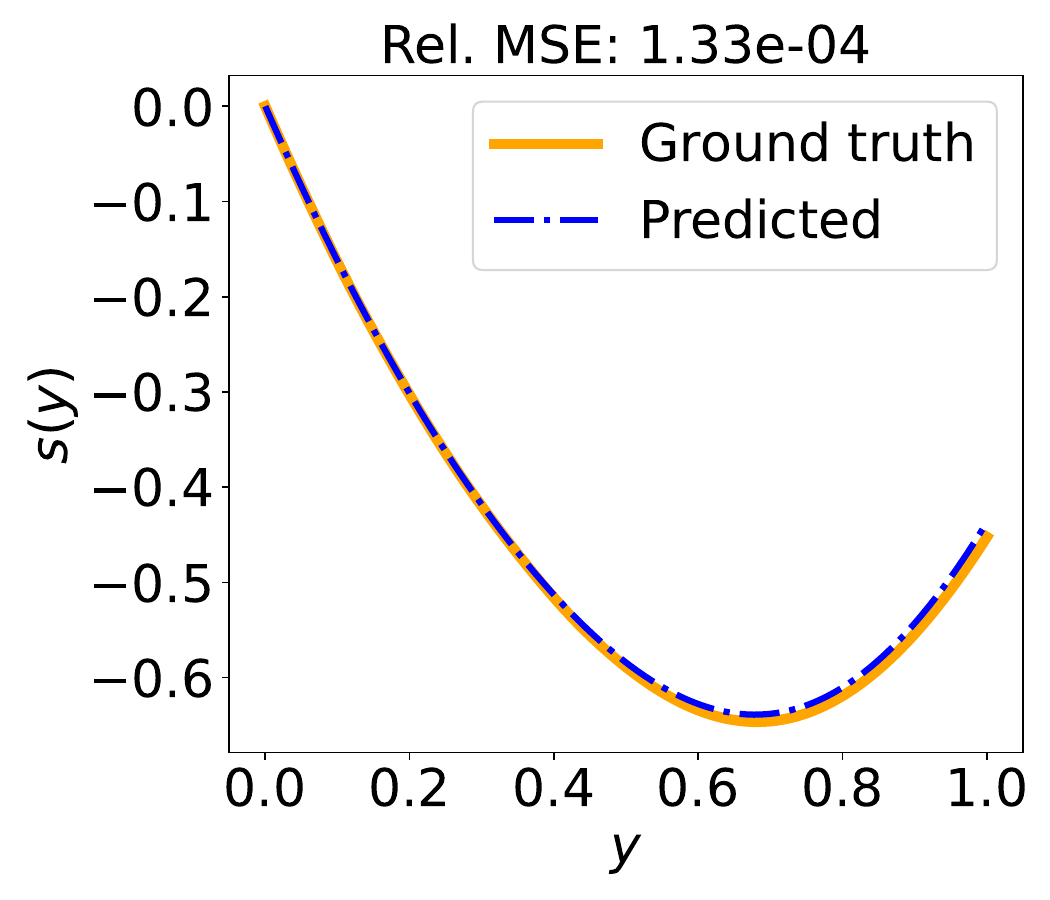}
        \caption{}
        \label{fig:1d_25per_train_s}
    \end{subfigure}    
    \hspace{2.0pt}
    \begin{subfigure}[b]{0.32\textwidth}
        \centering
        \includegraphics[width=\textwidth]{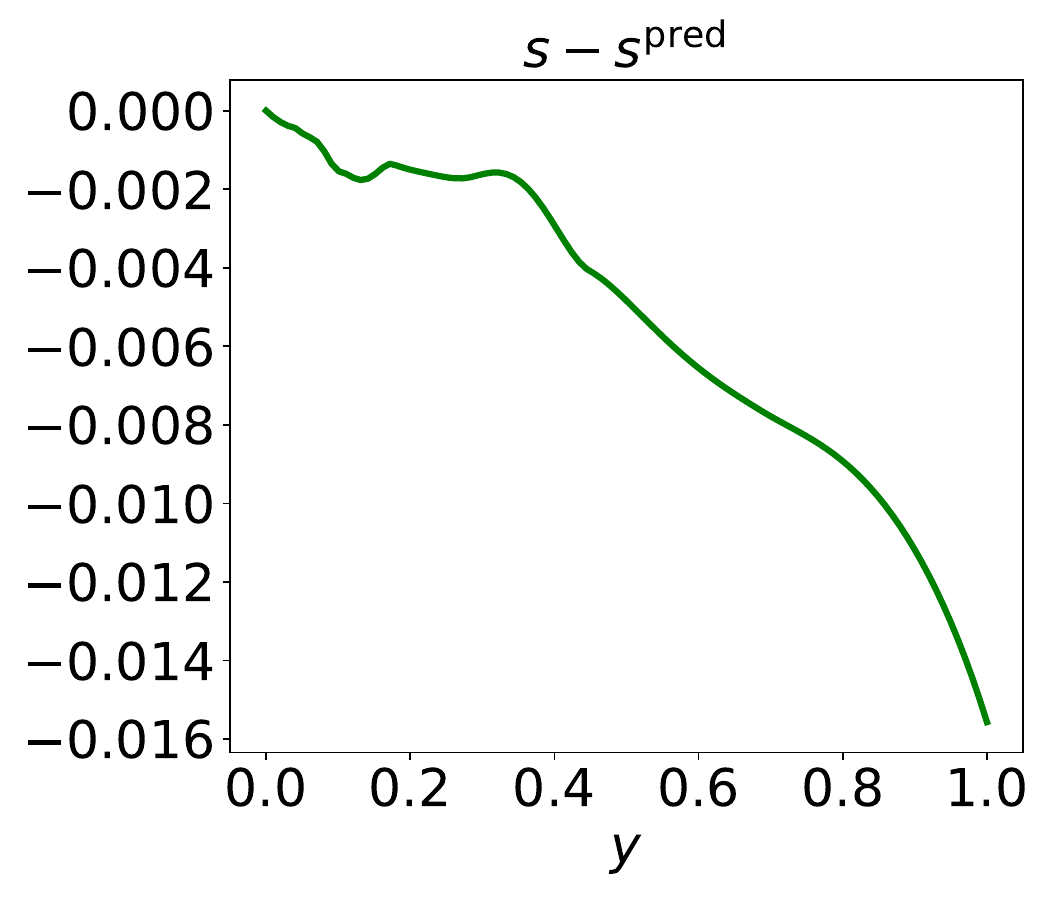}
        \caption{}
        \label{fig:1d_25per_train_s_error}
    \end{subfigure} \\
    \centering
    \begin{subfigure}[b]{0.32\textwidth}
        \centering
        \includegraphics[width=\textwidth]{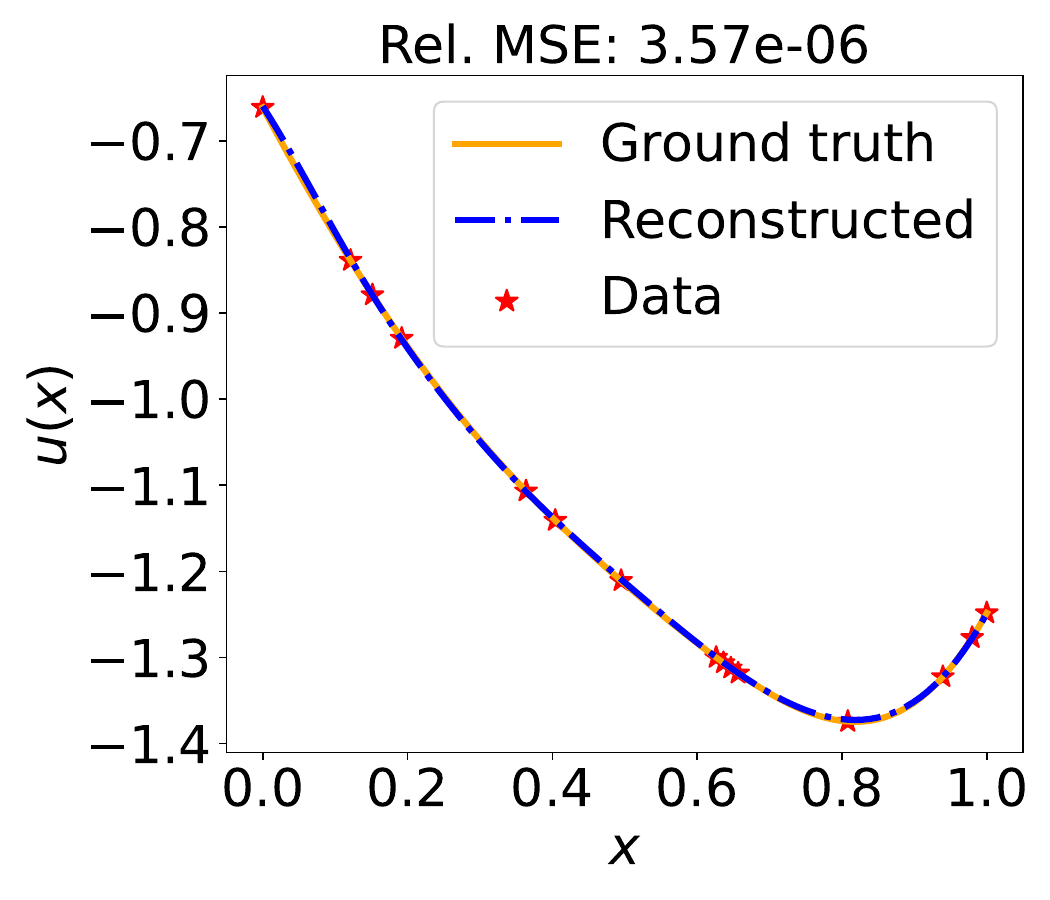}
        \caption{}
        \label{fig:1d_25per_test_u}
    \end{subfigure}
    \hspace{2.0pt}
    \begin{subfigure}[b]{0.32\textwidth}
        \centering
        \includegraphics[width=\textwidth]{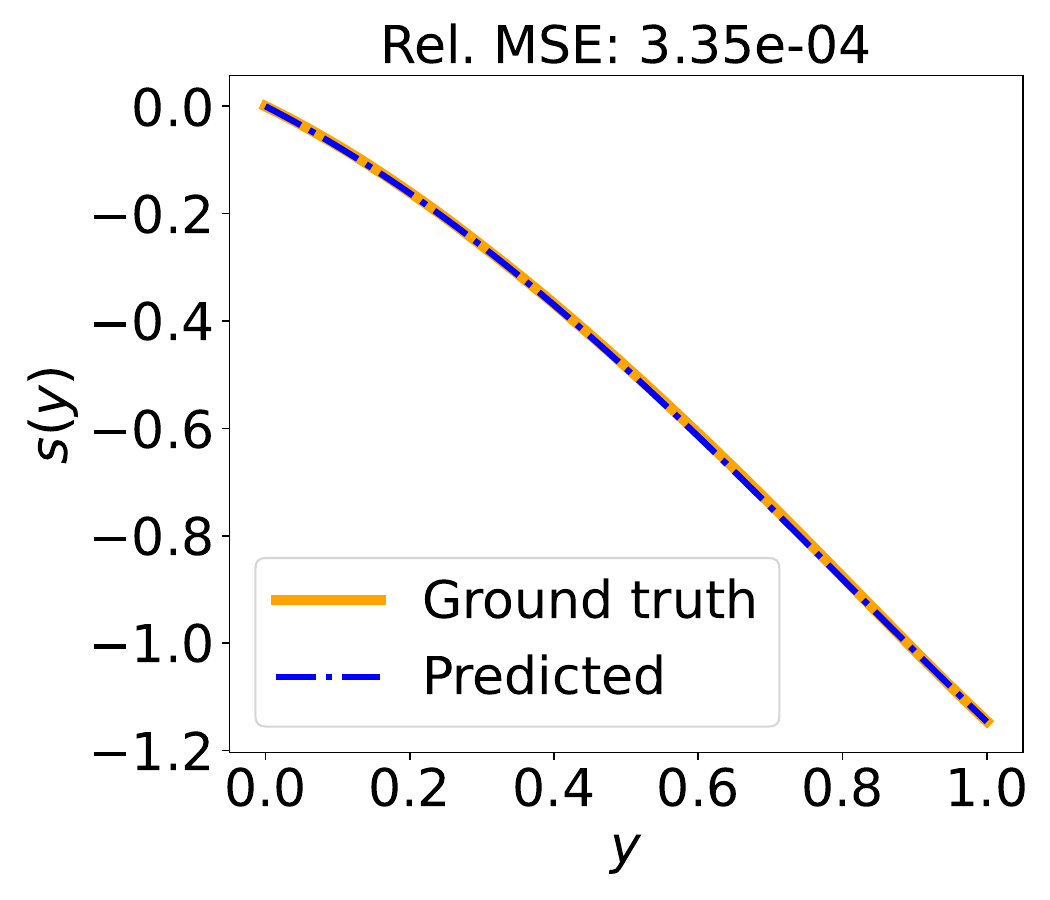}
        \caption{}
        \label{fig:1d_25per_test_s}
    \end{subfigure}    
    \hspace{2.0pt}
    \begin{subfigure}[b]{0.32\textwidth}
        \centering
        \includegraphics[width=\textwidth]{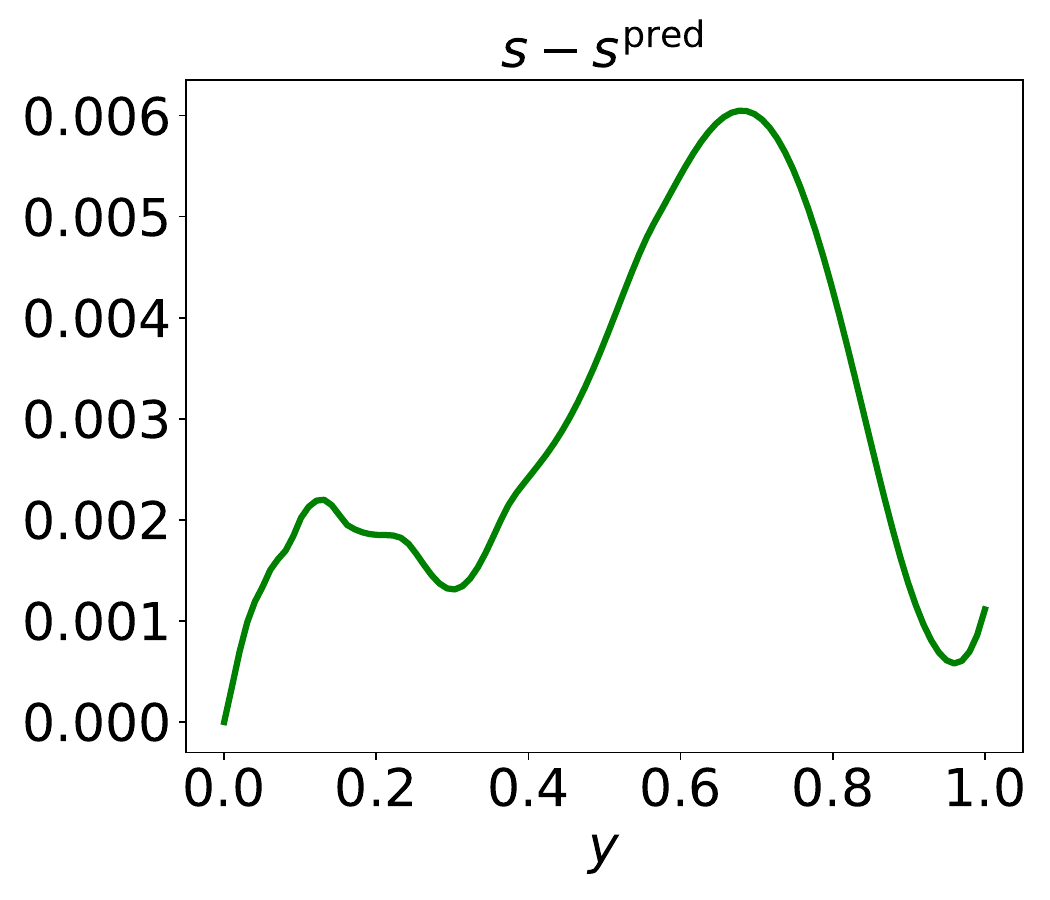}
        \caption{}
        \label{fig:1d_25per_test_s_error}
    \end{subfigure}
    \caption{Antiderivative Example: $25^{th}$ percentile PI-RINO prediction. The sample with the $25^{th}$ percentile error in the training set is shown in (a)--(c) and sample with the $25^{th}$ percentile error in the test set is shown in (d)--(f).  The first column (a,d) shows the queried input function data as ‘$\star$’ along with the reconstruction from dictionary learning. The second column (b,e) shows the PI-RINO predictions along with the true solutions, and the third column (c,f) shows the absolute error of output prediction distribution across the domain. All models use the \textit{Mish} activation function.}
    \label{fig:1d_antiderrivative_results-25}
\end{figure}

Figure~\ref{fig:1d_antiderrivative_results-worst} and \ref{fig:1d_antiderrivative_results-25} shows the prediction of the PI-RINO using the \textit{Mish} activation function for 
% two classes of input functions, with the first and second rows showcasing 
the worst-case sample (with the highest error in output function prediction) 
% while the second and fourth rows depict 
and the sample with the 25th percentile (Q1) of the error distribution (i.e., $75\%$ of cases have lower error than this sample). In each case, the first row ((a)--(c)) shows the corresponding sample from the training set and the second row ((d)--(f)) shows the sample from the test set. The first column shows the queried input function, which includes a point cloud of random data points shown as ‘$\star$’ along with the reconstructed input from dictionary learning and the true function. The second column shows the prediction of the output function along with the ground truth, while the last column shows the error of the output predictions relative to the true solution. The results demonstrate that PI-RINO shows good approximation in predicting the output fields from input data sampled at random locations. Table~\ref{table:error_statistics} details the mean, standard deviation, maximum and (Q1) errors of the training and test sets.

\begin{table}[!ht]
\centering
\begin{tabular}{@{}lcccccccc@{}}
\toprule
\multirow{2}{*}{\textbf{Problem}} & \multicolumn{4}{c}{\textbf{Training (Relative Errors)}} & \multicolumn{4}{c}{\textbf{Test (Relative Errors)}} \\
\cmidrule(lr){2-5} \cmidrule(lr){6-9}
 & Mean ($\mu$) & Std ($\sigma$) & Max & Q1 & Mean ($\mu$) & Std ($\sigma$) & Max & Q1 \\
\midrule
Antiderivative     & 1.75e-04 & 4.04e-04 & 2.90e-03 & 1.18e-04 & 7.08e-04 & 3.77e-03 & 9.22e-02 & 3.41e-04 \\
Heat conduction    & 1.17e-04 & 1.46e-04 & 1.30e-03 & 1.40e-04 & 1.71e-03 & 2.28e-03 & 1.77e-02 & 2.20e-03 \\
Biot (pressure)               & 5.29e-03 & 7.38e-03 & 5.67e-02 & 6.58e-03 & 5.22e-03 & 7.10e-03 & 4.38e-02 & 6.88e-03 \\
Biot (displacements)               & 2.39e-03 & 4.03e-03 & 4.28e-02 & 2.91e-03 & 2.29e-03 & 3.71e-03 & 2.57e-02 & 2.77e-03 \\
\bottomrule
\end{tabular}
\caption{Error statistics (mean, standard deviation, maximum, and first quartile) for training and test datasets across three benchmark problems.}
\label{table:error_statistics}
\end{table}

\subsection{Heat Conduction}\label{sec:num_examples_2d_poissons}

In this example we increase the complexity by considering a two-dimensional setting. The heat equation in two dimensions is given as:
\begin{equation}
-\kappa \left( \frac{\partial^2 s}{\partial y_1^2} + \frac{\partial^2 s}{\partial y_2^2} \right) = u(y_1,y_2), \quad (y_1,y_2) \in [0,1] \times [0,1]
\end{equation}
where we learn the solution operator that maps the forcing function $u(x_1, x_2)$ to the PDE solution $s(y_1,y_2)$. Dirichlet boundary conditions of $s=0$ are applied on all four boundaries. The conductivity is set to $\kappa=+1$. The forcing function $u(x_1,x_2)$ is a zero-mean Gaussian random field (with the radial basis function as the covariance functions \cite{bahmani2025resolution}). Just like in the 1D example, we simulate multi-resolution data by considering only sparse input data are available. To achieve this, the random fields are discretized on a fine grid with a resolution of $20 \times 20$, and the multi-resolution dataset is generated by setting $M_{\text{min}} = 200$ and $M_{\text{max}} = 280$.
The functions $u(x_1,x_2)$ are reconstructed from the sparse input data available using the pre-trained basis functions. These reconstructed $\tilde{u}^(y_1,y_2)$ are used to find the reference $s(y_1,y_2)$. The PI-RINO was trained using on 800 different training samples, and the results were tested on 200 samples from the test set.The errors for the predicted solutions presented in this section are computed with respect to the finite element (FE) solution, which is treated as the ground truth.

\begin{figure}[!ht]
    \centering
    \begin{subfigure}[b]{0.45\textwidth}
        \centering
        \includegraphics[width=\textwidth]{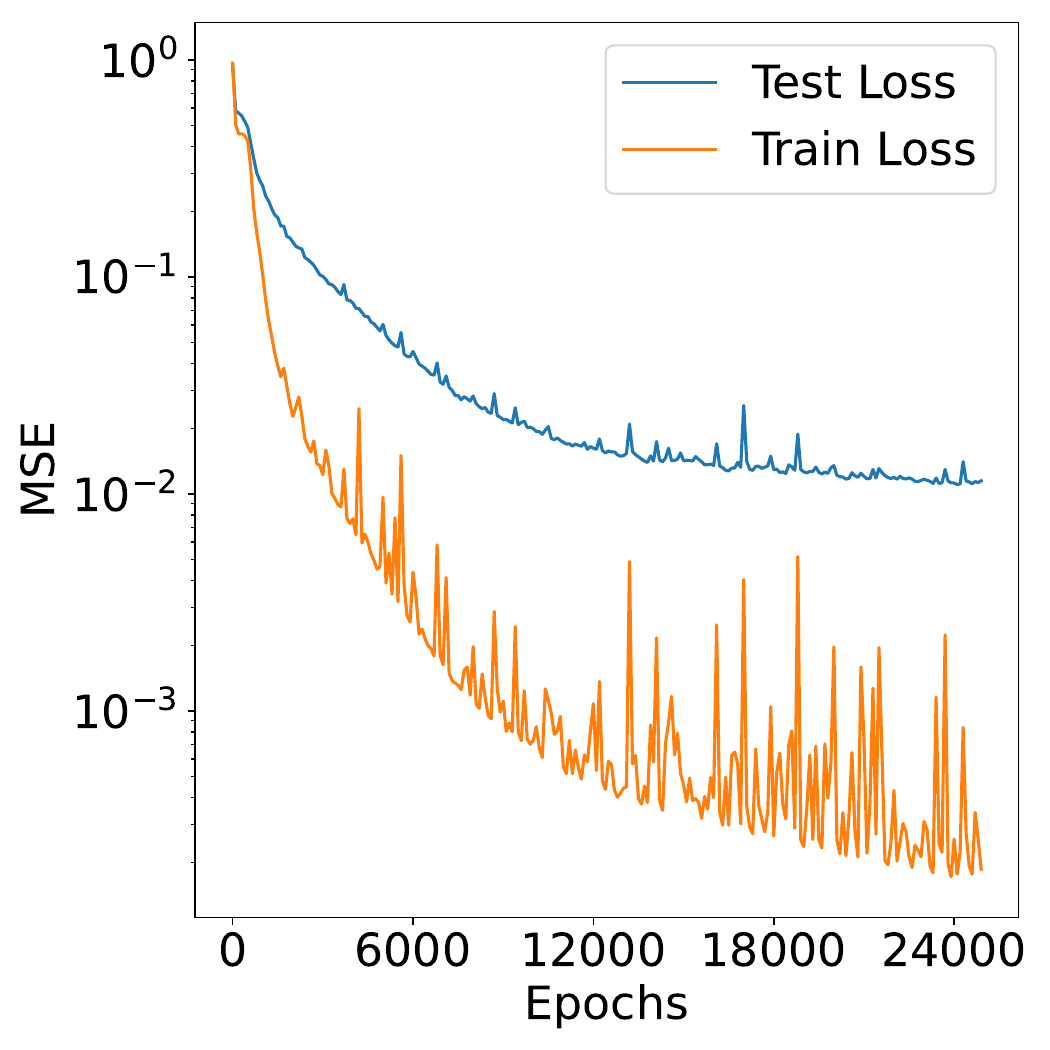}
        \caption{}
        \label{fig:2d_poissons_convergence}
    \end{subfigure}
    \hspace{2.0pt}
    \begin{subfigure}[b]{0.45\textwidth}
        \centering
        \includegraphics[width=\textwidth]{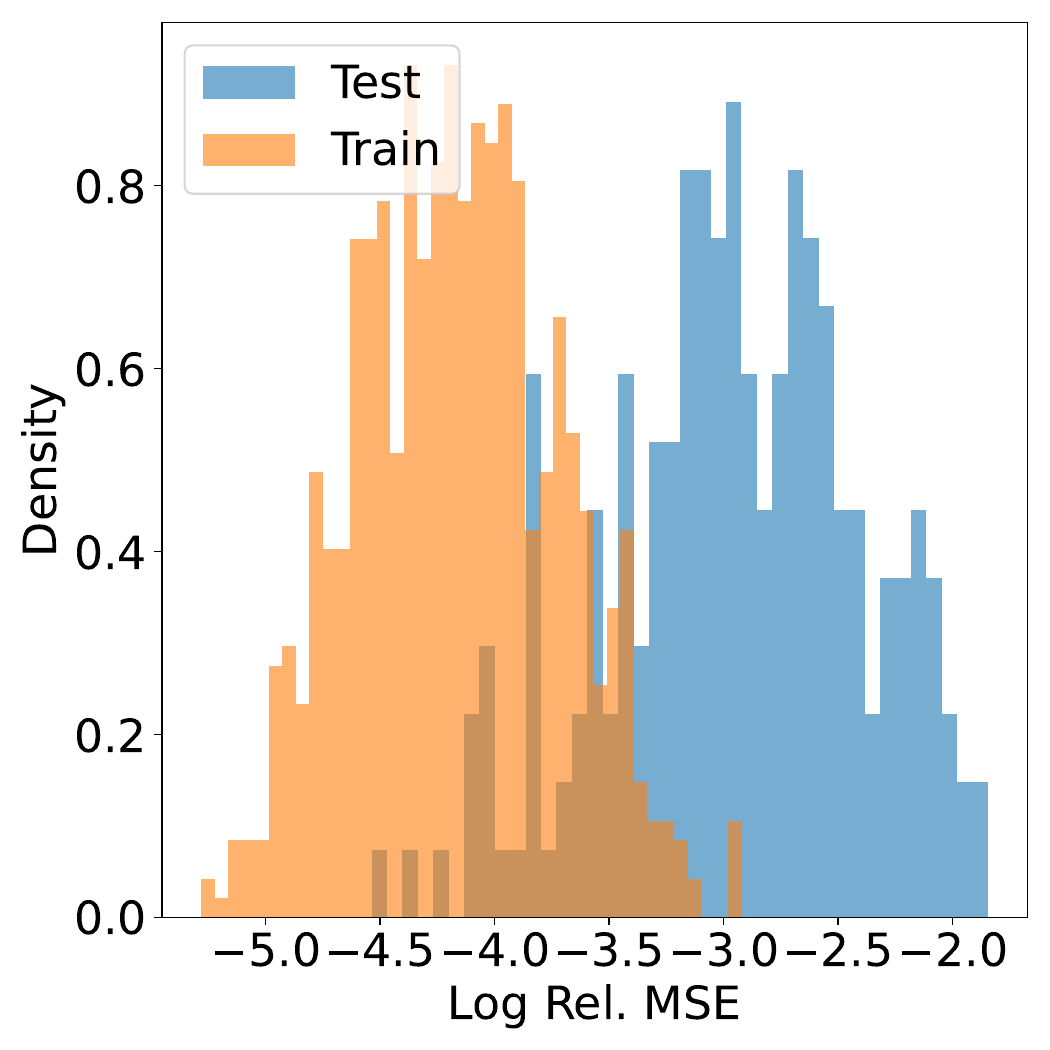}
        \caption{}
        \label{fig:2d_poissons_error_density}
    \end{subfigure}  \\
    \caption{2D Heat Example: (a) Convergence profile of loss function during optimization iterations for the PI-RINO, (b) Distribution of output function prediction errors (relative MSEs) for the training and testing datasets after training.
    }\label{fig:2d_poissons_congergence_error_density}
\end{figure}

The activation function for the neural operator is chosen to be \textit{Mish}. Figure~\ref{fig:2d_poissons_convergence} shows the convergence plots (objective vs. loss) for both the training and test datasets during the training process, while Figure~\ref{fig:2d_poissons_error_density} shows the distribution of the training and test errors (in terms of relative MSEs) after training is complete. It should be noted that although the test errors are slightly higher compared to training errors (unlike in the data-driven case where training and test errors may be comparable), the learned PDE solution adheres to the physics much more closely compared to the purely data-driven case, which is investigated further in Section~\ref{sec:autodiff_compare}.

\begin{figure}[!hbt]
    \centering
    \begin{subfigure}[b]{0.19\textwidth}
        \centering
        \includegraphics[width=\textwidth]{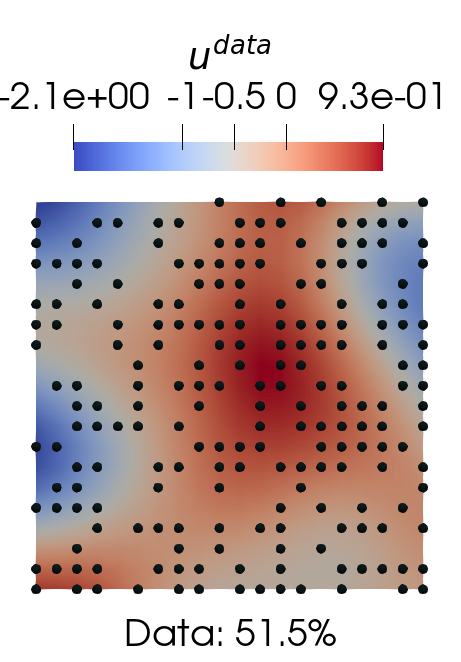}
        \caption{}
        \label{fig:2d_poissons_test_max_u_ground}
    \end{subfigure}
    \hspace{1.0pt}
    \begin{subfigure}[b]{0.19\textwidth} 
        \centering
        \includegraphics[width=\textwidth]{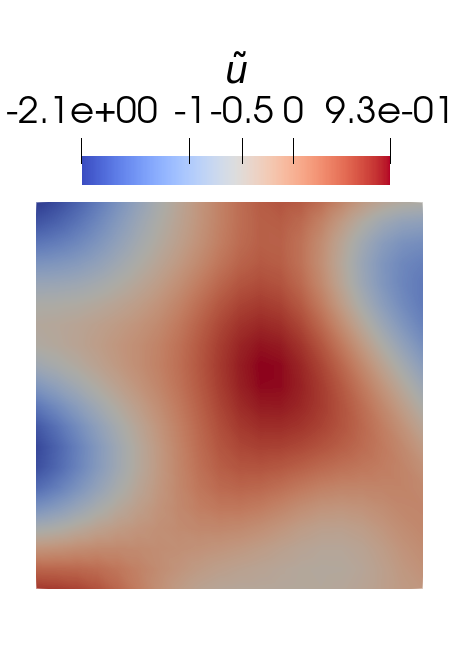}
        \caption{}
        \label{fig:2d_poissons_test_max_u_pred} 
    \end{subfigure}
    \hspace{1.0pt}
    \begin{subfigure}[b]{0.19\textwidth} 
        \centering
        \includegraphics[width=\textwidth]{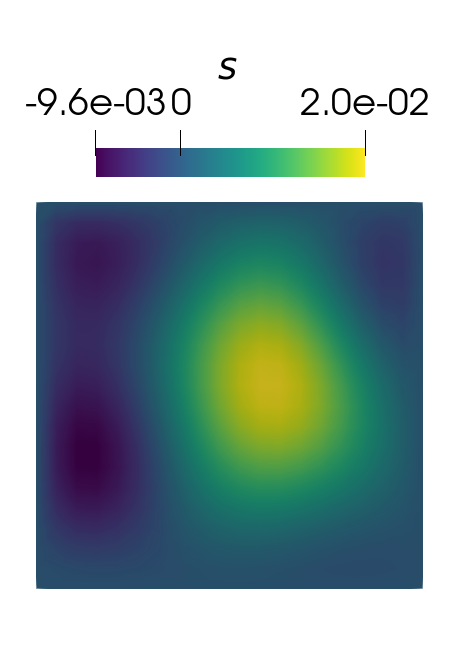}
        \caption{}
        \label{fig:2d_poissons_test_max_s_ground} 
    \end{subfigure}
    \hspace{1.0pt}
    \begin{subfigure}[b]{0.19\textwidth} 
        \centering
        \includegraphics[width=\textwidth]{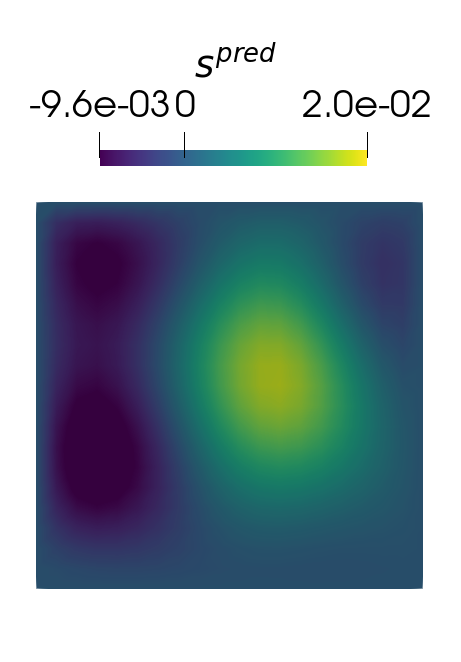}
        \caption{}
        \label{fig:2d_poissons_test_max_s_pred} 
    \end{subfigure}
    \hspace{1.0pt}
    \begin{subfigure}[b]{0.19\textwidth} 
        \centering
        \includegraphics[width=\textwidth]{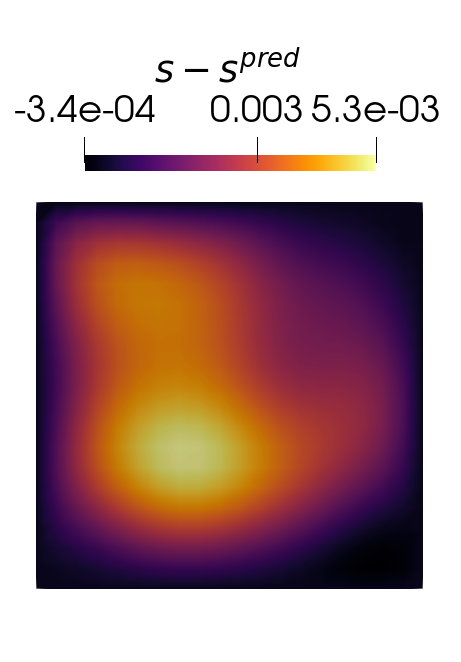}
        \caption{}
        \label{fig:2d_poissons_test_max_s_error} 
    \end{subfigure}\\
    \centering
    \begin{subfigure}[b]{0.19\textwidth}
        \centering
        \includegraphics[width=\textwidth]{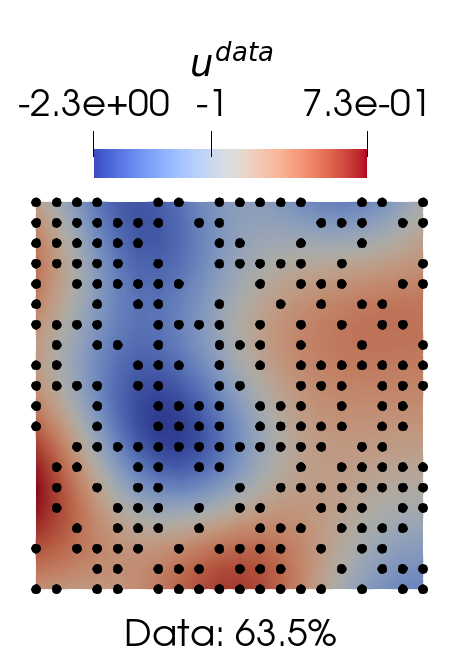}
        \caption{}
        \label{fig:2d_poissons_test_25per_u_ground}
    \end{subfigure}
    \hspace{1.0pt}
    \begin{subfigure}[b]{0.19\textwidth} 
        \centering
        \includegraphics[width=\textwidth]{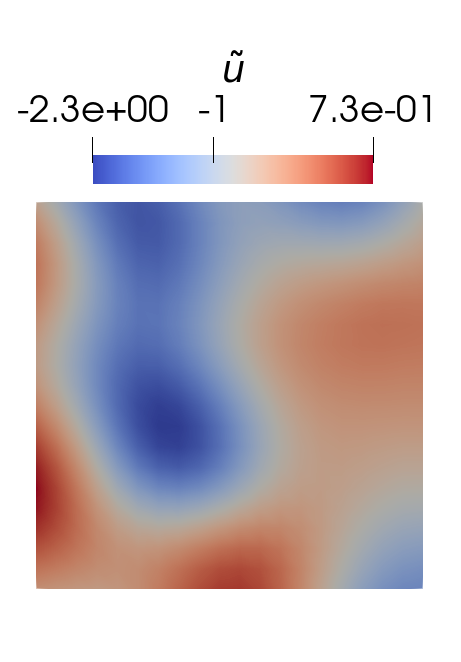}
        \caption{}
        \label{fig:2d_poissons_test_25per_u_pred} 
    \end{subfigure}
    \hspace{1.0pt}
    \begin{subfigure}[b]{0.19\textwidth} 
        \centering
        \includegraphics[width=\textwidth]{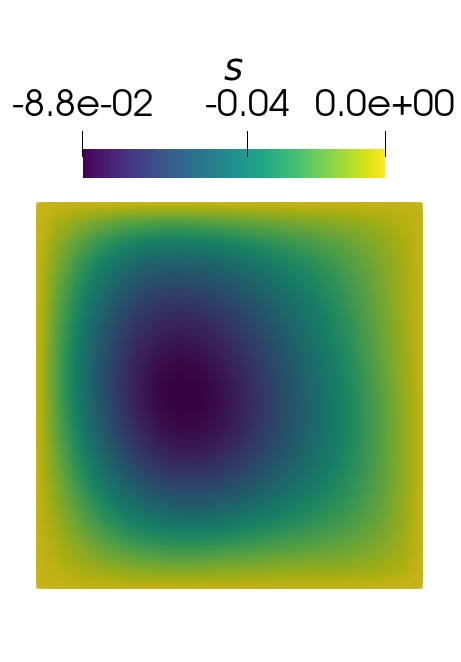}
        \caption{}
        \label{fig:2d_poissons_test_25per_s_ground} 
    \end{subfigure}
    \hspace{1.0pt}
    \begin{subfigure}[b]{0.19\textwidth} 
        \centering
        \includegraphics[width=\textwidth]{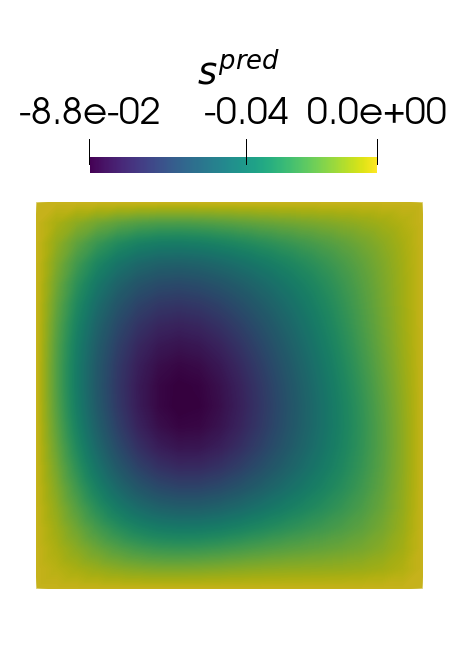}
        \caption{}
        \label{fig:2d_poissons_test_25per_s_pred} 
    \end{subfigure}
    \hspace{1.0pt}
    \begin{subfigure}[b]{0.19\textwidth} 
        \centering
        \includegraphics[width=\textwidth]{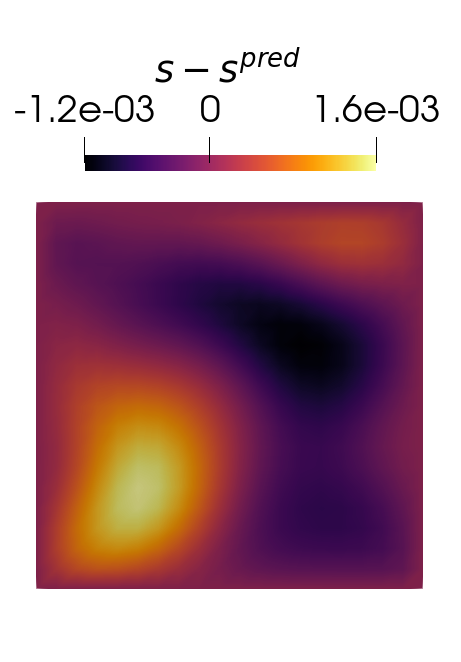}
        \caption{}
        \label{fig:2d_poissons_test_25per_s_error} 
    \end{subfigure}\\
    \caption{2D Heat Example: PI-RINO predictions for two samples: (a)--(e) shows the sample with the worst output predictions from the test set; (f)--(j) show the sample with the 25-percentile prediction (this mean 75\% of testing errors from the dataset are lower than this) from the test set. The first column (a, f) shows the queried input function data as `$\bullet$’, along with ground truth while the second column (b, g) shows the reconstructed full-field input function data from sparse observations using the learned dictionary. The third (c, g) and fourth (d, i) columns show the ground truth and PI-RINO predicted output fields. The last column (e, j) shows the output function prediction errors.}
    \label{fig:2d_poissons_results}
\end{figure}

Figure~\ref{fig:2d_poissons_results} shows the approximations for two cases from the test dataset. The top row (a--e) shows the results that produce the absolute highest errors from the testing dataset, while the bottom row (f--j) shows the sample at the 25th percentile (Q1) of the error distribution (ie, $75\%$ of cases have less error than this value). The first column (a,f) shows the queried multi-resolution input function data against the ground truth, while the second column (b,g) shows the full-field reconstructed input fuction $\tilde{u}$. The third column (c,h) shows the ground-truth PDE solution $s(x,y)$ obtained from a FE solver. The fourth column (d,i) shows the predicted solution $s^{pred}(x,y)$ from the PI-RINO. Finally, the last column (e,j) shows the error in PI-RINO predictions $s-s^{pred}$. Table~\ref{table:error_statistics} details the mean, standard deviation, maximum and (Q1) errors of the training and test sets. We see here that the proposed PI-RINO provides a high-accuracy solution, even in the worst case.

\subsection{Biot's Consolidation}\label{sec:num_examples_biot}

As a final experiment, we examine the fully coupled Biot's poroelastic equation. This equation describes the mechanical behavior of porous media influenced by the movement of fluid within the solid matrix. Biot's theory arises in a wide range of applications, including biomedical engineering~\cite{malandrino2019poroelasticity,cowin1999bone}, subsurface geomechanics~\cite{du2001poroelastic,terekhov2022finite}, and micromechanics~\cite{islam2020micromechanical}, among others. The foundational principles of poroelasticity center on the coupled interactions between deformation and diffusion. Fluid-to-solid coupling arises when changes in fluid pressure or mass induce deformation in the porous structure. Conversely, solid-to-fluid coupling occurs when mechanical stress in the solid matrix leads to corresponding changes in fluid pressure or mass. The governing equations employed here are formulated under the assumptions of quasi-static, linearized poroelasticity~\cite{coussy2004poromechanics}. We seek the displacement field $u(y,t)$ and the pressure field $p(y,t)$ that satisfy the following coupled system of equations:

\begin{subequations}
\begin{gather}
\frac{\partial}{\partial y}\left( \nu \frac{\partial u}{\partial y} \right) + \frac{\partial p}{\partial y} = 0 ,~\quad y,t \in [0,1]\times[0,1], \\
\frac{\partial}{\partial t}\left( ap + \frac{\partial u}{\partial y} \right) - \frac{\partial}{\partial y} \left( \kappa  \frac{\partial p}{\partial y} \right)=0,~\quad y,t \in [0,1]\times[0,1]
\end{gather}
\end{subequations}
with the boundary and initial conditions:
\begin{subequations}
\begin{gather}
    \nu\frac{\partial u}{\partial y}=-1 \text{  at  } y=0 \text{  and  } u=0 \text{ at  } y=1,\label{eq:bc1}\\
    p=0 \text{ at  } y=0 \text{ and  }\kappa \frac{\partial p}{\partial y}=0 \text{  at  } y=1,\label{eq:bc2} \\
    p(y) = \begin{cases}
                0, & y = 0 \\
                1, & y \neq 0
              \end{cases},~\quad\text{  at  } t=0\label{eq:ic} \\
    u(y) = 0, ~\quad\text{ at  } t=0.
\end{gather}
\end{subequations}
The problem setup described by the above system of equations corresponds to the non-dimensional form of a classical benchmark in poroelasticity—the Terzaghi consolidation problem. It basically models fluid flow through a porous medium represented by a one-dimensional vertical column with an impermeable fixed base and a permeable top subjected to a mechanical traction. The non-dimensionalization follows the methodology used in~\cite{bean2014immersed,Roy04032025}. In these equations, $\nu$ denotes the non-dimensional elasticity constant, $a$ represents the non-dimensional fluid compressibility, and $\kappa$ is the non-dimensional hydraulic conductivity. For this specific case, we assume an incompressible fluid ($a=0$) and set the elastic constant to unity ($\nu = 1$). In the context of subsurface mechanics, material properties,particularly the hydraulic conductivity $\kappa(x)$, are often spatially varying with depth and subject to uncertainty. Additionally, obtaining high-resolution, spatially resolved measurements of $\kappa(x)$ is typically impractical in real-world scenarios. Therefore, in this operator learning formulation, we assume sparse observations of $\kappa(x)$ across multiple realizations and aim to reconstruct the full-field $\kappa(x)$ profiles using the dictionary learning technique. $\kappa(x)$ were generated from sampling from a $\mathcal{GP}$ with mean $\mu=1.25$ and length-scale $l=0.2$ for the squared exponential kernel given above. The ultimate goal is to approximate the solution operator
\[
\mathcal{G}(\kappa(x)) \rightarrow \{ u(y,t),\; p(y,t) \},
\]
which maps the spatially varying hydraulic conductivity to the corresponding displacement and pressure fields over space and time. Please notice the change of notation here that $u$ now represents one of the solution fields (displacements), where previously $u$ was used to denote the input function fields, which in this case is $\kappa$.

In this example, the neural operator must approximate two output fields—the displacement and the pressure. The most straightforward approach is to construct a single PI-RINO model that outputs both fields simultaneously. An alternative, and more effective, approach is to construct two separate neural operators, each mapping from $\kappa(x)$ to one of the output fields, thereby using two distinct MLPs. These two operators interact through a shared physics-informed loss function. Empirically, the latter approach performs significantly better for poroelasticity-related problems, as previously observed in PINN-based frameworks~\cite{Roy04032025,haghighat2022physics}. This improvement may be attributed to the ability of each MLP to better capture the latent physical structure of its respective field when trained independently but coupled through physical constraints.

\begin{figure}[!ht]
    \centering
    \begin{subfigure}[b]{0.32\textwidth}
        \centering
        \includegraphics[width=\textwidth]{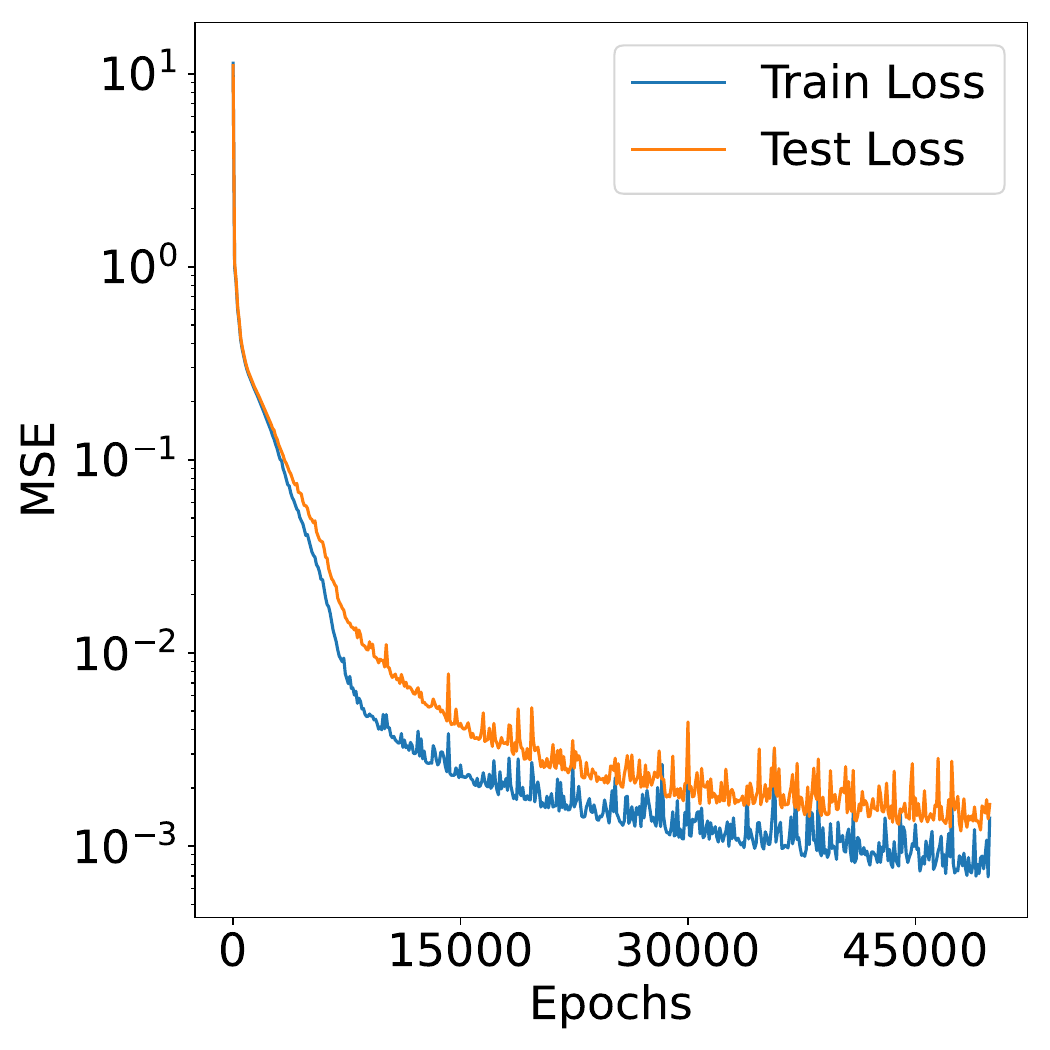}
        \caption{}
        \label{fig:biot_convergence}
    \end{subfigure}
    \hspace{2.0pt}
    \begin{subfigure}[b]{0.32\textwidth}
        \centering
        \includegraphics[width=\textwidth]{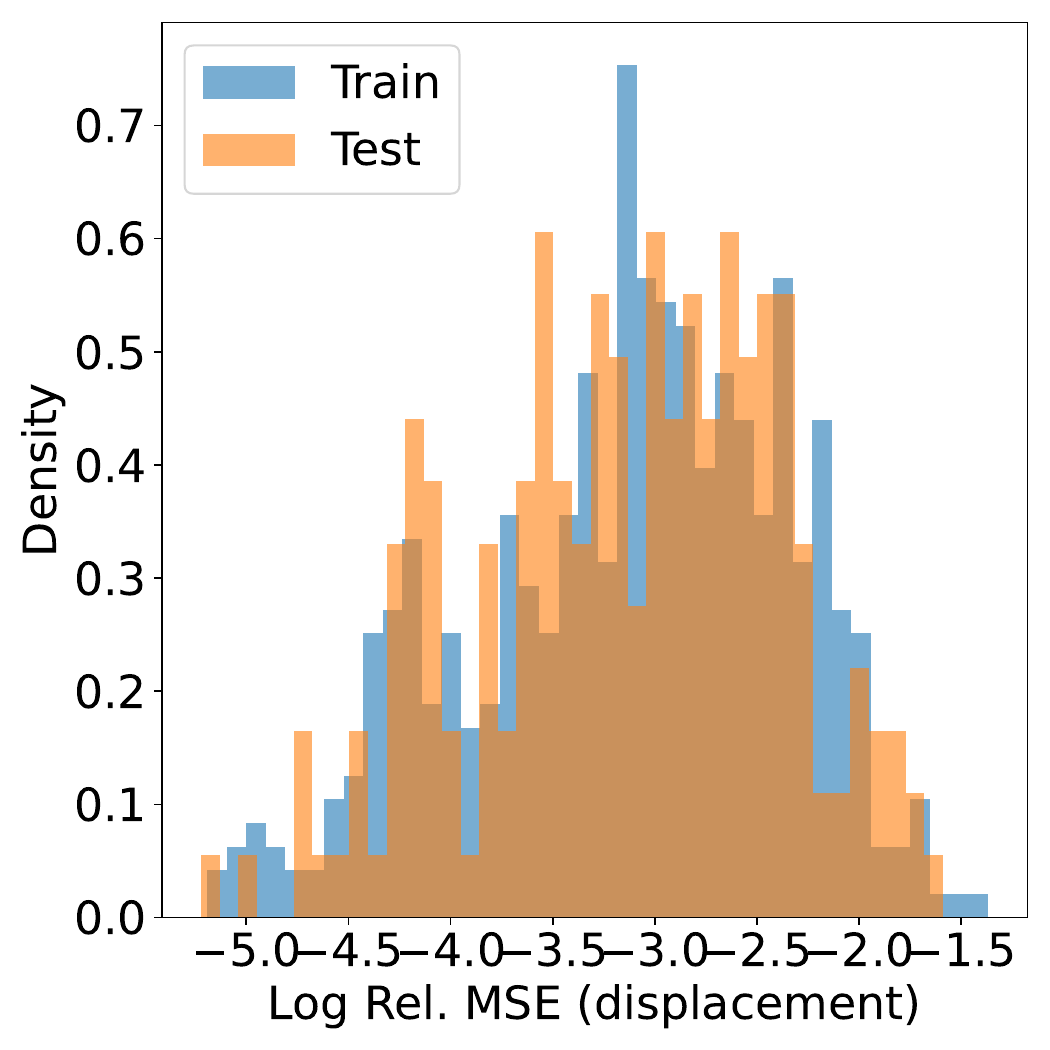}
        \caption{}
        \label{fig:biot_error_density_u}
    \end{subfigure} 
    \hspace{2.0pt}
    \begin{subfigure}[b]{0.32\textwidth}
        \centering
        \includegraphics[width=\textwidth]{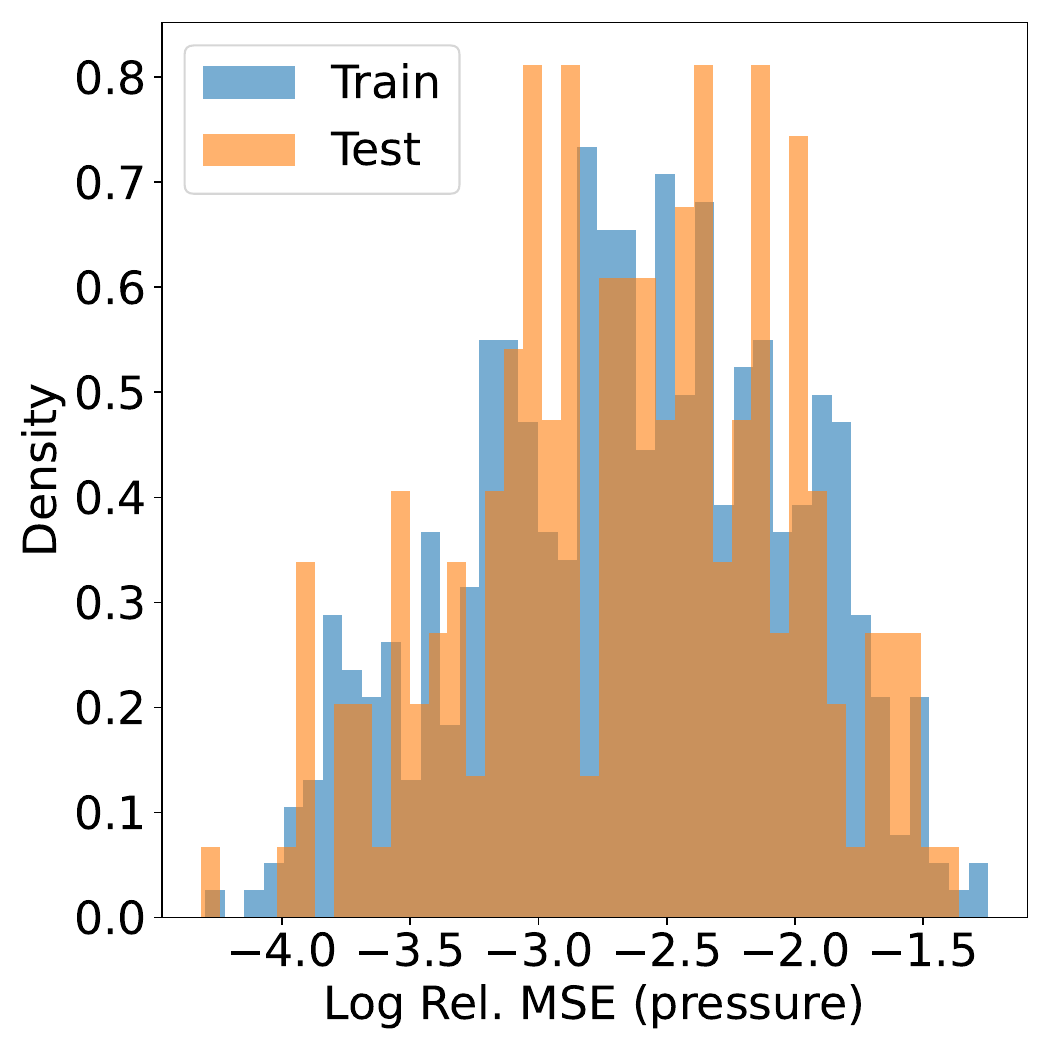}
        \caption{}
        \label{fig:biot_error_density_p}
    \end{subfigure} \\
    \caption{Biot consolidation: (a) Convergence profile of loss function during optimization iterations for the operator learning process; Distribution of output function prediction errors for the training and testing datasets after training for (b) displacements and (c) pressures.
    }\label{fig:biot_congergence_error_density}
\end{figure}

Table~\ref{table:error_statistics} presents the relative errors computed over the full training and testing datasets. 
For both, the mean and standard deviation of the errors are on the order of $\mathcal{O}(10^{-3})$. Figure~\ref{fig:biot_congergence_error_density}(a) shows the convergence during training for both the training and testing error, while Figure~\ref{fig:biot_congergence_error_density}(b, c) show histograms of the relative MSE for displacement and pressure, respectively after training.

\begin{figure}[!ht]
    \centering
    \begin{subfigure}[b]{0.35\textwidth}
        \centering
        \includegraphics[width=\textwidth]{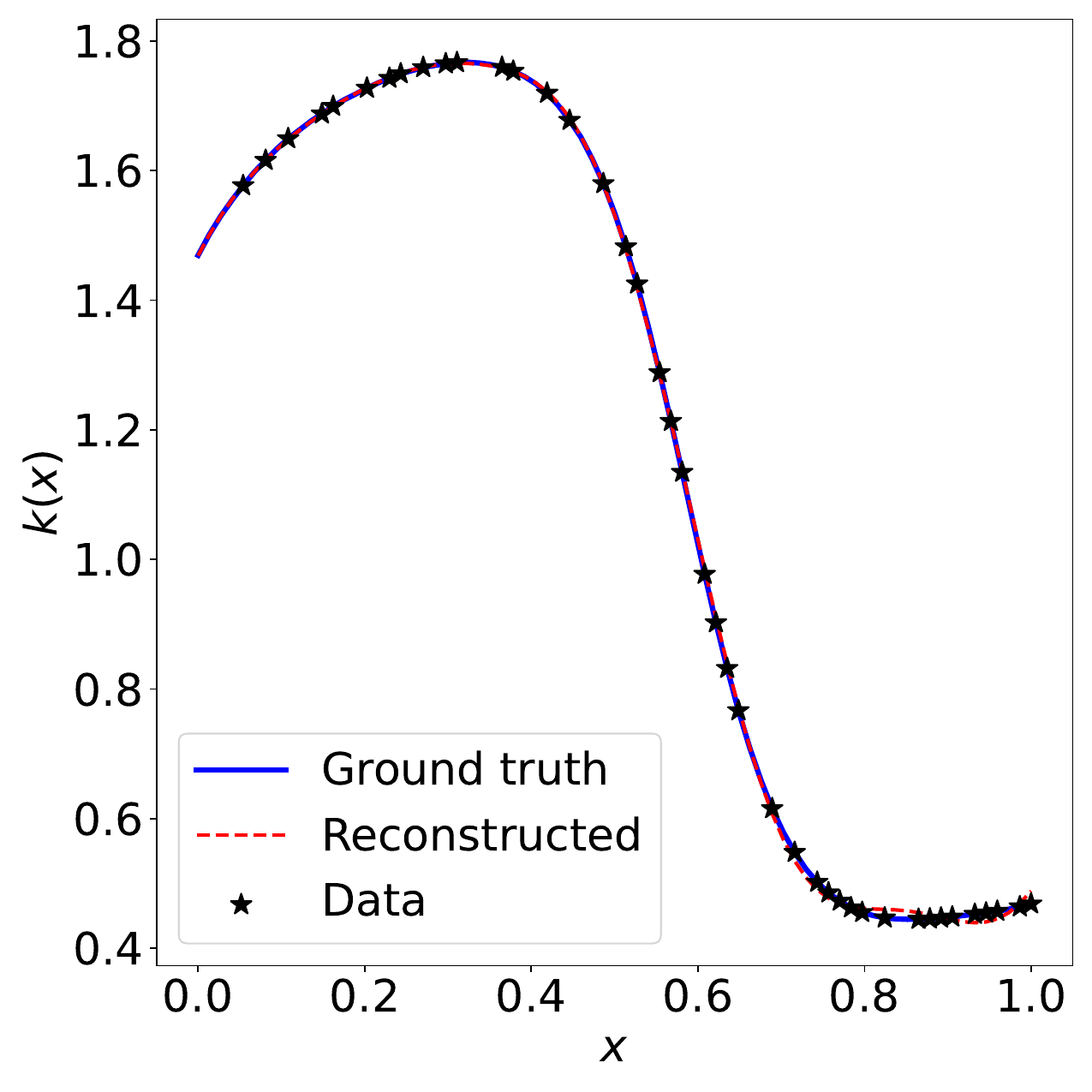}
        \caption{}
        \label{fig:biot_maxtest_k}
    \end{subfigure}\\
    \centering
    \begin{subfigure}[b]{0.22\textwidth}
        \centering
        \includegraphics[width=\textwidth]{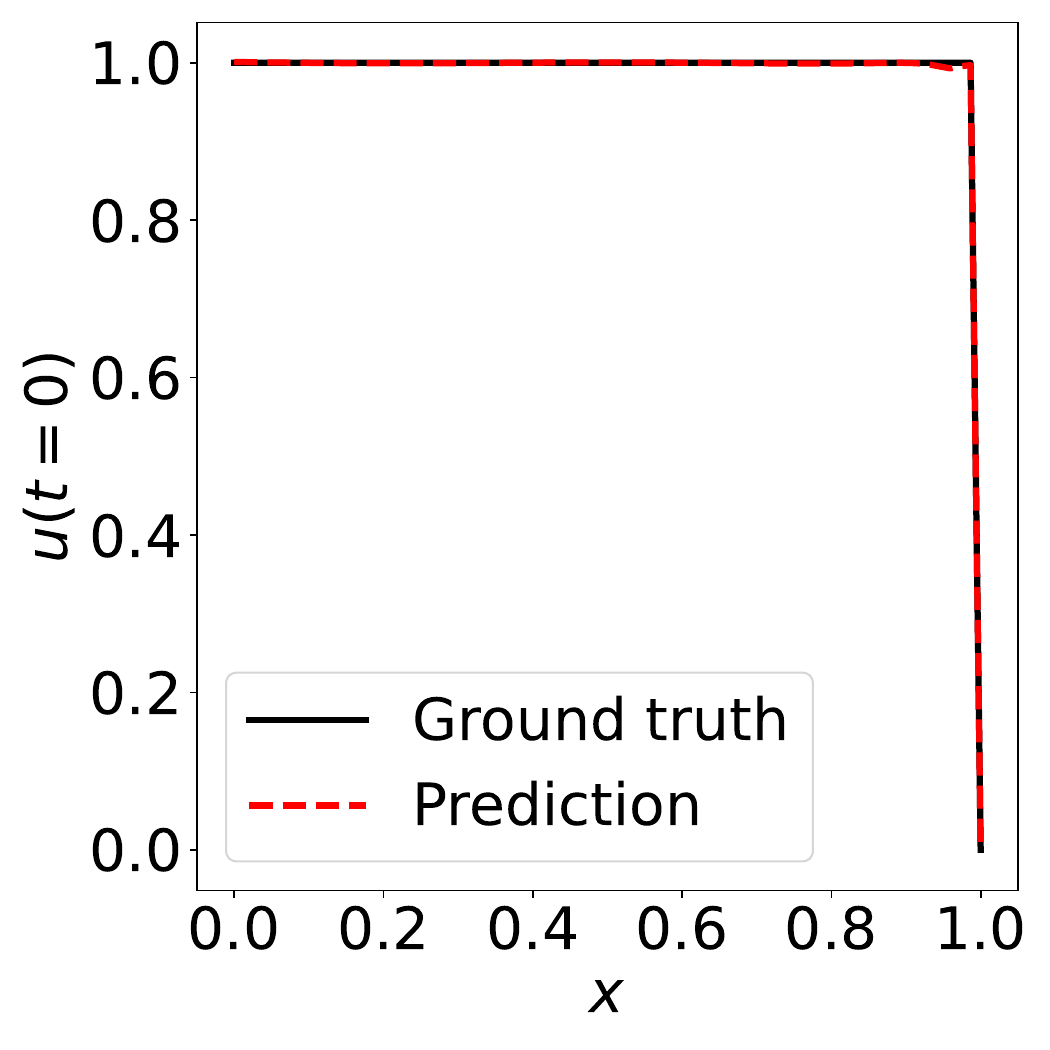}
        \caption{}
        \label{fig:biot_maxtest_pic}
    \end{subfigure}
    \hspace{1.0pt}
    \begin{subfigure}[b]{0.23\textwidth} 
        \centering
        \includegraphics[width=\textwidth]{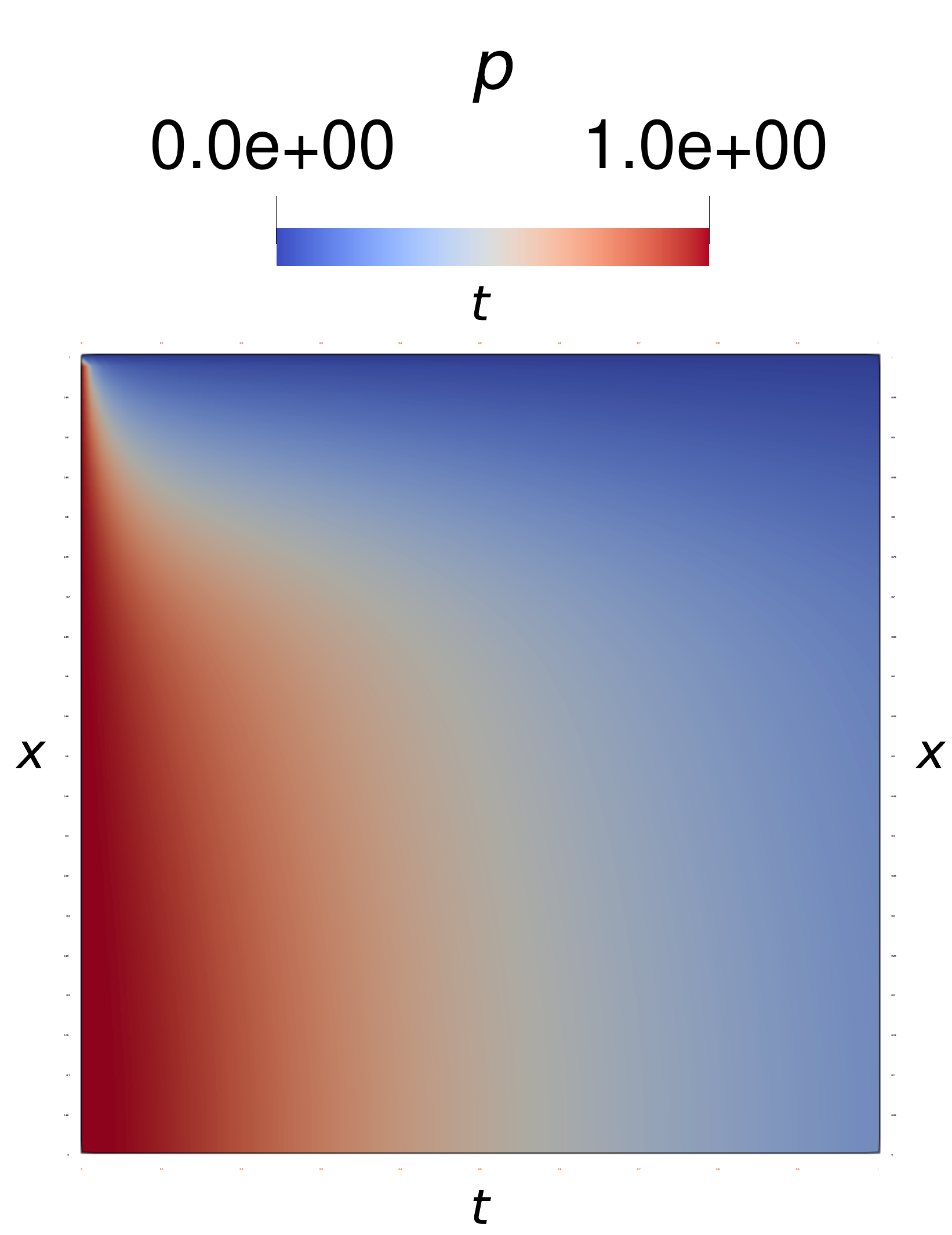}
        \caption{}
        \label{fig:biot_maxtest_p} 
    \end{subfigure}
    \hspace{1.0pt}
    \begin{subfigure}[b]{0.23\textwidth} 
        \centering
        \includegraphics[width=\textwidth]{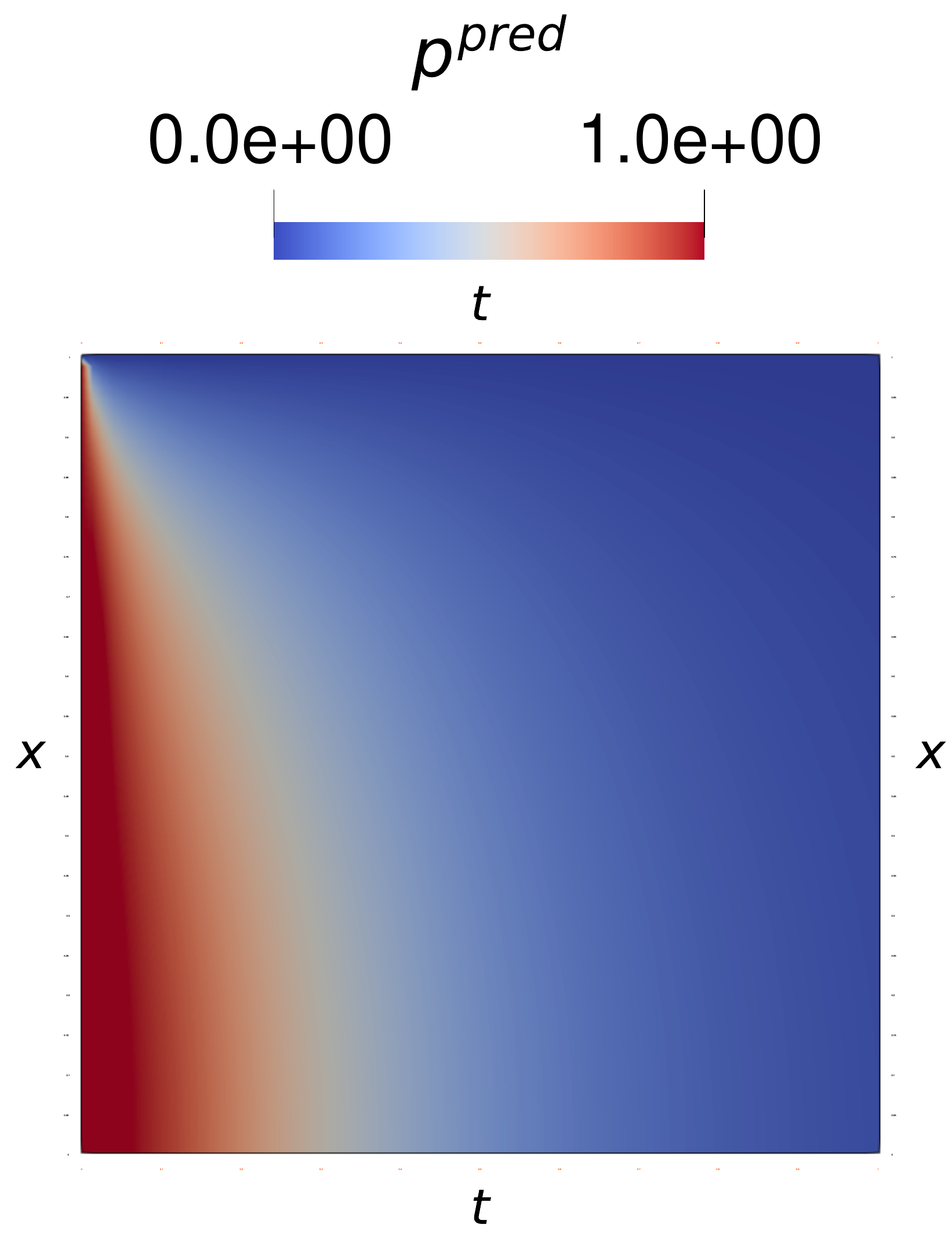}
        \caption{}
        \label{fig:biot_maxtest_ppred} 
    \end{subfigure}
    \hspace{1.0pt}
    \begin{subfigure}[b]{0.23\textwidth} 
        \centering
        \includegraphics[width=\textwidth]{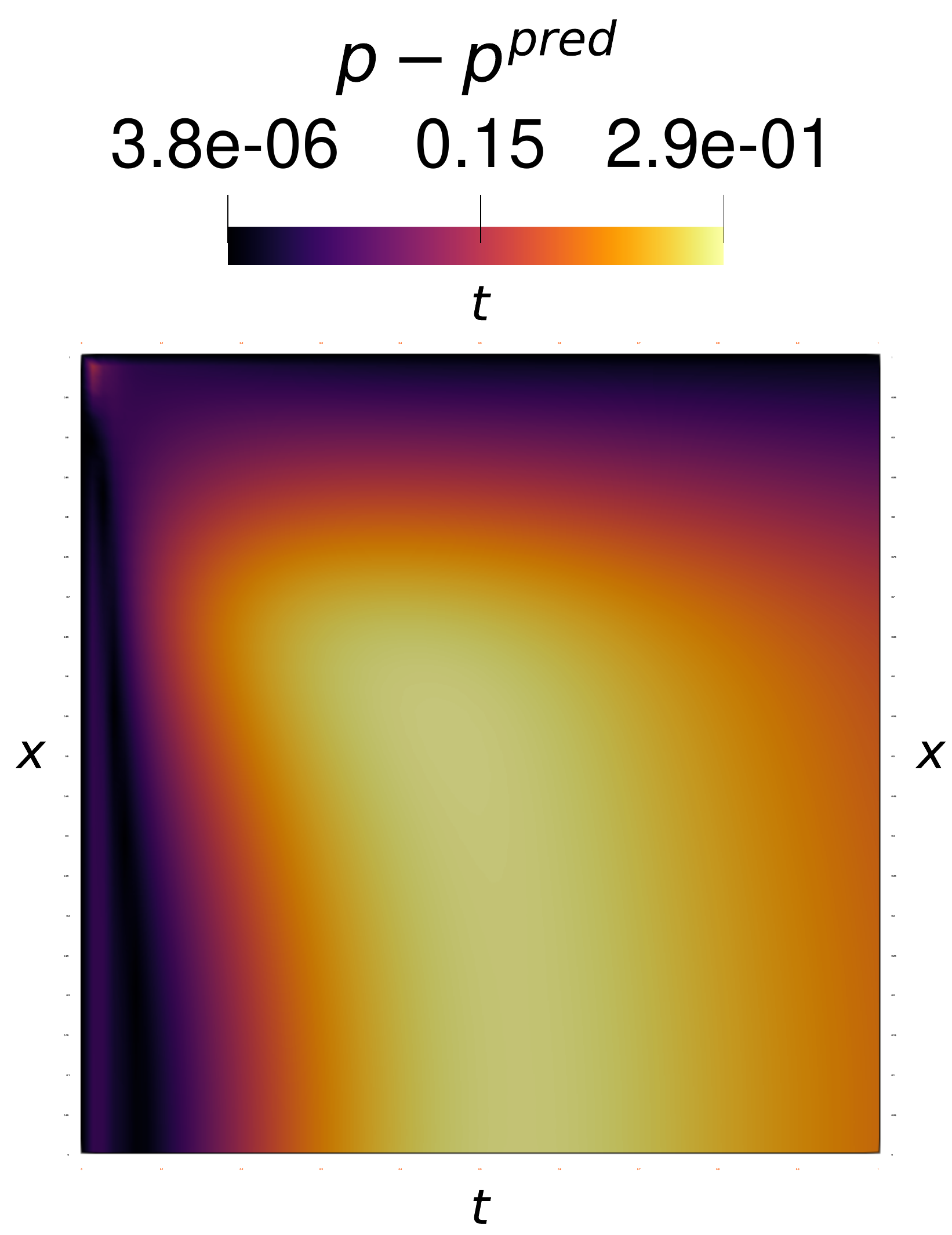}
        \caption{}
        \label{fig:biot_maxtest_error_p} 
    \end{subfigure}\\
    \centering
    \begin{subfigure}[b]{0.22\textwidth}
        \centering
        \includegraphics[width=\textwidth]{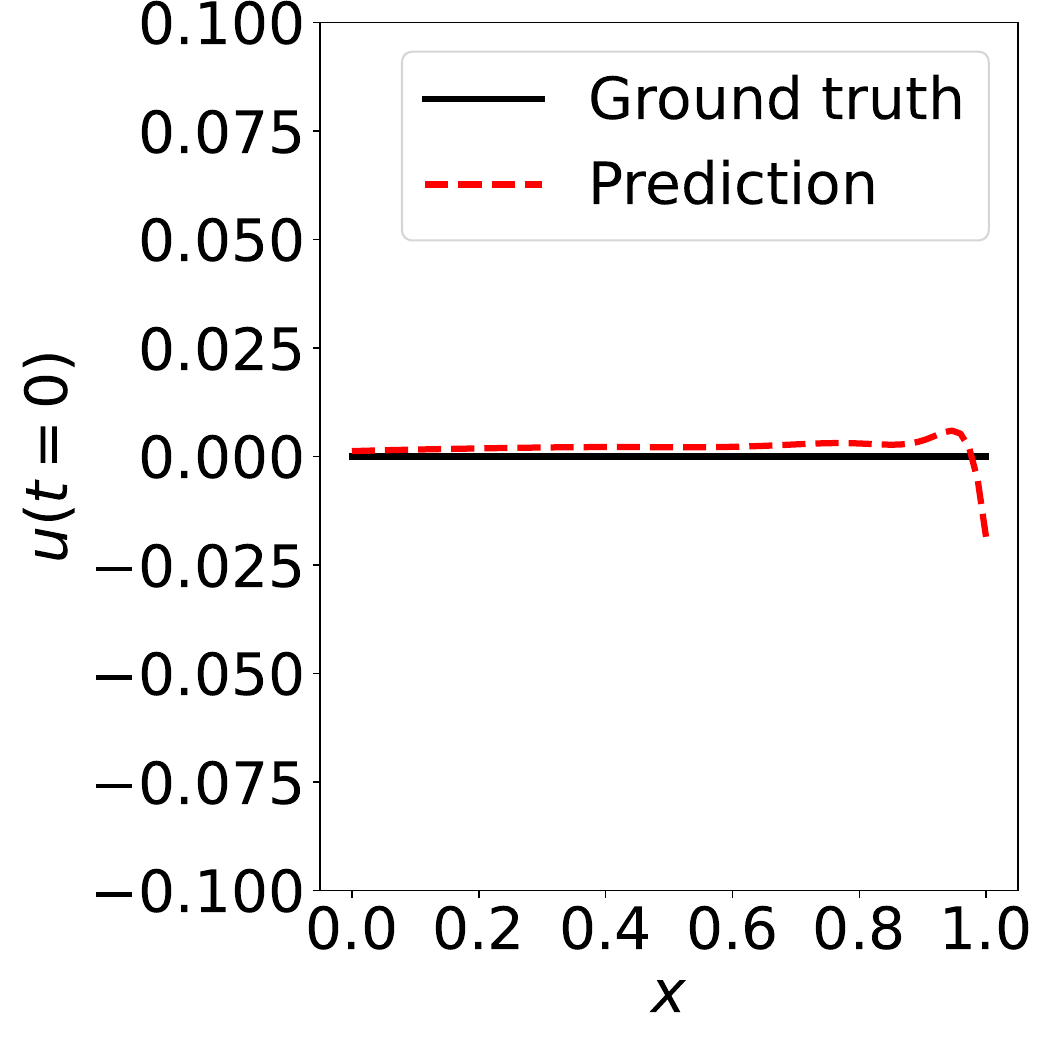}
        \caption{}
        \label{fig:biot_maxtest_uic}
    \end{subfigure}
    \hspace{1.0pt}
    \begin{subfigure}[b]{0.23\textwidth} 
        \centering
        \includegraphics[width=\textwidth]{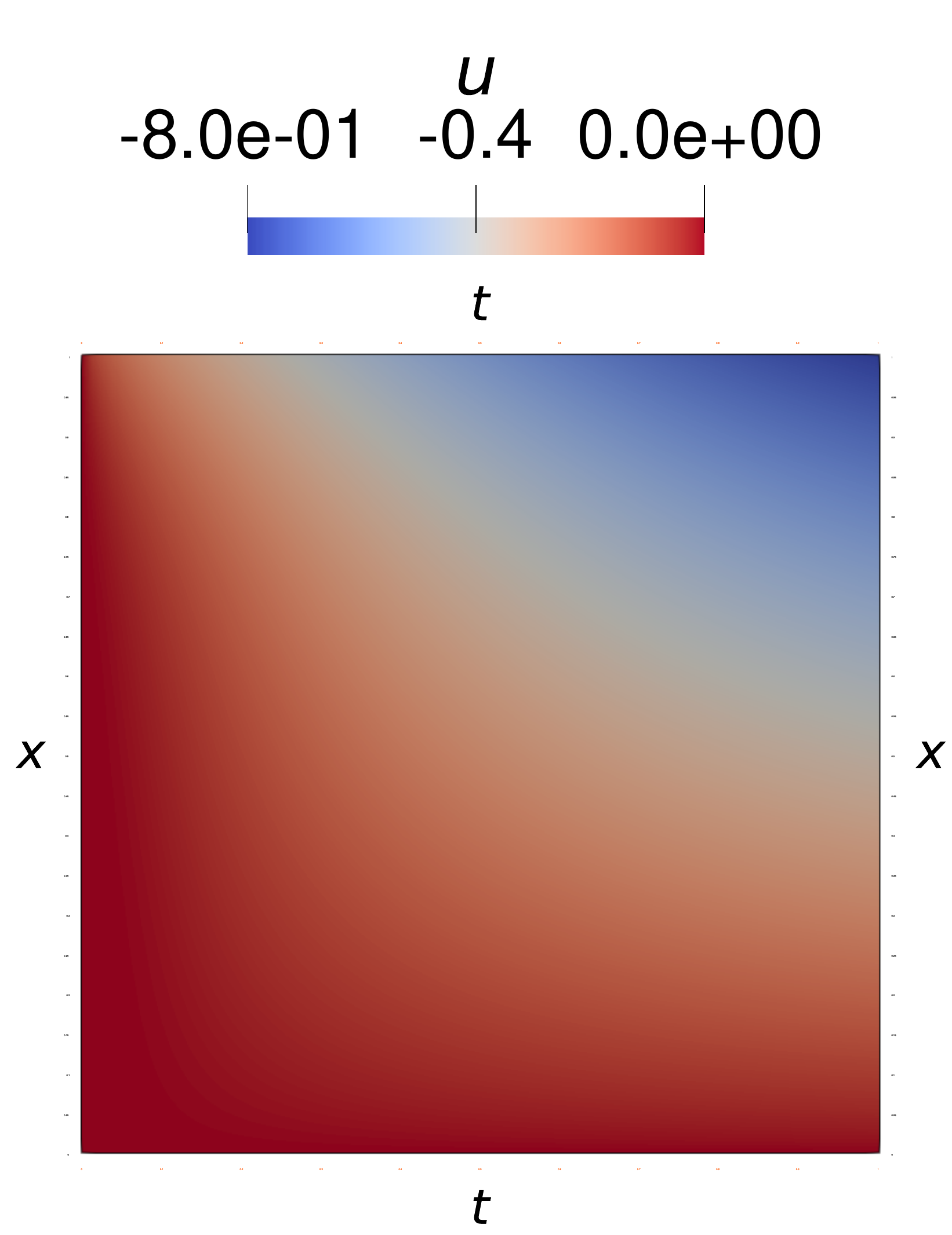}
        \caption{}
        \label{fig:biot_maxtest_u} 
    \end{subfigure}
    \hspace{1.0pt}
    \begin{subfigure}[b]{0.23\textwidth} 
        \centering
        \includegraphics[width=\textwidth]{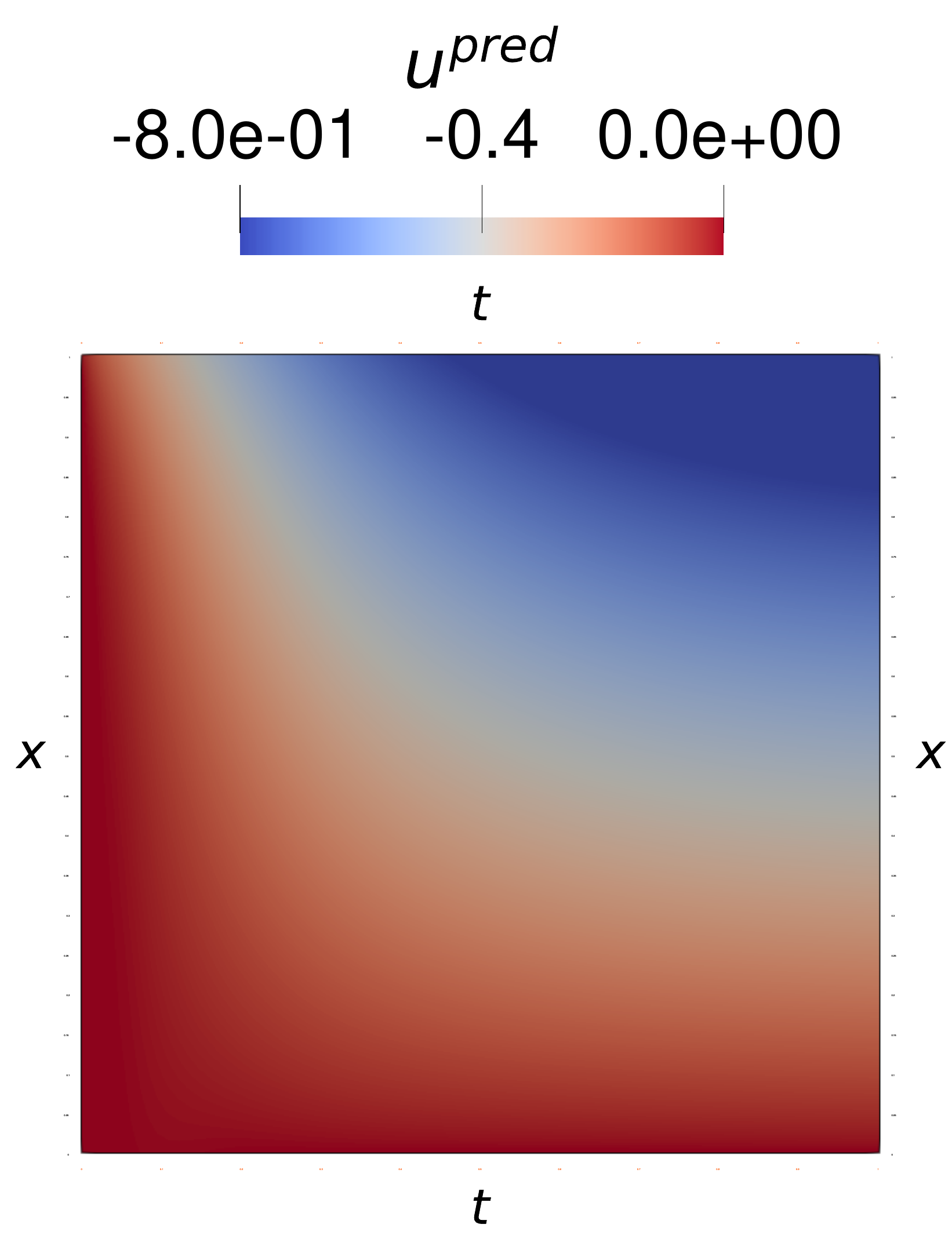}
        \caption{}
        \label{fig:biot_maxtest_upred} 
    \end{subfigure}
    \hspace{1.0pt}
    \begin{subfigure}[b]{0.23\textwidth} 
        \centering
        \includegraphics[width=\textwidth]{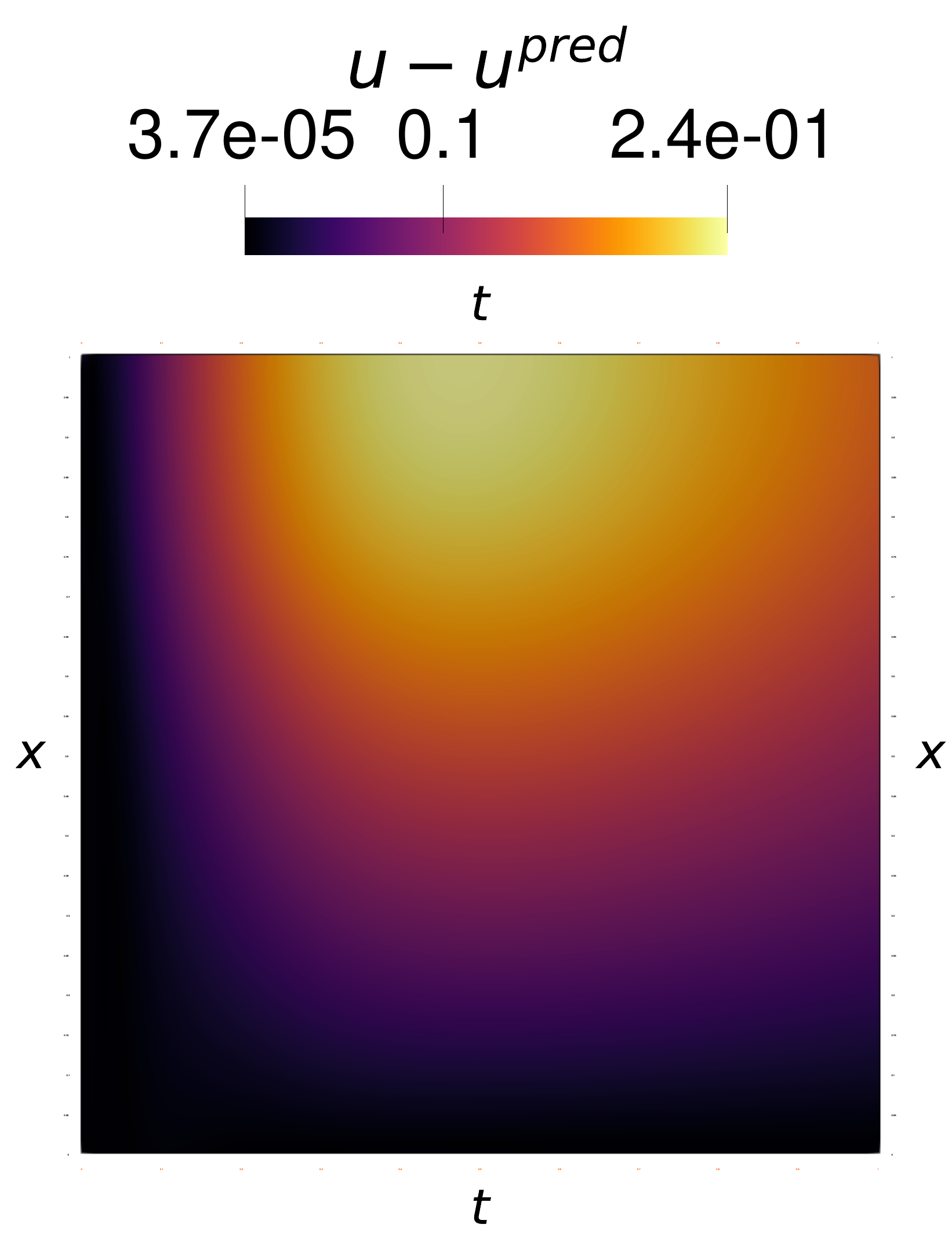}
        \caption{}
        \label{fig:biot_maxtest_error_u} 
    \end{subfigure}\\
    \caption{Biot's consolidation: PI-RINO predictions for worst-case sample with highest error in output function prediction from the test dataset. (a) The queried input function data are plotted as  `$\star$', along with the reconstructed function using the learned dictionary and compared with the ground truth. The middle row shows pressure predictions: (b) Initial pressure, (c) PI-RINO pressure prediction, (d) Ground truth pressure and (e) absolute error in pressure. The bottom row shows the displacement predictions: (f) Initial displacement, (g) PI-RINO displacement prediction, (h) Ground truth displacements and (i) absolute error in displacement.}
    \label{fig:biot_maxtest}
\end{figure}

\begin{figure}[!hbt]
    \centering
    \begin{subfigure}[b]{0.35\textwidth}
        \centering
        \includegraphics[width=\textwidth]{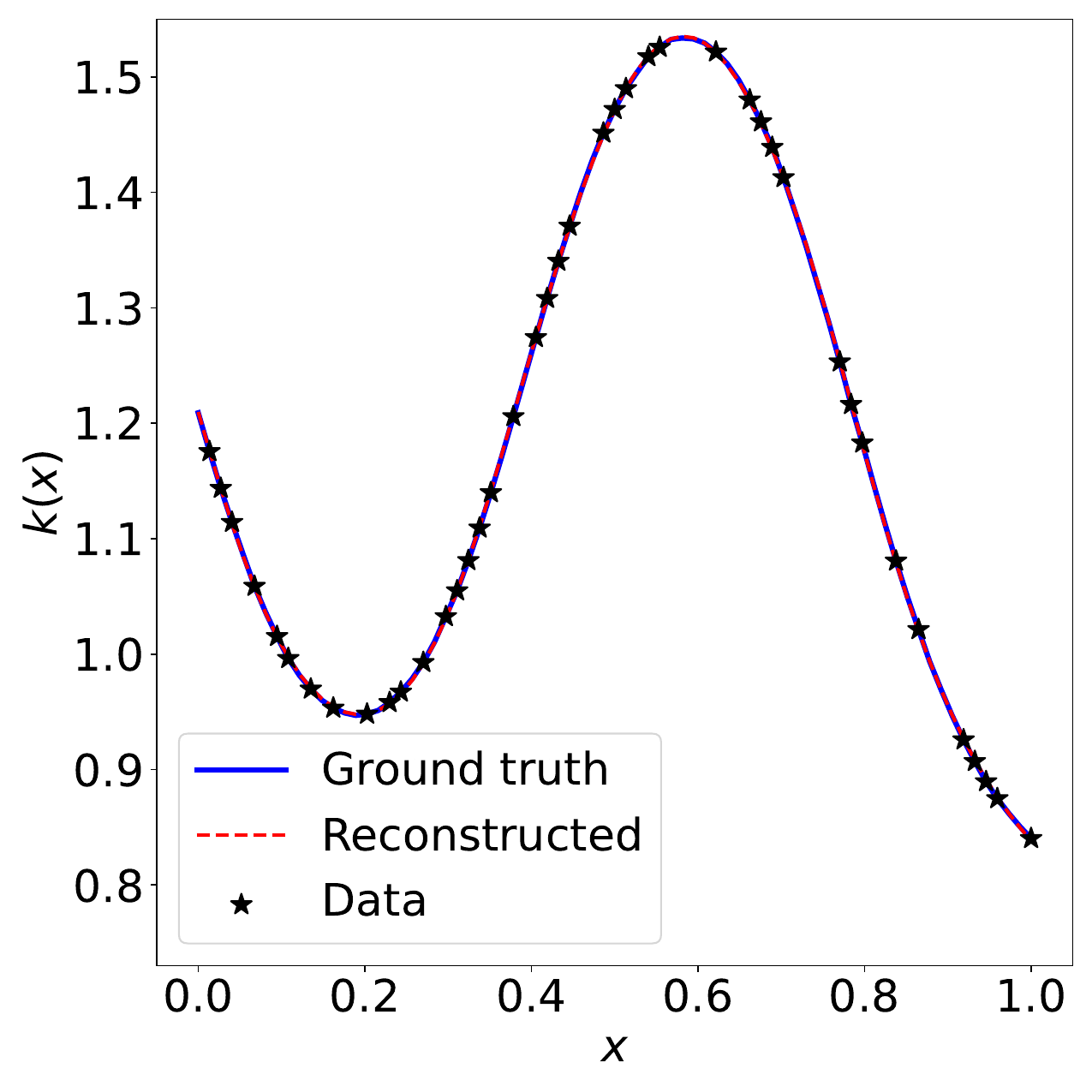}
        \caption{}
        \label{fig:biot_25pertest_k}
    \end{subfigure}\\
    \centering
    \begin{subfigure}[b]{0.22\textwidth}
        \centering
        \includegraphics[width=\textwidth]{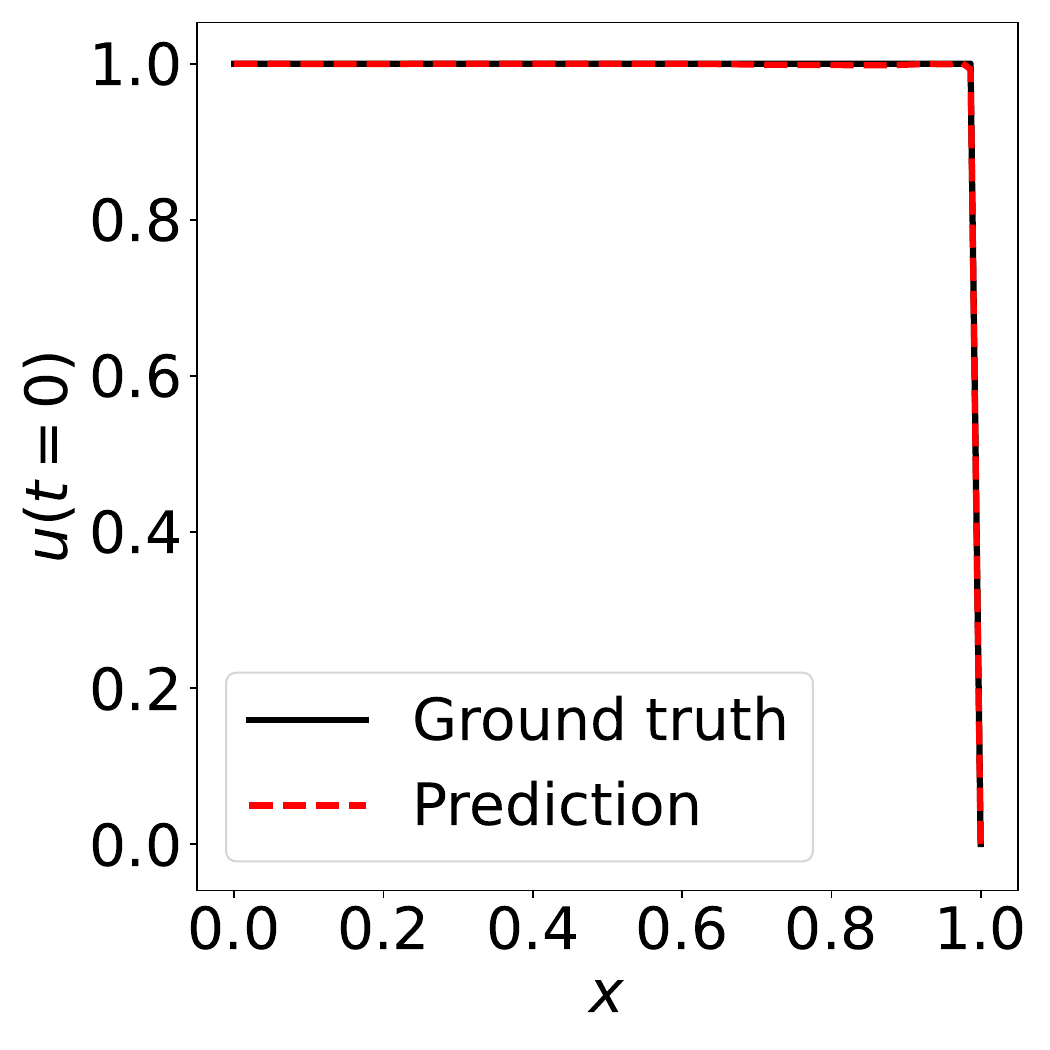}
        \caption{}
        \label{fig:biot_25pertest_pic}
    \end{subfigure}
    \hspace{1.0pt}
    \begin{subfigure}[b]{0.23\textwidth} 
        \centering
        \includegraphics[width=\textwidth]{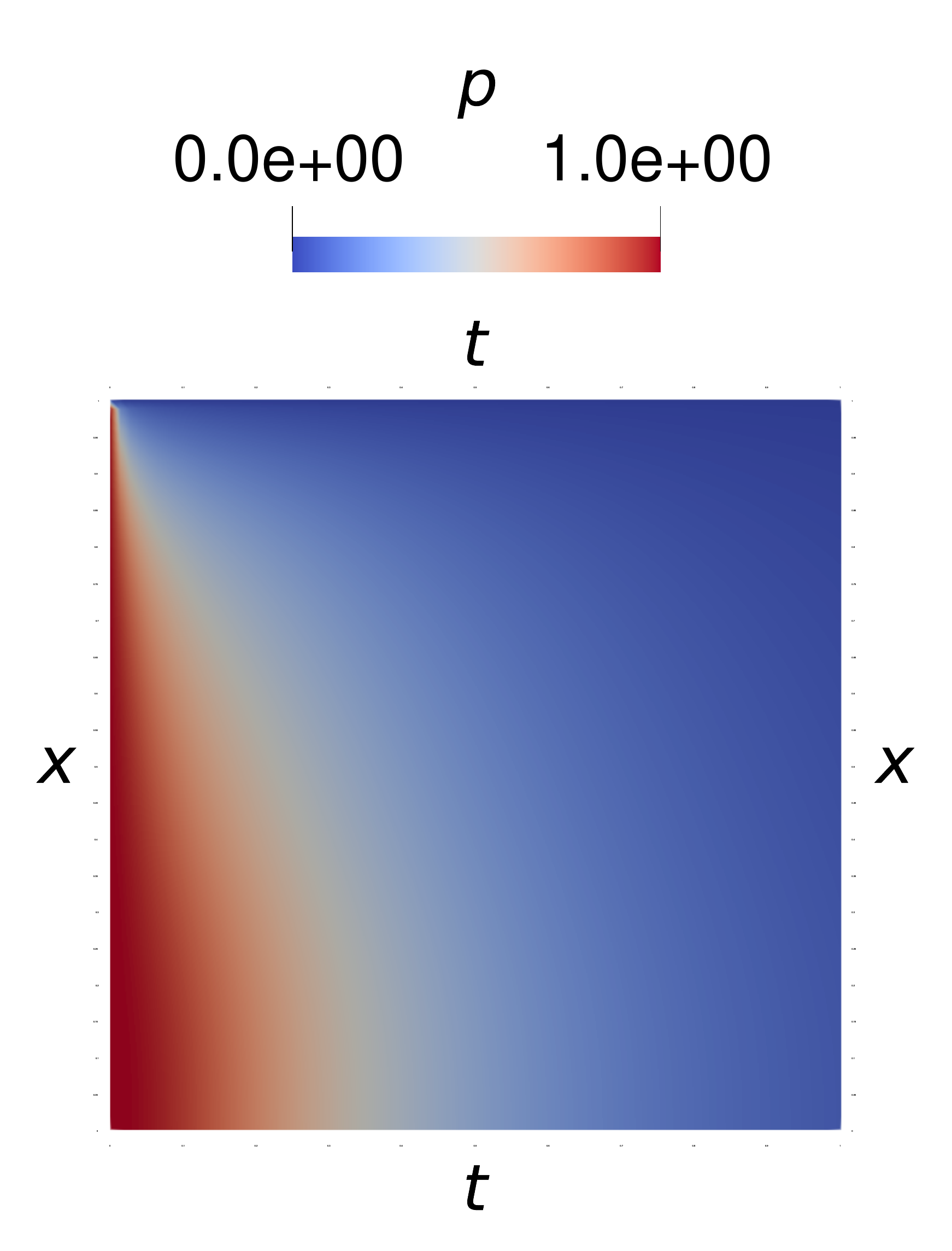}
        \caption{}
        \label{fig:biot_25pertest_p} 
    \end{subfigure}
    \hspace{1.0pt}
    \begin{subfigure}[b]{0.23\textwidth} 
        \centering
        \includegraphics[width=\textwidth]{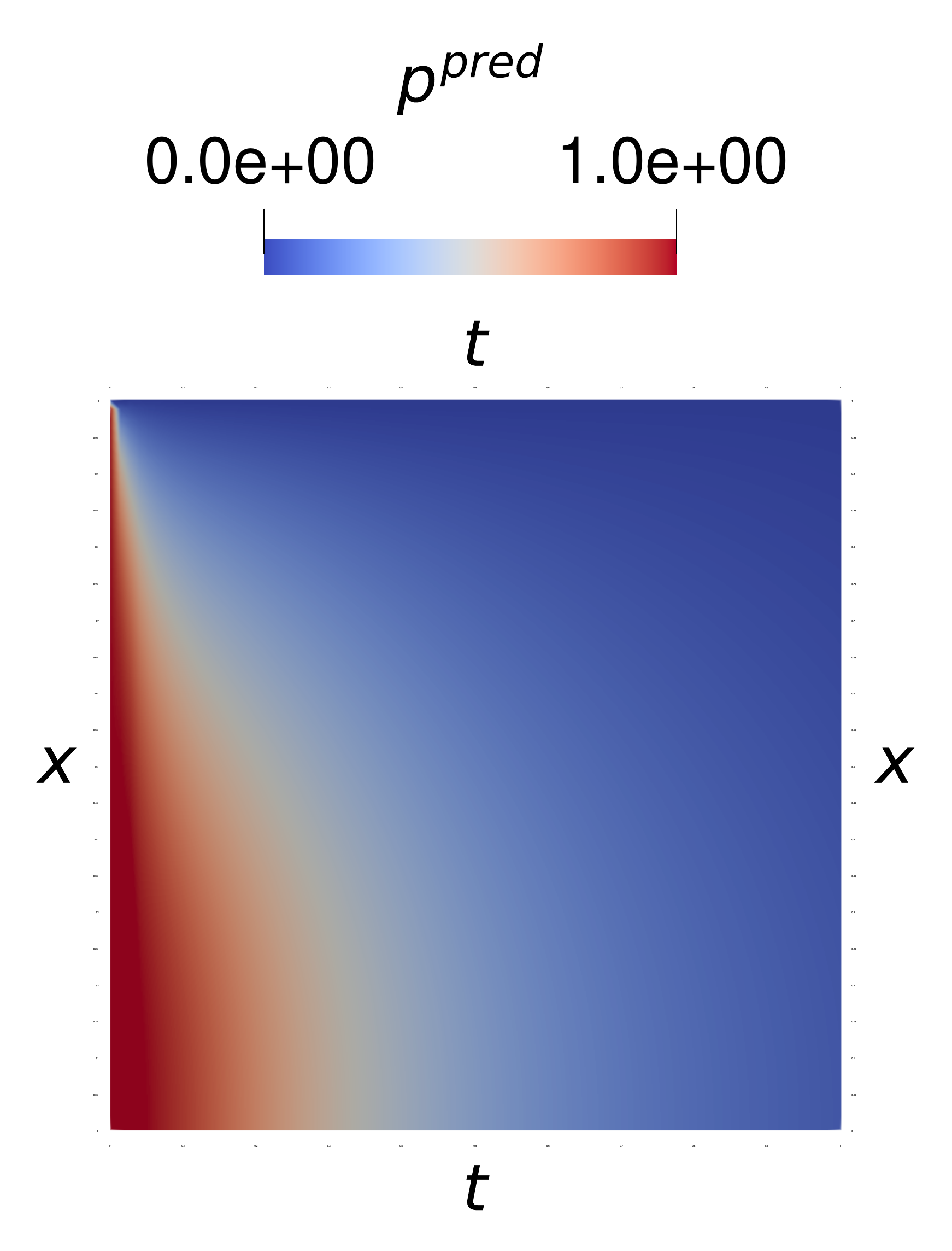}
        \caption{}
        \label{fig:biot_25pertest_ppred} 
    \end{subfigure}
    \hspace{1.0pt}
    \begin{subfigure}[b]{0.23\textwidth} 
        \centering
        \includegraphics[width=\textwidth]{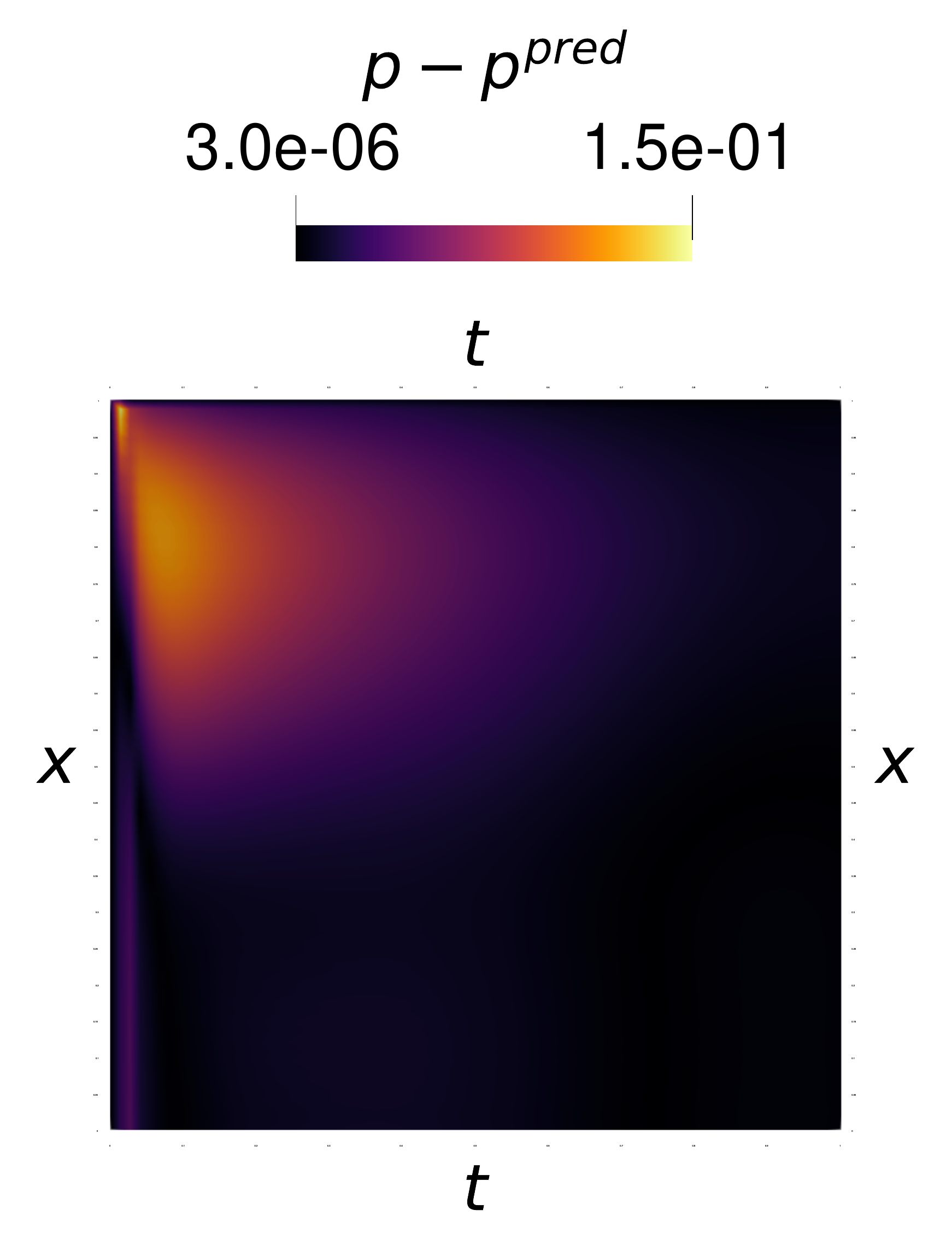}
        \caption{}
        \label{fig:biot_25pertest_error_p} 
    \end{subfigure}\\
    \centering
    \begin{subfigure}[b]{0.22\textwidth}
        \centering
        \includegraphics[width=\textwidth]{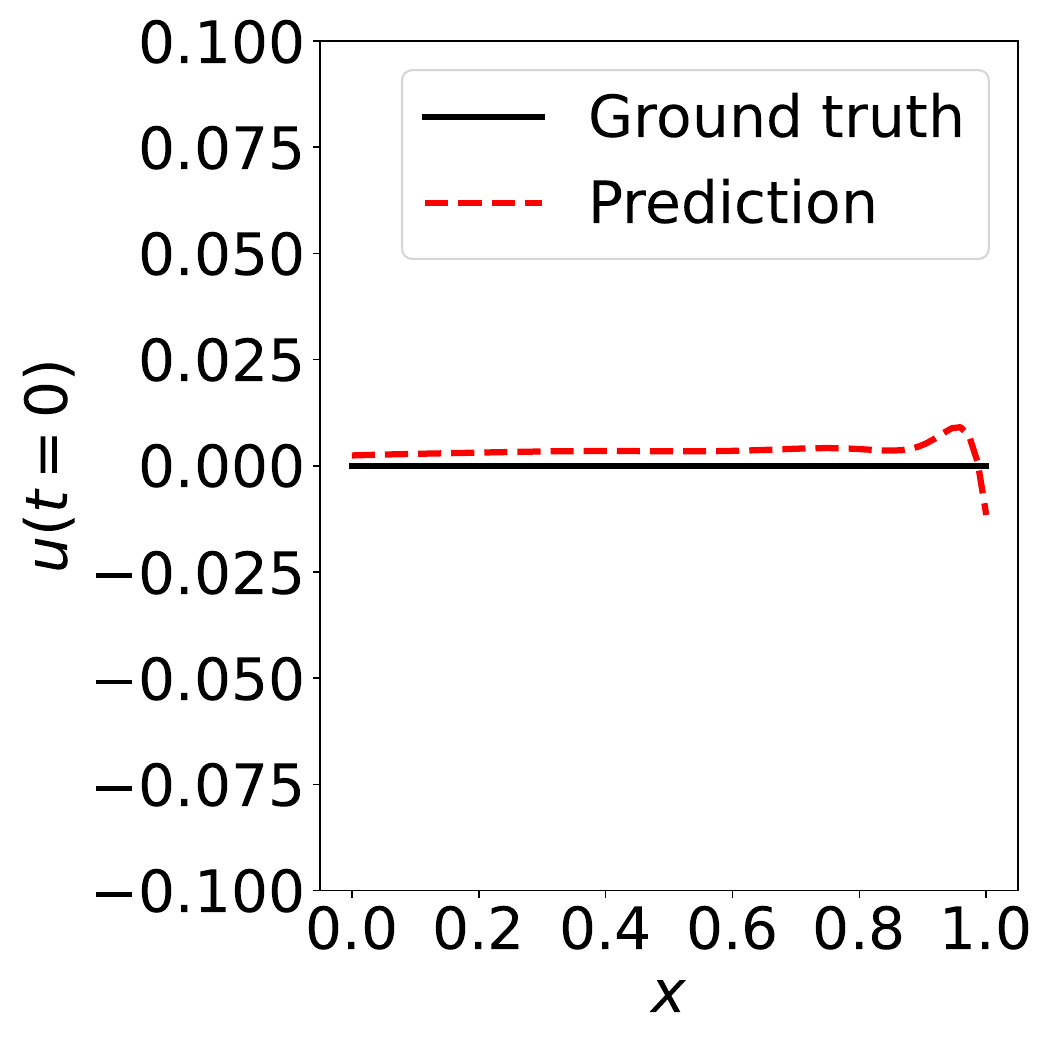}
        \caption{}
        \label{fig:biot_25pertest_uic}
    \end{subfigure}
    \hspace{1.0pt}
    \begin{subfigure}[b]{0.23\textwidth} 
        \centering
        \includegraphics[width=\textwidth]{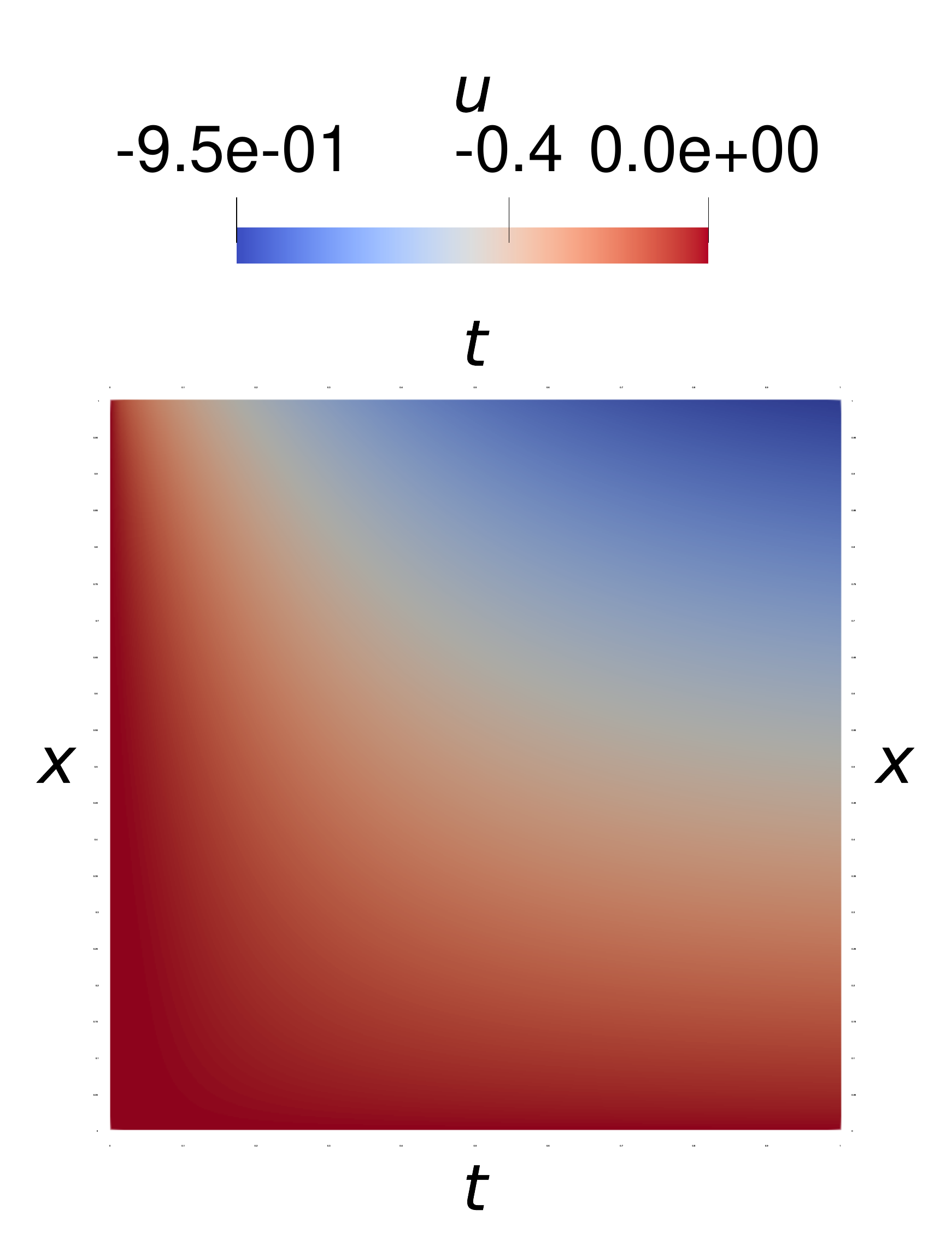}
        \caption{}
        \label{fig:biot_25pertest_u} 
    \end{subfigure}
    \hspace{1.0pt}
    \begin{subfigure}[b]{0.23\textwidth} 
        \centering
        \includegraphics[width=\textwidth]{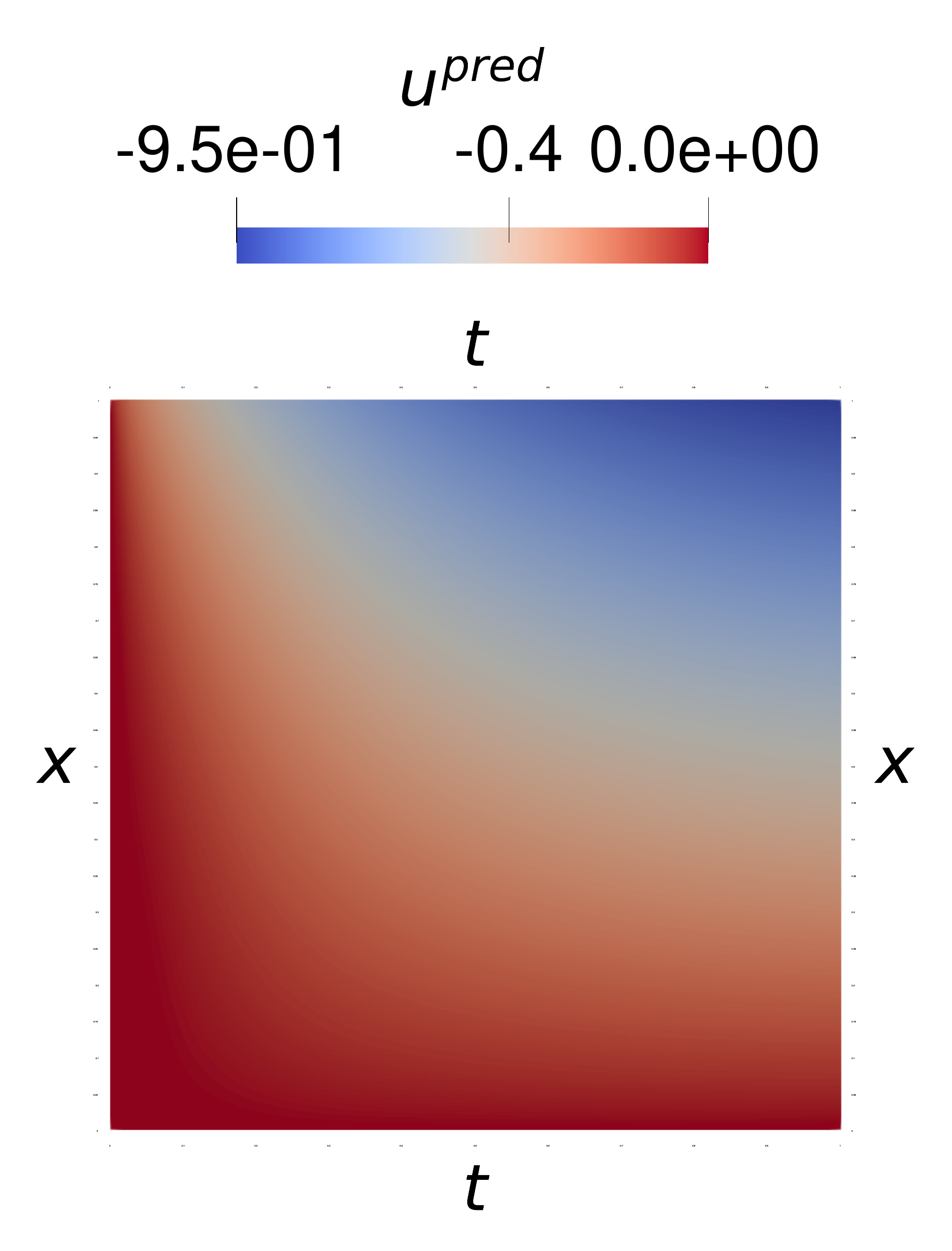}
        \caption{}
        \label{fig:biot_25pertest_upred} 
    \end{subfigure}
    \hspace{1.0pt}
    \begin{subfigure}[b]{0.23\textwidth} 
        \centering
        \includegraphics[width=\textwidth]{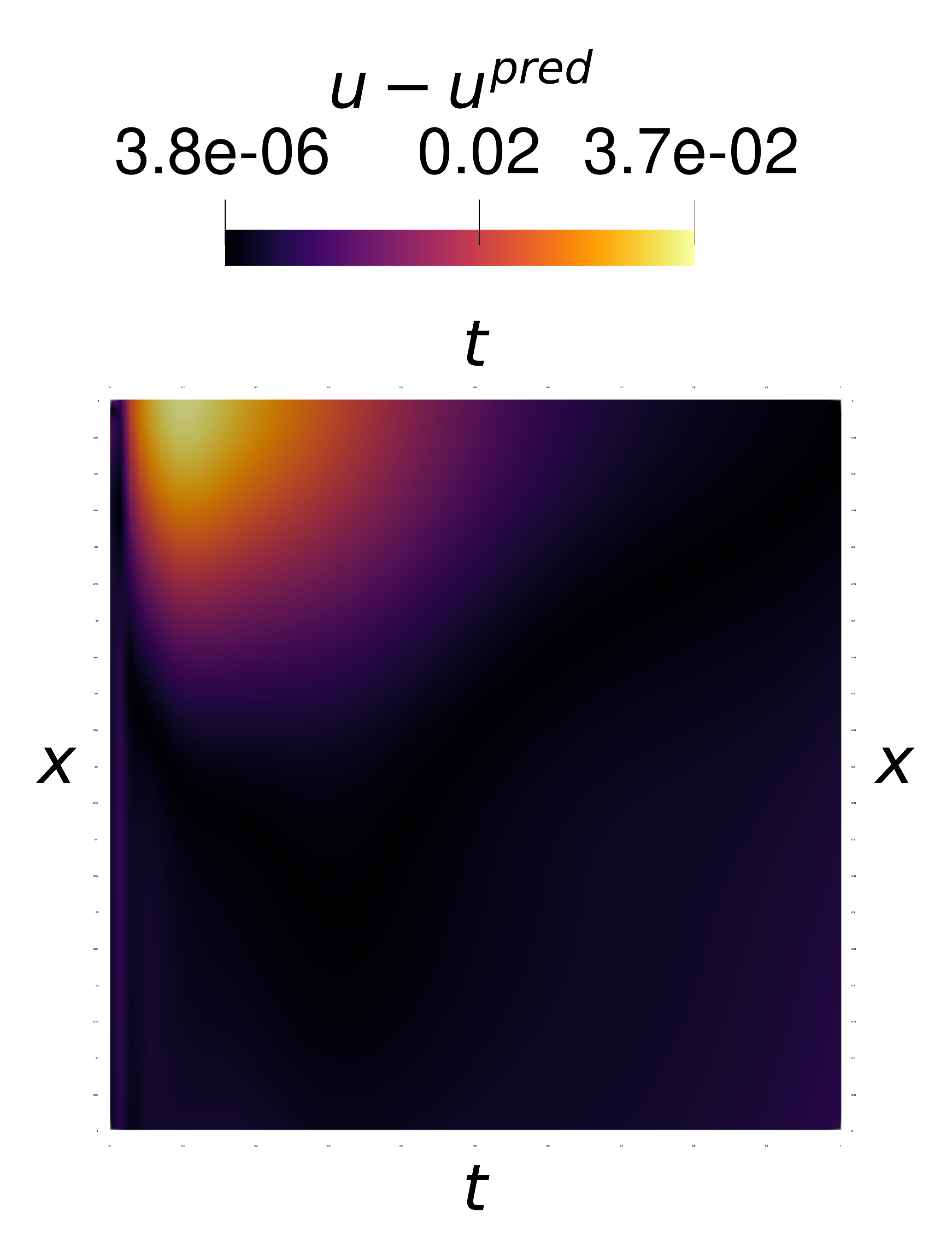}
        \caption{}
        \label{fig:biot_25pertest_error_u} 
    \end{subfigure}\\
    \caption{Biot consolidation: PI-RINO predictions for Q1 sample with 25th percentile of the error distribution (i.e., 75\% of cases have lower error than this sample) from the test dataset. (a) The queried input function data are plotted as `$\star$', along with the reconstructed function using the learned dictionary and compared with the ground truth. The middle row shows pressure predictions: (b) Initial pressure, (c) PI-RINO pressure prediction, (d) Ground truth pressure and (e) absolute error in pressure. The bottom row shows the displacement predictions: (f) Initial displacement, (g) PI-RINO displacement prediction, (h) Ground truth displacements and (i) absolute error in displacement.}
    \label{fig:biot_25pertest}
\end{figure}

Figures~\ref{fig:biot_maxtest} and~\ref{fig:biot_25pertest} illustrate the PI-RINO predictions for two representative test samples. Figure~\ref{fig:biot_maxtest} corresponds to the highest error in the test set and Figure~\ref{fig:biot_25pertest} corresponds to the 25th percentile (Q1) of the test error distribution, meaning that 75\% of test samples have lower error. In both cases, the top plot (a) shows the input function $\kappa(x)$ sampled at discrete points `$\star$', along with its full-field reconstruction $\tilde{\kappa}(x)$ from the dictionary learning approach compared to the ground truth. The second (b--e) and third (f--i) rows show the predicted pressure and displacement fields, respectively. While the worst-case sample exhibits relatively high prediction errors (up to 30\% for pressure and 31\% for displacement), the Q1 sample shows significantly lower errors, no more than 15\% for pressure and 4\% for displacement,indicating robust average performance.

\section{Comparison with Automatic Differentiation}\label{sec:autodiff_compare}

In this section, we compare the performance of the proposed framework, which employs a numerical method (finite differences) to enforce the physics loss, to the traditional approach that relies on automatic differentiation (Autodiff). To ensure a fair comparison, both models use the same network architecture and hyperparameters; the only difference lies in the method used to enforce the PDE constraints: finite difference versus Autodiff. We aimed to keep these implementations as close as possible, using the same platform (Python, with both based on PyTorch) and trained on an Apple M3 Max CPU. We report the computational gain factor `Cost', defined as the ratio of the time spent using the Autodiff-based method to the time spent using the FD-based method; for example, $\text{Cost}=2$ the Autodiff-based method takes twice as long to train using identical resources.

\begin{figure}[!hbt]
    \centering
    \begin{subfigure}[b]{0.45\textwidth}
        \centering
        \includegraphics[width=0.85\linewidth]{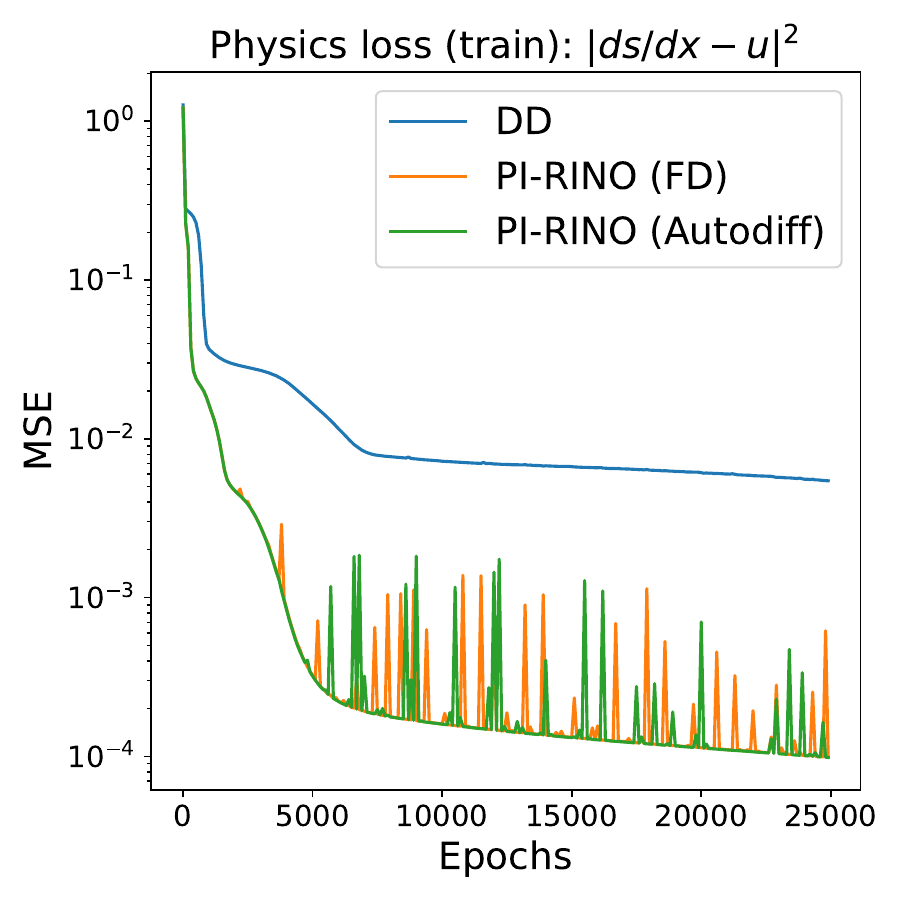}
        \caption{}
        \label{fig:1d_physics_compare_fd_auto_physics_loss_train}
    \end{subfigure}
    \begin{subfigure}[b]{0.45\textwidth}
        \centering
        \includegraphics[width=0.85\linewidth]{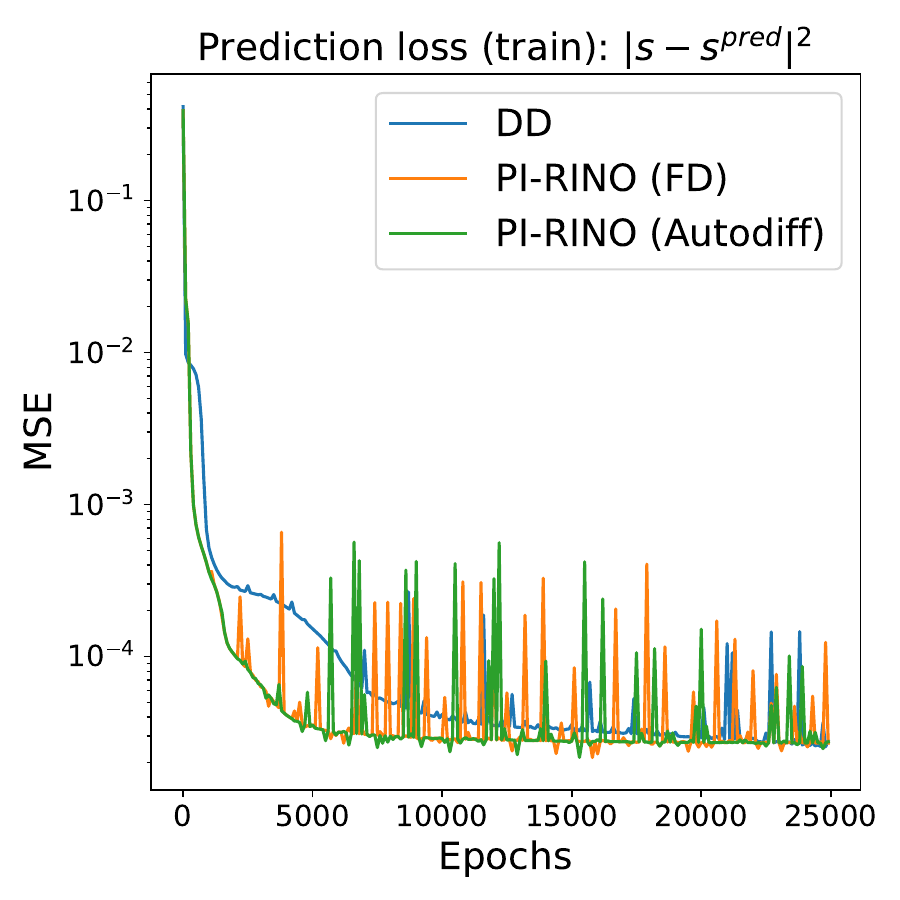}
        \caption{}
        \label{fig:1d_physics_compare_fd_auto_pred_loss_train}
    \end{subfigure}
    \begin{subfigure}[b]{0.45\textwidth}
        \centering
        \includegraphics[width=0.85\linewidth]{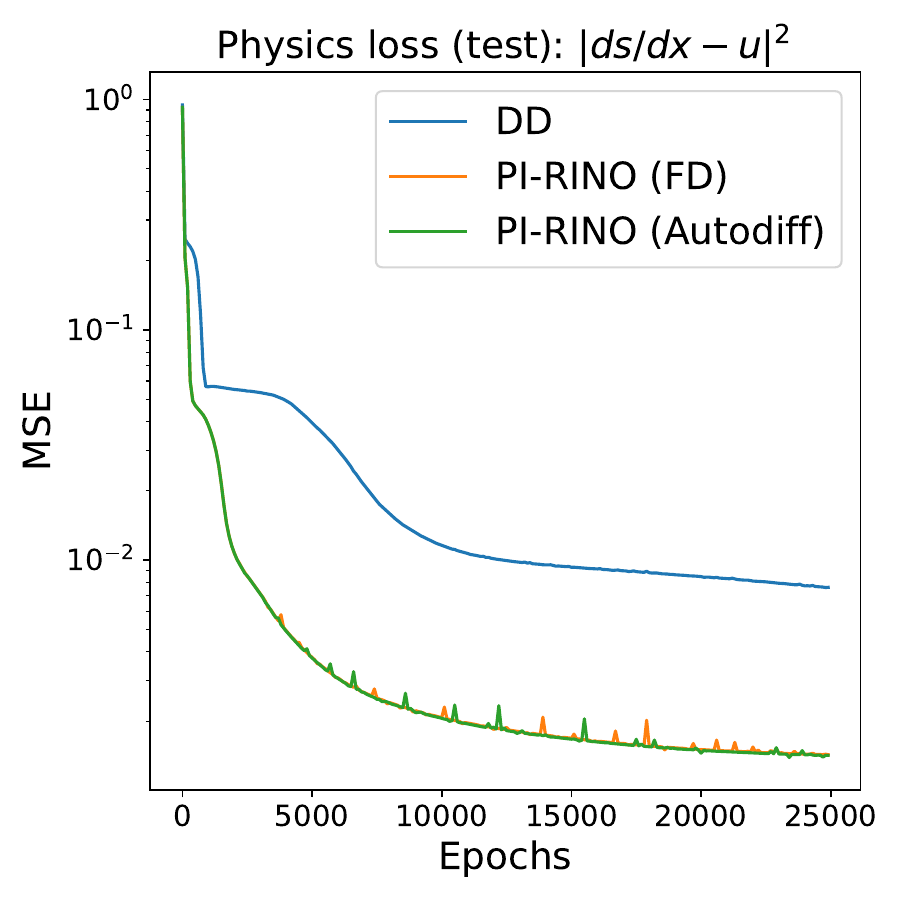}
        \caption{}
        \label{fig:1d_physics_compare_fd_auto_physics_loss_test}
    \end{subfigure}
    \begin{subfigure}[b]{0.45\textwidth}
        \centering
        \includegraphics[width=0.85\linewidth]{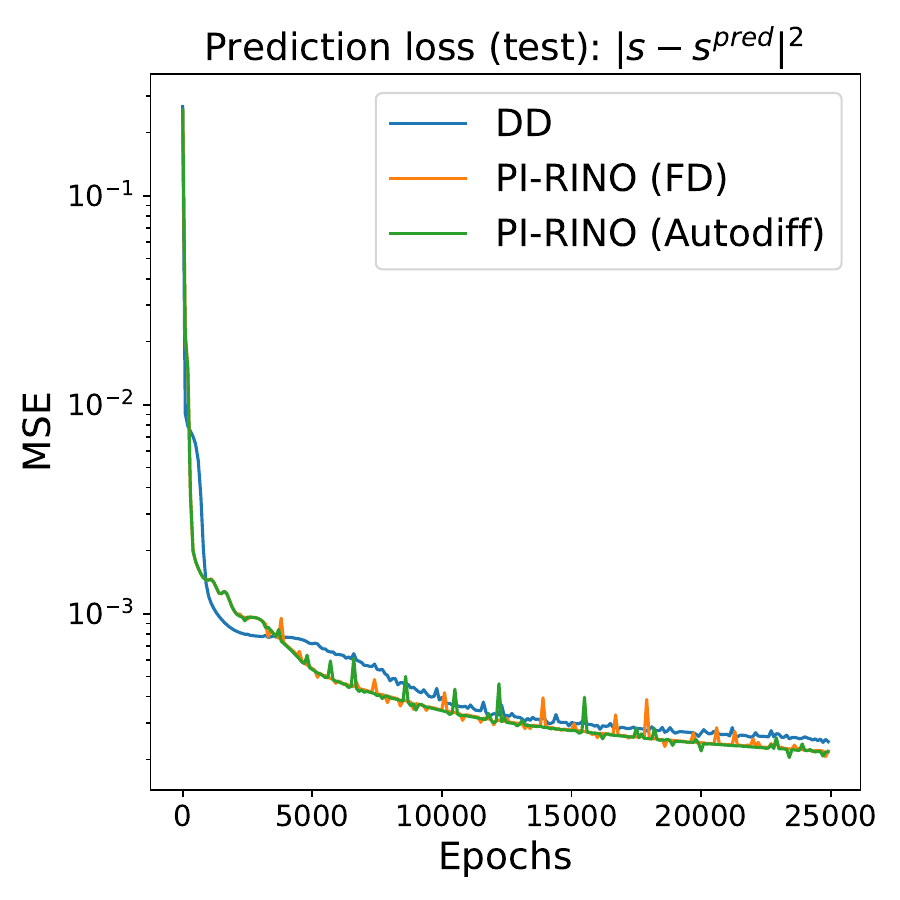}
        \caption{}
        \label{fig:1d_physics_compare_fd_auto_pred_loss_test}
    \end{subfigure}
    \begin{subfigure}[b]{0.45\textwidth}
        \centering
        \includegraphics[width=0.85\linewidth]{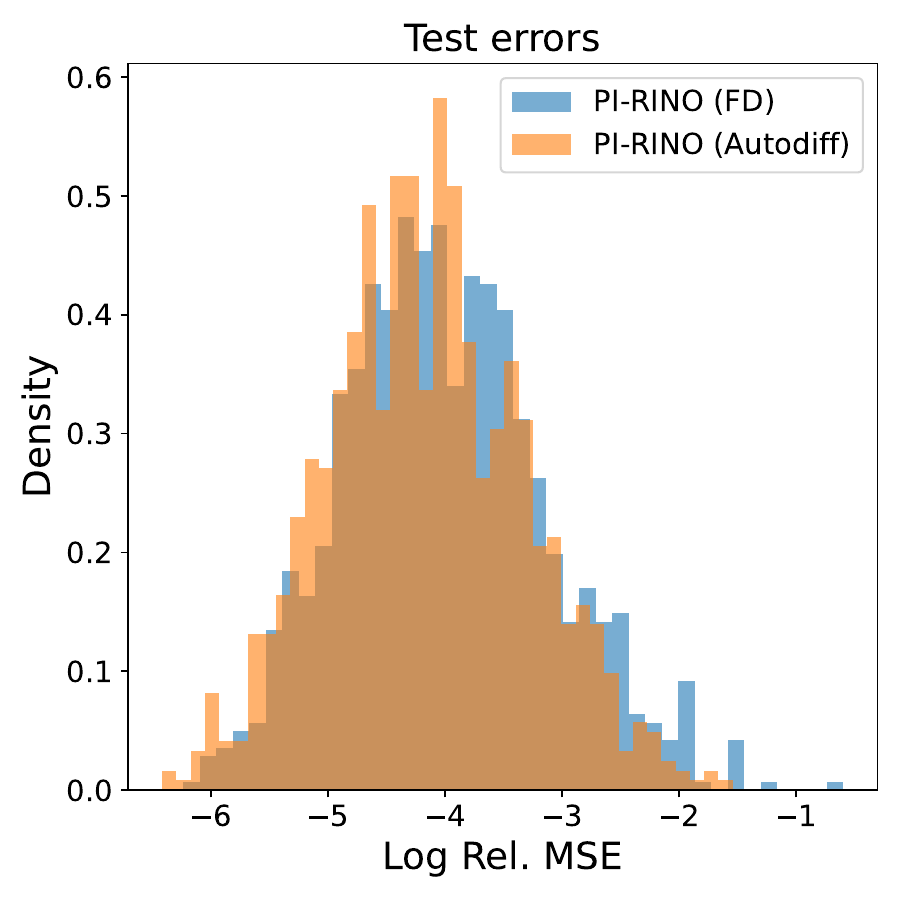}
        \caption{}
        \label{fig:1d_density_compare_fd_auto}
    \end{subfigure}

    \caption{Antiderivative: Comparison of the two PI-RINO frameworks—one using automatic differentiation (Autodiff) and the other using finite differences (FD) for physics enforcement—with a data-driven (DD) baseline - convergence of physics and prediction losses for training dataset (a and b), testing dataset (c and d); and distribution of output prediction errors (test) for the PI-RINO models (e).}
    \label{fig:1d_compare_fd_auto_convergence}
\end{figure}

\begin{figure}[!hbt]
    \centering
    \begin{subfigure}[b]{0.45\textwidth}
        \centering
        \includegraphics[width=0.85\linewidth]{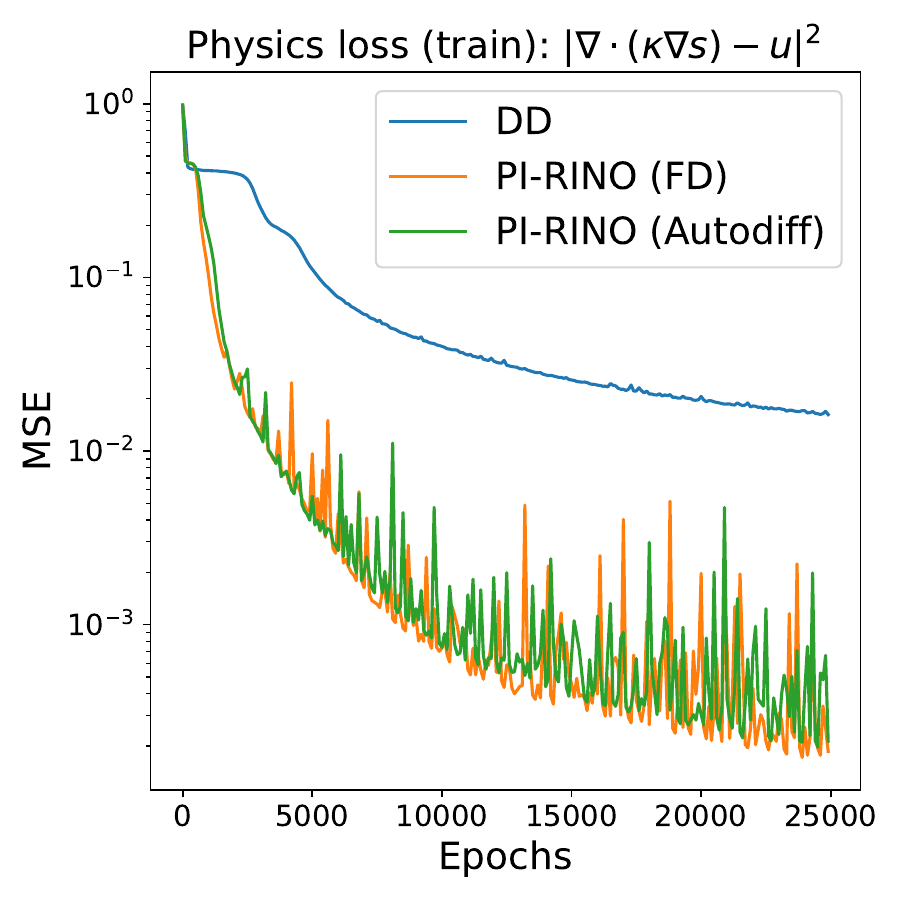}
        \caption{}
        \label{fig:2d_physics_compare_fd_auto_physics_loss_train}
    \end{subfigure}
    \begin{subfigure}[b]{0.45\textwidth}
        \centering
        \includegraphics[width=0.85\linewidth]{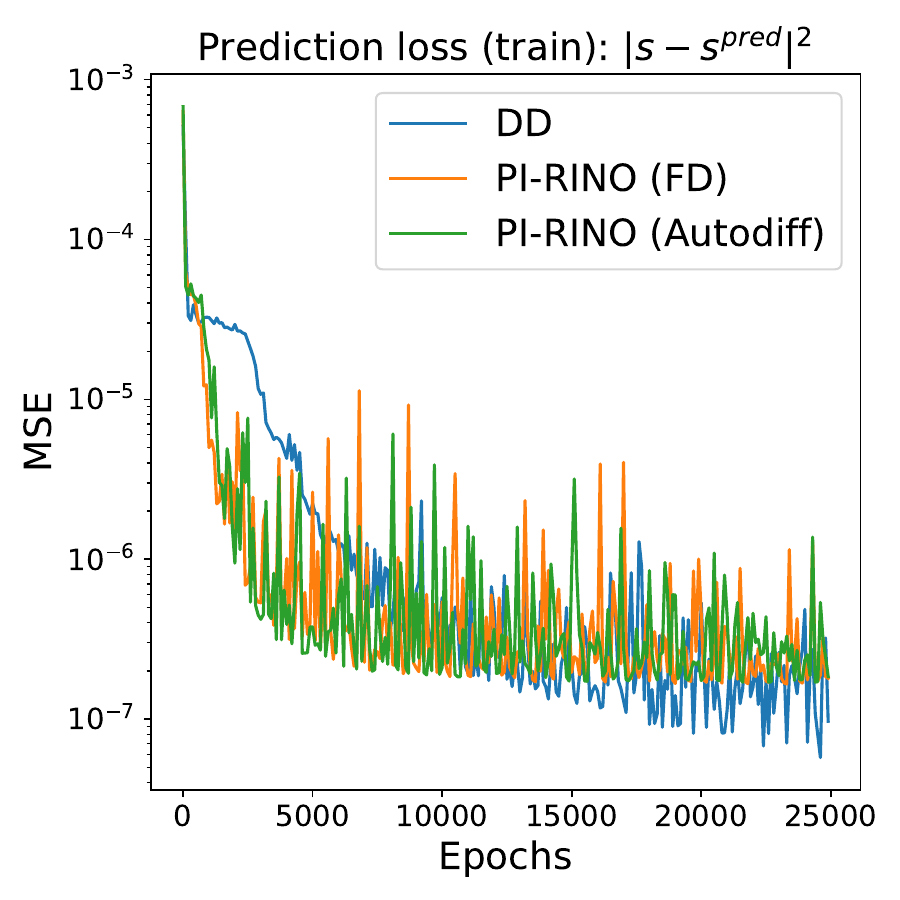}
        \caption{}
        \label{fig:2d_physics_compare_fd_auto_pred_loss_train}
    \end{subfigure}
    \begin{subfigure}[b]{0.45\textwidth}
        \centering
        \includegraphics[width=0.85\linewidth]{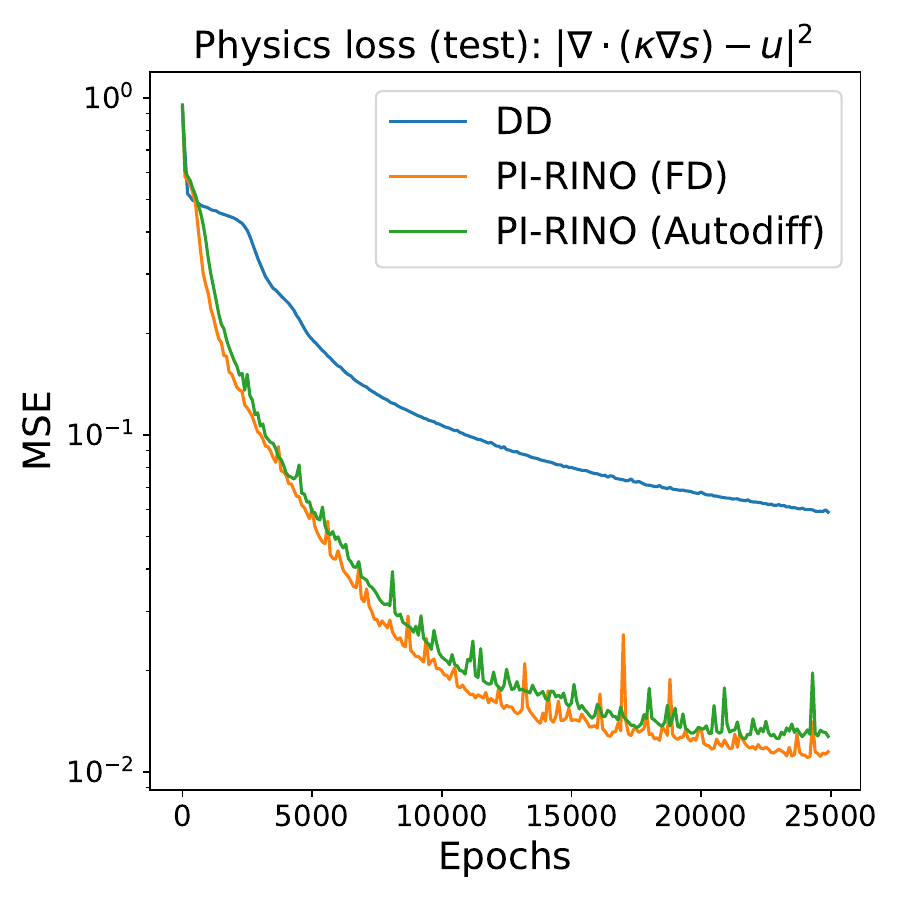}
        \caption{}
        \label{fig:2d_physics_compare_fd_auto_physics_loss_test}
    \end{subfigure}
    \begin{subfigure}[b]{0.45\textwidth}
        \centering
        \includegraphics[width=0.85\linewidth]{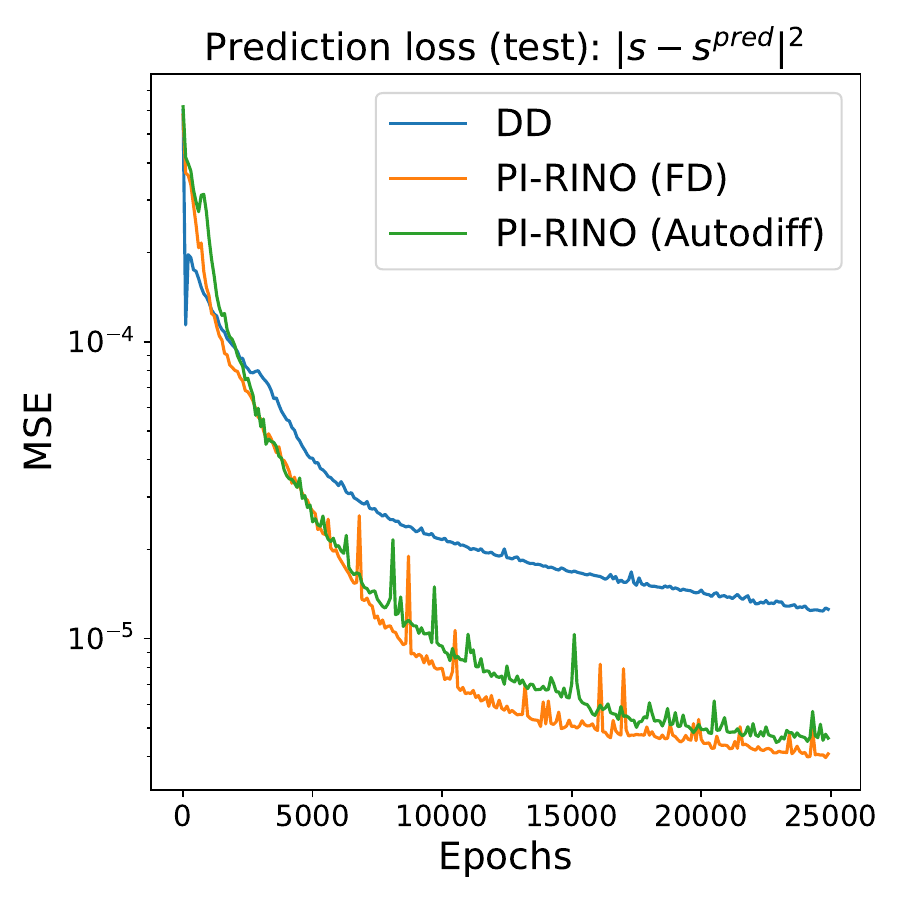}
        \caption{}
        \label{fig:2d_physics_compare_fd_auto_pred_loss_test}
    \end{subfigure}
    \begin{subfigure}[b]{0.45\textwidth}
        \centering
        \includegraphics[width=0.85\linewidth]{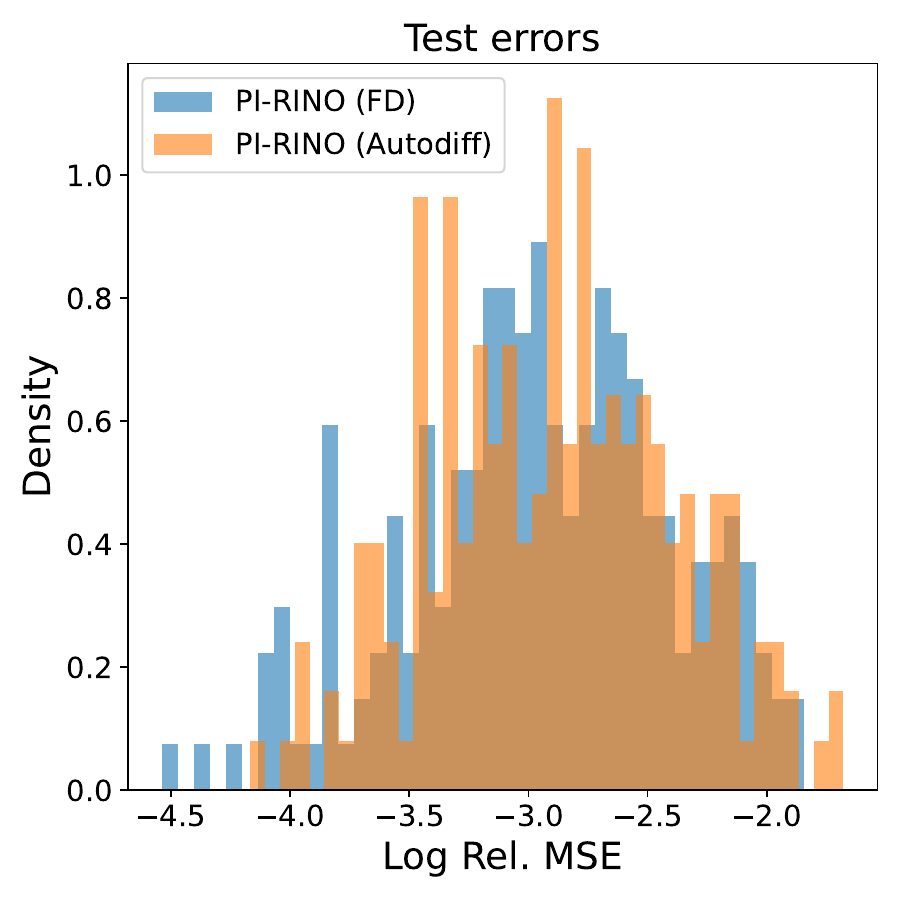}
        \caption{}
        \label{fig:2d_density_compare_fd_auto}
    \end{subfigure}

    \caption{2D Heat: Comparison of the two PI-RINO frameworks—one using automatic differentiation (Autodiff) and the other using finite differences (FD) for physics enforcement—with a data-driven (DD) baseline - convergence of physics and prediction losses for training dataset (a and b), testing dataset (c and d); and distribution of output prediction errors (test) for the PI-RINO models (e).}
    \label{fig:2d_compare_fd_auto_convergence}
\end{figure}

Tables~\ref{table:1d_autodiff_compare_antiderrivative} and \ref{table:2d_autodiff_compare_antiderrivative} compare the performance of the PI-RINO using FD and Autodiff for the antiderivative and 2D heat equation, respectively. Both tables present the relative cost (where again cost is scaled such that Cost $=1$ for FD), as well as the prediction errors on the training and test sets. These error measures include the mean $\pm1$ standard deviation testing and training error (computed over the full training/test datasets), the maximum training/test error, and the first quartile errors. In all cases, the testing and training errors between FD and Autodiff are comparable. There is no discernible advantage to either method from the perspective of the errors.  Crucially, the Cost of Autodiff is approximately $2.3\times$ greater than FD for the antiderivative problem and more than $10\times$ greater for the heat equation.
Hence, by constraining the physics with FD we realize a considerable savings that is expected due to the additional computational overhead of backpropagating gradients through the entire network. Moreover, both frameworks exhibit similar convergence in physics loss and prediction loss,  as illustrated in Figures~\ref{fig:1d_density_compare_fd_auto} and \ref{fig:2d_density_compare_fd_auto}, and significantly outperform a purely data-driven (DD) baseline in enforcing physical consistency on both the training set. These observations suggest that using FD in place of Autodiff does not result in any substantial loss of physical fidelity. Interestingly, the physics-informed frameworks also yield better convergence in the prediction loss compared to the DD baseline on the test set for the 2D heat example(Figure~\ref{fig:2d_physics_compare_fd_auto_pred_loss_test}). 

%The performance of the Autodiff-based PI-RINO framework was also analyzed on the testing dataset for both the problems, results of which are summarized in Appendix~\ref{app:autodiff_results}. 

\begin{table}[!hbt]
\centering
\begin{tabular}{@{}lcc@{}}
\toprule
\textbf{Metric} & \textbf{PI-RINO (FD)} & \textbf{PI-RINO (Autodiff)} \\
\midrule
Cost      & 1 & 2.30 \\
\midrule
\textit{Test Performance} & & \\
Mean ± std. dev. of test error & (7.08e-04 ± 3.77e-03) & (5.74e-04 ± 2.61e-03) \\
Maximum test error              & 9.22e-02 & 5.36e-02 \\
25th percentile test error      & 3.41e-04 & 2.16e-05 \\
\midrule
\textit{Training performance} & & \\
Mean ± std. dev. of train error & (1.75e-04 ± 4.04e-04) & (2.00e-04 ± 5.51e-04)\\
Maximum training error           & 2.90e-03 & 5.53e-03 \\
25th percentile training error   & 1.18e-04 & 8.41e-06 \\
\bottomrule
\end{tabular}
\caption{Antiderivative: Comparison of computational cost and output functions prediction errors for the proposed PI-RINO implemented with FD and Autodiff.}
\label{table:1d_autodiff_compare_antiderrivative}
\end{table}

\begin{table}[!hbt]
\centering
\begin{tabular}{@{}lcc@{}}
\toprule
\textbf{Metric} & \textbf{PI-RINO (FD)} & \textbf{PI-RINO (Autodiff)} \\
\midrule
Cost      & 1 & 10.32 \\
\midrule
\textit{Test Performance} & & \\
Mean ± std. dev. of test error & (2.19e-03 ± 2.63e-03) & (2.47e-03 ± 3.24e-03) \\
Maximum test error              & 1.43e-02 & 2.10e-02 \\
25th percentile test error      & 2.67e-03 & 2.94e-03 \\
\midrule
\textit{Training performance} & & \\
Mean ± std. dev. of train error & (1.15e-04±1.41e-04) & (1.04e-04 ± 1.29e-04)\\
Maximum training error           & 1.20e-03 & 1.20e-03 \\
25th percentile training error   & 1.34e-04 & 1.23e-04 \\
\bottomrule
\end{tabular}
\caption{2D Heat Example: Comparison of computational cost and output functions prediction errors for the proposed PI-RINO implemented with FD and Autodiff.}
\label{table:2d_autodiff_compare_antiderrivative}
\end{table}

\textbf{Remark.} Here we develop an alternative approach to formulating the physics-informed loss function using numerical finite-difference schemes in place of the traditionally used Autodiff. The motivation is to reduce the computational overhead associated with backpropagating higher-order derivatives through the entire network. We do not claim that the proposed finite-difference approach (restricted here to first- and second-order schemes) is universally the best option. A systematic comparison of numerical methods for enforcing physics constraints remains an interesting and open research direction, which the authors are actively pursuing. It should also be noted that in certain scenarios higher-order discretization schemes (e.g., Runge–Kutta or related approaches) may be required, potentially increasing the computational cost of finite-difference enforcement. Furthermore, the problems considered in this study are pedagogical in nature and limited to square domains, where structured sampling is straightforward. In more complex geometries, the cost of discretization and mesh generation can be substantial, in which case Autodiff may remain the more practical choice.

\section{Conclusion}\label{sec:conclusion}

In this study, we present a physics-informed operator learning framework that maps unevenly distributed (point cloud) input data to candidate solution functions governed by an underlying partial differential equation (PDE). This is achieved by first projecting the arbitrarily discretized input functions onto a latent embedding space of consistent dimensionality, which is learned and parameterized by neural networks. A second neural network then takes these embedded coordinates as input and approximates the action of the PDE operator in the \textit{physical} space. To enforce the governing PDEs, we integrate a finite-difference solver within the physical space, resulting in a surrogate model that emulates the underlying PDE operator using sparse and misaligned measurements of input functions. 
%The proposed framework demonstrates strong approximation accuracy on test data, with relative errors ranging from $\mathcal{O}(10^{-4})$ for simple 1D problems to $\mathcal{O}(10^{-2})$ for coupled systems. 
The accuracy of the proposed scheme on previously unseen data is assessed using multiple benchmark problems.
Furthermore, we compare the proposed method against the traditional autodifferentiation-based approach and show that it achieves comparable relative errors while offering significant computational speedups—approximately $2\times$ for 1D problems and up to $10\times$ for 2D problems. This work offers important insights that can potentially enable the extension of physics-informed operator learning to problems with complex geometries, sparse and missing data, multi-fidelity, and/or multi-modal (simulation + experimental) data sources.

\section*{Acknowledgments}

This material is based upon work supported by the U.S. Department of Energy, Office of Science, Office of Advanced Scientific Computing Research, under Award Number DE-SC0024162.

\section*{Author Contributions}

\textbf{Sumanta Roy}:Methodology, Software, Validation, Formal analysis, Investigation, Data Curation, Writing – original draft preparation, Writing – review and editing, Visualization.\\
\textbf{Bahador Bahmani}: Conceptualization, Methodology, Software, Validation, Formal analysis, Investigation, Data Curation, Supervision, Writing – original draft preparation, Writing – review and editing.\\
\textbf{Ioannis G. Kevrekidis}: Validation, Formal analysis, Investigation, Resources, Supervision, Writing – review and editing, Funding Acquisition, Project Administration.\\
\textbf{Michael D. Shields}: Conceptualization, Validation, Formal analysis, Investigation, Resources, Supervision, Writing – review and editing, Funding Acquisition, Project Administration.

\bibliographystyle{unsrtnat}
\bibliography{bibliography}

%% Everything about Appendix here: %%
\section*{Appendix}
\appendix
\section{Dictionary learning for the input function embeddings}\label{app:dictionary_learning}

We briefly summarize the dictionary learning algorithm introduced by \cite{bahmani2025resolution} which we use in this study as Step-A for embedding the multi-resolution input function data which is available to us a point cloud as detailed in Section~\ref{sec:operator_learning_dataset}. It features a collection of $N$ input signals denoted by $\mathcal{D}_\text{in} = \{\mathcal{D}_\text{in}^{(i)}\}_{i=1}^N$, where each element $\mathcal{D}_\text{in}^{(i)}$ contains an arbitrarily discretized version of the $i$-th function realization $\boldsymbol{u}^{(i)}$ (see Figure~\ref{fig:multi_fidelity_dataset}). The aim is to adaptively learn a set of basis functions which represents (reconstructs) the signals from the point cloud data available. These basis functions are represented by artificial neural networks. We want to build the dictionary $\boldsymbol{\Psi}(\boldsymbol{x}) = \{\psi_l(\boldsymbol{x})\}_{l=1}^P$, where $P = |\boldsymbol{\Psi}|$, which will basically be a common set of $P$ basis functions, such that  $\boldsymbol{\Psi(x)} = \{\psi_l(\boldsymbol{x})\}_{l=1}^P \equiv [\psi_1(\boldsymbol{x}), \ldots, \psi_P(\boldsymbol{x})]^T$. These are defined \textit{continuously} in the domain of $u$, such that the input signal realizations can be represented (and reconstructed) sufficiently well by a linear combination of these basis functions, as follows:
\begin{equation*}
    \tilde{u}^{(i)}(\boldsymbol{x}) \approx \sum_{l=1}^P \alpha_l^{(i)} \, \psi_l(\boldsymbol{x}) = \boldsymbol{\Psi}^T(\boldsymbol{x})\, \boldsymbol{\alpha}^{(i)}
\end{equation*}
where $\alpha_l^{(i)}$ is the coefficient corresponding to the $l$-th basis function for the $i$-th signal realization, and the job is basically to find these these basis functions and the corresponding coefficients which act as embeddings for our operator. There are a number of ways of choosing the basis-functions, for example, we can choose orthogonal Fourier bases~\cite{konidaris2011value,huang2011approximation}, orthogonal polynomials~\cite{masjedjamei2002three,canuto1982approximation}, Legendre-polynomials~\cite{schweizer2021legendre,wang2021much}, radial basis functions~\cite{d2024learning,chakravarthy1997function} etc. In particular, orthogonal basis functions $\psi_l(\boldsymbol{x}; \boldsymbol{\theta}_l)$ which are parameterized by neural networks (with parameters $\boldsymbol{\theta}_l$ for the $l$-th neural network basis function) are used, which are learnt in a pure data-driven manner. Thus we obtain:
\begin{equation}
    \tilde{u}^{(i)}(\boldsymbol{x}) \approx \boldsymbol{\Psi}^T(\boldsymbol{x}; \boldsymbol{\theta}) \, \boldsymbol{\alpha}^{(i)},
\end{equation}
where $\boldsymbol{\theta}$ concatenates the parameters of all neural network basis functions. The basis functions are learned successively (orthogonal basis functions are added successively thus reducing the total approximation error). The embedding coefficients are obtained by constructing an error function as follows:
\begin{equation}
\boldsymbol{\alpha} = \text{Proj}_{\Psi}[u(\boldsymbol{x})] \triangleq \arg\min_{\boldsymbol{\alpha}} \left\| u(\boldsymbol{x}) - \boldsymbol{\Psi}^T(\boldsymbol{x}; \boldsymbol{\theta}) \, \boldsymbol{\alpha} \right\|_2^2 + \mathcal{R}(\boldsymbol{\alpha}).
\end{equation}
In the above equation, the projection error between the original signal $u(\boldsymbol{x})$ and its projection $\boldsymbol{\Psi}^T(\boldsymbol{x}; \boldsymbol{\theta})\boldsymbol{\alpha}$ is reduced using the $L_2$-norm as the error measure with respect to a well-defined inner product $\|r(x)\|_2^2 = \langle r(x), r(x) \rangle$. The projection operator is given as $\text{Proj}_{\Psi}(\cdot) : \mathcal{U} \mapsto \mathbb{R}^{|\boldsymbol{\Psi}|}$ with respect to the dictionary $\boldsymbol{\Psi}$ with $|\boldsymbol{\Psi}| = Q$ number of basis-functions (neural networks), which returns the coordinates of the projection of the input function $u(\boldsymbol{x})$ onto the subspace spanned by the dictionary’s basis functions. The second term in the loss function is the regulation term, which improves the stabulity of the optimization while promoting sparsity. Let us choose \( \mathcal{R}(\boldsymbol{\alpha}) = \lambda \|\boldsymbol{\alpha}\|_2^2 \), i.e., \( L_2 \)-regularization. Then the objective function becomes:
\begin{align*}
J(\boldsymbol{\alpha}) &= \| \boldsymbol{u} - \Psi^T \boldsymbol{\alpha} \|_2^2 + \lambda \| \boldsymbol{\alpha} \|_2^2. \\
&= (\boldsymbol{u} - \Psi^T \boldsymbol{\alpha})^T (\boldsymbol{u} - \Psi^T \boldsymbol{\alpha}) + \lambda \| \boldsymbol{\alpha} \|_2^2 \\
&= \boldsymbol{u}^T \boldsymbol{u} - 2 \boldsymbol{\alpha}^T \Psi \boldsymbol{u} + \boldsymbol{\alpha}^T \Psi \Psi^T \boldsymbol{\alpha} + \lambda \| \boldsymbol{\alpha} \|_2^2. 
\end{align*}
The gradient of $J(\boldsymbol{\alpha})$ is set to zero, and is solved for $\boldsymbol{\alpha}$, to obtain:
\begin{align*}
\Rightarrow \nabla_{\boldsymbol{\alpha}} J &= -2 \boldsymbol{\Psi} \boldsymbol{u} + 2 (\boldsymbol{\Psi} \boldsymbol{\Psi}^T + \lambda \boldsymbol{I}) \boldsymbol{\alpha} \\
&= 0 \\ 
\Rightarrow (\boldsymbol{\Psi} \boldsymbol{\Psi}^T + \lambda \boldsymbol{I}) \boldsymbol{\alpha} &= \boldsymbol{\Psi} \boldsymbol{u} \\
\boxed{
\boldsymbol{\alpha} = (\boldsymbol{\Psi \Psi}^T + \lambda \boldsymbol{I})^{-1} \boldsymbol{\Psi} \boldsymbol{u}
}
\end{align*}
which can also be written as: 
\begin{equation}
\alpha_l = \text{Proj}_{\psi_l}[u(\boldsymbol{x})] = \frac{ \langle u(\boldsymbol{x}), \psi_l(\boldsymbol{x}) \rangle }{ \langle \psi_l(\boldsymbol{x}), \psi_l(\boldsymbol{x}) \rangle + \lambda }, \quad \text{for } l = 1, \dots, Q.
\end{equation}
It is to be noted that we need to find both the basis functions and the coefficients, in the learning process which makes the optimization challenging. This is taken care of by the staggered approach, where in the first step we consider the basis function parameters as fixed, and find the corresponding coordinates. Then, we keep the obtained coefficients fixed and train the basis function neural networks. This is shown in Algorithm~\ref{alg:pi_rino}.

Each of the basis functions $\psi_l$ are parameterized using SIRENs~\cite{sitzmann2020implicit}, as given by:
\begin{equation}
\psi_l(\boldsymbol{x}; \boldsymbol{\theta}_l) = \boldsymbol{W}^{(l)}_{n+1} \left( \phi_n \circ \phi_{n-1} \circ \cdots \circ \phi_1 \right)(\boldsymbol{x}) + \boldsymbol{\beta}^{(l)}_{n+1}; \quad \phi_m(h_{m-1}^{(l)}) \triangleq \sin\left( \boldsymbol{W}^{(l)}_m h^{(l)}_{m-1} + \boldsymbol{\beta}^{(l)}_m \right), 
\end{equation}
where $h^{(l)}_m$ denotes the $m$-th hidden state, with $h_0 \triangleq \boldsymbol{x}$ and $0 \le m \le n$. $\boldsymbol{W}^{(l)}_m$ and $\boldsymbol{\beta}^{(l)}_m$ represent the typical weights and biases of the MLP, respectively. The vector $\boldsymbol{\theta}_l$ concatenates all the trainable parameters of the SIREN, i.e., $\boldsymbol{\theta}_l \equiv \left\{ \boldsymbol{W}^{(l)}_m, \boldsymbol{\beta}^{(l)}_m \right\}_{m=1}^{n+1}$.

\begin{remark}
    \label{remark:orthogonal_basis_function}
    \textit{Two functions $\psi_i(x)$ and $\psi_j(x)$ are orthogonal over the domain $x$ if their inner product $\langle \psi_i, \psi_j \rangle$ is zero for $i \ne j$, i.e.,}
\begin{equation}
    \langle \psi_i, \psi_j \rangle \triangleq \int_x \psi_i(x)\, \psi_j(x) \, dx = 0, \quad \text{if } i \ne j. \tag{10}
\end{equation}
\end{remark}

\section{Finite difference physics losses}\label{app:fd_loss}

In this section, we describe the finite difference (FD) method used as a replacement for automatic differentiation in computing the physics loss term in Eq.~\eqref{eq:pi-rino_loss_function}. %Traditional FD methods rely on discretizing the domain using a structured mesh, where mesh points correspond to the intersections of horizontal and vertical lines—see Fig.~\ref{fig:fd_convolution_schematic} for a schematic illustration in two dimensions.

To enforce the physics loss, we generate structured collocation points $\boldsymbol{y}_{[p,q]}$, where $p$ and $q$ denote the grid indices along the $\boldsymbol{y_1}$ and $\boldsymbol{y_2}$ directions, respectively. At these collocation points, we evaluate the neural network output $s_{[p,q]}$ (more precisely, $s_{\boldsymbol{\theta}:[p,q]}^{\text{pred}}$) and the reconstructed input function $u_{[p,q]}$ (i.e., $\tilde{u}_{[p,q]}$). Following the FD methodology, we construct a finite difference stencil which, when centered at a collocation point, replaces the continuous derivatives in the PDE with discrete difference quotients. These quotients involve values of the solution function (and the reconstructed input) at surrounding mesh points. For instance, consider the representative PDE $\Delta^2 s = u$. The physics loss for the $i$-th sample can be expressed as:
\begin{equation}
    \mathcal{L}_{\text{pde}}^{(i)} = \left | \frac{\partial^2 s^{(i)}}{\partial y_1^2} + \frac{\partial^2 s^{(i)}}{\partial y_2^2} - u^{(i)} \right |^2.
\end{equation}
By applying the FD stencil at the collocation point $\boldsymbol{y}_{[p,q]}$, the physics loss at that point becomes:
\begin{equation}
    \mathcal{L}_{\text{pde}:[p,q]}^{(i)} \approx \Bigg| 
    \frac{s^{(i)}_{[p+1,q]} - 2s^{(i)}_{[p,q]} + s^{(i)}_{[p-1,q]}}{\Delta y_1^2} 
    + 
    \frac{s^{(i)}_{[p,q+1]} - 2s^{(i)}_{[p,q]} + s^{(i)}_{[p,q-1]}}{\Delta y_2^2}
    - u^{(i)}_{[p,q]} 
    \Bigg|^2.
\end{equation}
The total FD-based physics loss is then computed by convolving this stencil across all interior collocation points in the domain and summing the individual contributions:
\begin{equation}
    \mathcal{L}_{\text{pde}}^{(i)} \approx \sum_{p,q} \mathcal{L}_{\text{pde}:[p,q]}^{(i)}.
\end{equation}
The table below summarizes the governing PDEs and their corresponding physics loss expressions at the $[p,q]$-th collocation point for all examples presented in this study:

\begin{table}[!hbt]
\renewcommand{\arraystretch}{1.9}
\small
\centering
\caption{Physics-informed loss expressions used for different PDE cases.}
\label{tab:fd_losses}
\begin{tabular}{|p{3.0cm}|p{3.7cm}|p{8.8cm}|}
\hline
\textbf{Case} & \textbf{PDE} & $\boldsymbol{\mathcal{L}^{(i)}_{\text{pde}: [p,q]}}$ \\
\hline

\rule{0pt}{3.5ex}\textbf{Anti-derivative} 
& $\displaystyle  \frac{ds}{dy} = u(y)$ 
& $\displaystyle \Bigg| \frac{s^{(i)}_{[p+1]} - s^{(i)}_{[p-1]}}{2\Delta y} - u^{(i)}_{[p,q]}\Bigg|^2$  \\
\hline

\rule{0pt}{3.5ex}\textbf{2D heat conduction} 
& $\displaystyle \kappa \left( \frac{\partial^2 s}{\partial y_1^2} + \frac{\partial^2 s}{\partial y_2^2} \right) = u(y_1,y_2)$ 
& $\displaystyle
\Bigg| \kappa \left(
\frac{s^{(i)}_{[p+1,q]} - 2s^{(i)}_{[p,q]} + s^{(i)}_{[p-1,q]}}{\Delta y_1^2}
+
\frac{s^{(i)}_{[p,q+1]} - 2s^{(i)}_{[p,q]} + s^{(i)}_{[p,q-1]}}{\Delta y_2^2}
\right)
- u^{(i)}_{[p,q]}\Bigg|^2$ \\
\hline

\rule{0pt}{3.5ex}\textbf{Biot’s consolidation}\tablefootnote{In this case, the input function (multi-resolution) is $\kappa(x)$, while $u(y,t)$ and $p(y,t)$ are the solution functions. Readers should note the slight change in notation used here.}
& 
\(
\begin{aligned}
\frac{\partial}{\partial y} \left( \nu \frac{\partial u}{\partial y} \right) + \frac{\partial p}{\partial y} &= 0 \\
\frac{\partial}{\partial t} \left( a p + \frac{\partial u}{\partial y} \right)
- \frac{\partial}{\partial y} \left( \kappa \frac{\partial p}{\partial y} \right) &= 0
\end{aligned}
\)
& 
\(
\begin{aligned}
\Bigg|\gamma \left( \frac{u^{(i)}_{[p+1,q]} - 2u^{(i)}_{[p,q]} + u^{(i)}_{[p-1,q]}}{\Delta y^2} \right)
+ \left( \frac{p^{(i)}_{[p+1,q]} - p^{(i)}_{[p-1,q]}}{2\Delta y} \right)\Bigg|^2 \\
\Bigg|a \left( \frac{p^{(i)}_{[p,q]} - p^{(i)}_{[p,q-1]}}{\Delta t} \right)
+ \left( \frac{u^{(i)}_{[p+1,q]} - u^{(i)}_{[p+1,q-1]} - u^{(i)}_{[p-1,q]} + u^{(i)}_{[p-1,q-1]}}{2\Delta y \Delta t} \right) \\
- \kappa^{(i)}_{[p,q]} \left( \frac{p^{(i)}_{[p,q+1]} - 2p^{(i)}_{[p,q]} + p^{(i)}_{[p,q-1]}}{\Delta y^2} \right) \Bigg|^2
\end{aligned}
\)
\\
\hline
\end{tabular}
\end{table}

\section{Hyper-parameter settings}\label{app:hyperparameters}
\begin{table}[!hbt]
\centering
\caption{Hyper-parameter settings for the examples outlined in this study.}
\label{tab:hyperparams}
\begin{tabular}{|l|c|c|c|}
\hline
\textbf{Hyper-parameter} & \textbf{Anti-derivative operator} & \textbf{2D heat conduction} & \textbf{Biot's consolidation} \\
\hline
\# training samples  & 150 & 800 & 500 \\
$[M_{min},M_{max}]$ & [10,60] & [100, 280] & [35,55] \\
Embedding size & 10 & 57 & 10 \\
Activation function & Mish & Mish & Tanh \\
Collocation points & 100 & 400 & 5625 \\
Epochs & 25000 & 25000 & 50000 \\
Learning rate & 5e-5 & 1e-4 & 1e-4\\
Layer architecture & [11,128,128,128,1] & [59,128,128,128,128,1] & [12,100,100,100,1]$\times$2 \\
\hline
\end{tabular}
\end{table}

\end{document}